\documentclass[11pt]{article}
\usepackage[preprint]{acl}
\usepackage{tabularx}

\usepackage[T2A,T1]{fontenc}
\usepackage{times}
\usepackage{latexsym}
\usepackage[utf8]{inputenc}
\usepackage{microtype}
\usepackage{inconsolata}
\usepackage{graphicx}
\usepackage{booktabs}
\usepackage{adjustbox}
\usepackage{tabularx}
\usepackage{color, colortbl}
\usepackage{makecell}

\usepackage{todonotes}
\usepackage{enumitem}
\usepackage{subcaption}
\usepackage[nameinlink,capitalize,noabbrev]{cleveref}
\usepackage{comment}
\usepackage{multirow}
\usepackage{paralist} 
\usepackage{longtable}
\usepackage{pdfpages}
\newcommand{\HPLT}{HPLT~v2}
\newcommand{\LUMI}{LUMI}

\usepackage{acronym}
\acrodef{clm}[CLM]{causal language model}
\acrodef{cc}[CC]{Common Crawl}
\acrodef{cc0}[CC0]{Creative Commons Zero}
\acrodef{ia}[IA]{Internet Archive}
\acrodef{lid}[LID]{language identification}
\acrodef{lm}[LM]{language model}
\acrodef{llm}[LLM]{large language model}
\acrodef{mlm}[MLM]{masked language model}
\acrodef{mt}[MT]{machine translation}
\acrodef{nlp}[NLP]{natural language processing}
\acrodef{nlu}[NLU]{natural language understanding}
\acrodef{tld}[TLD]{top-level domain}

\title{An Expanded Massive Multilingual Dataset for \\High-Performance Language Technologies (HPLT)}

\author{
    \textbf{Laurie Burchell}\textsuperscript{1},
    \textbf{Ona de Gibert}\textsuperscript{2},
    \textbf{Nikolay Arefyev}\textsuperscript{3},
    \textbf{Mikko Aulamo}\textsuperscript{2},
    \textbf{Marta Bañón}\textsuperscript{4},
    \\
    \textbf{Pinzhen Chen}\textsuperscript{1},
    \textbf{Mariia Fedorova}\textsuperscript{3},
    \textbf{Liane Guillou}\textsuperscript{1},
    \textbf{Barry Haddow}\textsuperscript{1},
    \textbf{Jan Haji\v{c}}\textsuperscript{5},
    \\
    \textbf{Jind\v{r}ich Helcl}\textsuperscript{5},
    \textbf{Erik Henriksson}\textsuperscript{6},
    \textbf{Mateusz Klimaszewski}\textsuperscript{1},
    \textbf{Ville Komulainen}\textsuperscript{6},
    \\
    \textbf{Andrey Kutuzov}\textsuperscript{3},
    \textbf{Joona Kytöniemi}\textsuperscript{6},
    \textbf{Veronika Laippala}\textsuperscript{6},
    \textbf{Petter Mæhlum}\textsuperscript{3},
    \\
    \textbf{Bhavitvya Malik}\textsuperscript{1},
    \textbf{Farrokh Mehryary}\textsuperscript{6},
    \textbf{Vladislav Mikhailov}\textsuperscript{3},
    \textbf{Nikita Moghe}\textsuperscript{7},
    \\
    \textbf{Amanda Myntti}\textsuperscript{6},
    \textbf{Dayyán O'Brien}\textsuperscript{1},
    \textbf{Stephan Oepen}\textsuperscript{3},
    \textbf{Proyag Pal}\textsuperscript{1},
    \textbf{Jousia Piha}\textsuperscript{6},
    \\
    \textbf{Sampo Pyysalo}\textsuperscript{6},
    \textbf{Gema Ramírez-Sánchez}\textsuperscript{4},
    \textbf{David Samuel}\textsuperscript{3},
    \textbf{Pavel Stepachev}\textsuperscript{1},
    \\
    \textbf{Jörg Tiedemann}\textsuperscript{2},
    \textbf{Dušan Vari\v{s}}\textsuperscript{5},
    \textbf{Tereza Vojt\v{e}chová}\textsuperscript{5},
    \textbf{Jaume Zaragoza-Bernabeu}\textsuperscript{4}
\\
 \textsuperscript{1}University of Edinburgh,
\textsuperscript{2}University of Helsinki,
 \textsuperscript{3}University of Oslo,
 \\
 \textsuperscript{4}Prompsit Language Engineering,
 \textsuperscript{5}Charles University,
 \textsuperscript{6}University of Turku,
 \textsuperscript{7}Amazon\thanks{Work was done prior to joining Amazon.}
\\
 \small{
   \textbf{Contact:} \url{https://hplt-project.org}
 }
}

\begin{document}

\maketitle

\begin{abstract}

Training state-of-the-art large language models
requires vast amounts of clean and diverse textual data. However, building suitable multilingual datasets remains a challenge. In this work, we present \HPLT{}, a collection of high-quality multilingual monolingual and parallel corpora, extending prior work of the HPLT project. 
The monolingual portion of the data contains 8T tokens covering 193 languages, while the parallel data contains  380M sentence pairs covering 51 languages. We document the entire data pipeline and release the code to reproduce it. We provide extensive analysis of the quality and characteristics of our data. Finally, we evaluate the performance of language models and machine translation systems trained on \HPLT{}, demonstrating its value.

%We conduct extensive analysis through indirect quality indicators, human evaluation and automatic register labeling. Additionally, we perform evaluations of language models trained on our data, including masked and autoregressive models, and machine translation systems.
\end{abstract}

\section{Introduction}
% Laurie 
In order to train the state-of-the-art \acfp{llm} required for modern NLP, large amounts of high-quality textual training data are essential. However, obtaining a sufficient quantity of such data is far from easy. In addition, effective NLP research requires open training data so that results can be replicated and verified.

% In this paper, we build on the work of \citet{de-gibert-etal-2024-new} (hereafter referred to as HPLT~v1.2), by downloading and processing 4.5 PB of web crawl data sourced from Internet Archive\footnote{\url{https://archive.org}} and Common Crawl.
% \footnote{\url{https://commoncrawl.org}} 
In this paper, we introduce a new set of text corpora dubbed \HPLT{}.\footnote{\url{https://hplt-project.org/datasets/v2.0}} It is extracted from 4.5 petabytes (PB) of the \ac{ia}\footnote{\url{https://archive.org}} and \ac{cc}\footnote{\url{https://commoncrawl.org}} data. 
We build on the work of \citet{de-gibert-etal-2024-new} (hereafter referred to as HPLT~v1.2) with an improved extraction pipeline and a much larger set of input crawls to produce the \HPLT{} collection of monolingual and parallel corpora. To our knowledge, our new corpus is the only large-scale text collection extracted from the \ac{ia}, apart from HPLT~v1.2.  We release \HPLT{} under the permissive \ac{cc0} license\footnote{\url{https://creativecommons.org/share-your-work/public-domain/cc0/}. We do not claim ownership of any of the text from which this data has been extracted.} and provide the code to replicate our pipeline. Our main contributions can be summarised as:
\begin{compactitem}
    \item We release monolingual corpora covering 193 languages and containing approximately 52 trillion characters and 8 trillion tokens.% sourcing our data from both the \ac{ia} and \ac{cc}.
    \item We derive parallel corpora from our monolingual data for 50 languages paired with English, containing over 380 million sentence pairs.
    \item We make the tools and pipelines used to create the collection openly available.\footnote{\url{https://github.com/hplt-project/HPLT-textpipes}}
    \item We conduct an in-depth analysis of our data including descriptive statistics, manual inspection, and automatic register labelling.
    \item We demonstrate the quality of \HPLT{} by using it to train a range of high-performing language and machine translation models.
\end{compactitem}

\begin{figure*}[thbp]
    \centering
    \includegraphics[width=\linewidth]{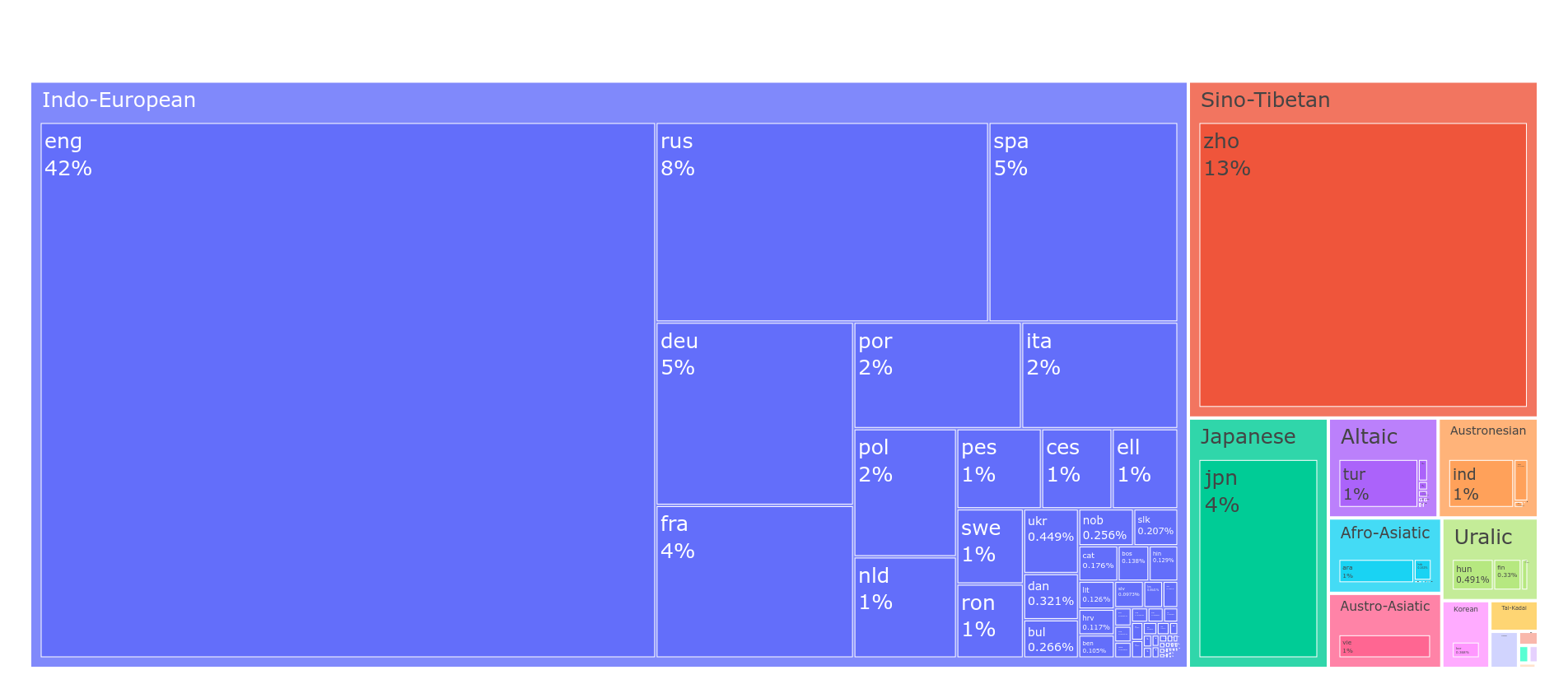}
    \caption{The distribution of documents in the \HPLT{} cleaned dataset by language family and variety. Shortened ISO 639-3 language codes are used here; see the full plots at \url{https://hplt-project.org/datasets/v2.0} }
    \label{fig:lang_families}
\end{figure*}

\section{Related work}
\label{sec:related_work}
% Laurie and Ona

The increasing data demands of state-of-the-art \acp{llm} have driven a rapid growth in both the number and the size of text corpora. We provide a summary of some well-known collections in \Cref{sec:multi_collections}. Whilst \acp{llm} trained on ostensibly English data have shown impressive multilingual capabilities \cite{armengol-estape-etal-2022-multilingual}, of particular relevance to this work is the growing shift towards explicitly multilingual corpora. Compared with earlier efforts (e.g. OSCAR \cite{oscar}, CC-100 \cite{conneau-etal-2020-unsupervised} and mC4 \cite{xue-etal-2021-mt5}), more recent multilingual datasets cover increasing numbers of languages, e.g. CulturaX \cite{culturax} and MADLAD-400 \cite{madlad400}. %Glot500-C \cite{glot500}, the MaLA Corpus \cite{ji2024emma} and Serengeti \cite{serengeti}.
\HPLT{} continues this trend by aiming for significant coverage of a wide range of languages. We note that the majority of previous multilingual datasets are sourced from \ac{cc}, whereas much of \HPLT{} is composed of \ac{ia} crawls. This means that \HPLT{} can be used in conjunction with these existing datasets as a complementary source.

Producing large-scale datasets by crawling the Web is helpful for scale, but raises questions around dataset quality such as the prevalence of boilerplate, explicit material or non-linguistic content \cite{kreutzer2022quality}. One way to tackle low-quality data is through human audit and curation (e.g. ROOTS \cite{roots}, Glot500-c \cite{glot500}, Serengeti \cite{serengeti} and the MaLA Corpus \cite{ji2024emma}). However, such an approach is difficult to scale. Instead, we ensure the quality of \HPLT{} through a robust dataset construction pipeline (\Cref{sec:construction}) and by verifying our data through extensive analysis and downstream evaluation (\Cref{sec:analysis,sec:evaluation}).

% % Why is parallel data relevant to this day?
% Finally, we also present a collection of high-quality parallel data.
% %, which remains crucial in enhancing multilingual capabilities of current models.
% While the traditional encoder-decoder paradigm has shifted into Casual Language Models (CLM) with decoder-only architectures that primarily rely on monolingual data, recent studies have demonstrated that incorporating parallel data during the pretraining stage significantly boosts their performance on MT, downstream multilingual and cross-lingual tasks \cite{kale-etal-2021-nmt5,  briakou-etal-2023-searching}.

% %\paragraph{Our work}
% Our work fits in this scenario by providing a web-crawled multilingual collection \HPLT{} designed to address key challenges in existing datasets. 
% %While our dataset does not cover as many languages as some larger collections, 
% Our dataset significantly increases the quantity of per-language data.
% By balancing scale, diversity, and quality, our approach aims to set a new standard for multilingual data creation.

In addition to large-scale monolingual data in multiple languages, \HPLT{} contains high-quality parallel data. Whilst \acp{clm} with decoder-only architectures rely primarily on monolingual data, recent studies have shown that incorporating parallel data during the pretraining stage significantly boosts multilingual, cross-lingual and \ac{mt} performance for such models \cite{kale-etal-2021-nmt5,  briakou-etal-2023-searching, alves_tower_2024}. Because of this,
we expect that there is still significant demand for parallel data. 
%that the parallel portion of \HPLT{} remains a valuable resource.

This work is a direct successor of our HPLT~v1.2 dataset introduced in \cite{de-gibert-etal-2024-new}. 
Compared to HPLT~v1.2, we feature more data (21 billion vs. 5 billion documents) using an improved pipeline (\Cref{sec:construction}), resulting in a significantly larger dataset (52 trillion characters compared to 42 trillion). The \HPLT{} collections are of higher quality than those in HPLT~v1.2, as shown through comparative analysis and evaluation (\Cref{sec:analysis,sec:evaluation}). For clarity, we list the main differences between HPLT~v1.2 and \HPLT{} here:

\begin{compactitem}
\item The size of the source web collections is 2.5x larger, totalling 4.5 petabytes of compressed web data.
\item The text extraction pipeline uses \texttt{Trafilatura} rather than \texttt{warc2text}, which results in more efficient boilerplate removal.
\item Language identification uses a modified version of \texttt{OpenLID} rather than \texttt{CLD2}, increasing coverage from 75 to 193 language varieties.
\item Documents are annotated with their compliance to the \texttt{robots.txt} files of the corresponding websites, which can be used to filter out documents explicitly forbidden for crawling by website owners. The cleaned variant of the \HPLT{} dataset contains only \texttt{robots.txt} compliant documents.
\item Deduplication is done at collection level rather than globally.
\item Documents are annotated for PII information.
\item Document quality scores computed with \texttt{web-docs-scorer} rather than segment-level language model based scores
\item Simpler and more interpretable filtering and cleaning criteria.
\end{compactitem}

\section{Dataset description} 
% Andrey for monolingual
In this section, we describe the \HPLT{} collection of monolingual and parallel corpora, before explaining how it was constructed in \Cref{sec:construction}. 

\subsection{Monolingual datasets}
\label{sec:mono_desc}
The monolingual portion of \HPLT{} covers 193 language varieties\footnote{Language varieties are labelled with an ISO 639-3 code denoting the variety plus an ISO 15924 four-letter code denoting the script, separated by an underscore: e.g., \texttt{gla\_Latn}.} and is published in two variants: `deduplicated' (21 terabytes) and `cleaned' (15 terabytes). In the latter variant, the documents filtered by our cleaning heuristics (see \Cref{sec:monolingual_proc}) are excluded. For training \acp{llm}, we recommend using the cleaned variant, but we also publish the datasets before cleaning (`deduplicated') so that it is possible to apply custom cleaning pipelines to the \HPLT{} data. In total, the deduplicated monolingual \HPLT{} datasets contain approximately 7.6 trillion white-space separated tokens and 52 trillion characters, extracted from 21 billion documents. \HPLT{} is published in the JSONL format, with one document per line.

\Cref{fig:lang_families} shows the distribution of documents in the cleaned monolingual data by language families and language variety. Indo-European languages, and especially English, make up the majority of the data. Unfortunately, this is the reality of current web crawls; increasing the amount of data available for other languages is not an easy task and is important future work. \Cref{tab:monostats_long} gives a full breakdown of the statistics of the monolingual data.

\begin{table}[t]
\centering\small
\setlength{\tabcolsep}{1.2ex}
\begin{tabular}{lrrrr}
\toprule
\textbf{} &\multicolumn{2}{c}{\textbf{Raw}} &\multicolumn{2}{c}{\textbf{Filtered}} \\
\cmidrule(lr){2-3}
\cmidrule(lr){4-5}
 &\textbf{Pairs} &\textbf{Eng. words} &\textbf{Pairs}  & \textbf{Eng. words} \\\midrule
 
\textbf{Total} & 1277M & 16849M & 380M &6780M \\
\textbf{Median} &11M & 170M &4M &80M \\
\bottomrule
\end{tabular}
\caption{Counts in millions (M) of sentence pairs and English words in the parallel \HPLT{} data before filtering (Raw) and after  filtering and deduplication (Filtered), both in total and the median across all languages.}
\label{tab:v2_bitext_data_stats_summary}
\end{table}

\subsection{Parallel datasets}
\label{sec:parallel_desc}
We extract parallel data from the monolingual \HPLT{} to cover 50 languages paired with English. We aimed for a diverse range of language varieties and scripts in the low to medium resource range (listed in Table~\ref{tab:v2_bitext_data_stats_full}). We align these to English since this configuration has the highest potential for finding high-quality parallel data. We release our data in both XML %TMX
%\footnote{Translation Memory Exchange \url{http://xml.coverpages.org/tmxSpec971212.html}}
and bitext format.\footnote{\url{https://opus.nlpl.eu/HPLT/corpus/version/HPLT}} 

\Cref{tab:v2_bitext_data_stats_summary} gives the number of sentence pairs and English words per language prior to filtering (Raw) and after processing (Filtered). We provide both the total over the entire dataset and the median count by language variety. Our results show that the deduplicated \HPLT{} parallel corpora have a 70\% reduction in sentence pairs compared to the raw data. The final dataset contains over 380 million sentence pairs, with the English side of the dataset containing over 6 billion words. The median number of sentence pairs by language variety is 4 million, but individual sizes vary greatly by language: the smallest, Sinhala, contains around 273 thousand pairs, whereas the largest, Finnish, contains over 29 million pairs. We give full statistics for each included language variety in \Cref{tab:v2_bitext_data_stats_full} in the appendix. 

We assume the large number of Finnish sentence pairs is due to the pipeline's bias toward European languages. In contrast, languages such as Japanese and Korean, which we would expect to have larger corpora, may have lower counts because of lower-quality monolingual data and limited support in key pipeline components such as sentence splitting and tokenization. This results in reduced yields during data cleaning and filtering for non-European languages written in non-Latin scripts.

\paragraph{Doc\HPLT{}} We also introduce context, by providing the documents that our bitext data comes from. These documents are annotated with both sentence and paragraph alignment. Overall, we provide 74,078,581 aligned documents covering all 51 languages. This ranges from 38,199,407 documents in English to 20,448 documents in Xhosa.\footnote{\url{https://opus.nlpl.eu/legacy/DocHPLT.php}}
% ONA

\paragraph{Multi\HPLT{}} We leverage the English-centric \HPLT{} parallel resources to further create a multi-way parallel corpus, obtained by pivoting via English sentences. This corpus includes 1275 language pairs and contains over 16.7 billion parallel sentences.\footnote{\url{https://opus.nlpl.eu/MultiHPLT/corpus/version/MultiHPLT}} 
% 16,723,245,979

A similar pivoting is also done for DocHPLT and we  create the additional bitexts through sentences aligned to the same English document. However, anchoring sentences in aligned documents creates a different subset across the pivoted language pairs.

\section{Dataset construction}
\label{sec:construction}

\begin{figure}[t]
    \centering
    \includegraphics[width=\linewidth]{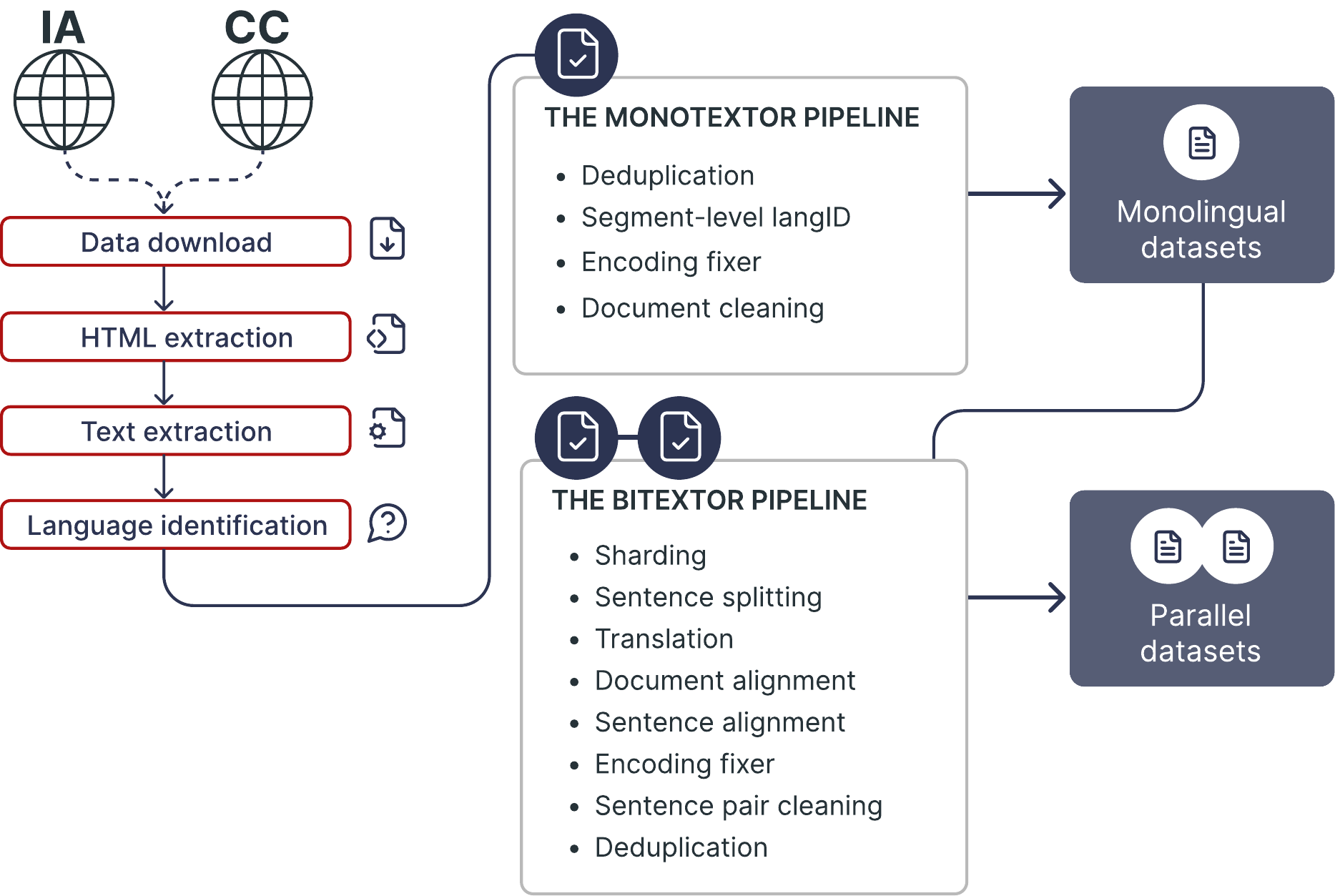}
    \caption{Overview of the data acquisition and processing pipeline for \HPLT{}.}
    \label{fig:pipeline}
\end{figure}

In the following section, we explain the dataset construction pipeline for \HPLT{}. We first extract text from web crawls via HTML (\Cref{sec:text_extraction}), deduplicate and clean this monolingual text (\Cref{sec:monolingual_proc}), and finally extract and process the parallel data (\Cref{sec:parallel_extraction}). \Cref{fig:pipeline} provides a high-level overview of the pipeline.

% The construction of our datasets starts with finding and extracting texts in roughly 200 natural languages from web crawls, this step is explained in section~\ref{sec:text_extraction}. These texts are then processed by two different pipelines to build the monolingual (section~\ref{sec:monolingual_proc}) and bilingual (section~\ref{sec:bilingual_proc}) datasets.

% section writing: Nikolay (with helpers)
\subsection{Text extraction from web crawls}
\label{sec:text_extraction}

\paragraph{Sources} In total, we ingest 4.5 PB  of web crawl data to build \HPLT{}. 3.7 PB is sourced from \ac{ia} from crawls conducted mostly between 2012 and 2020, with the remaining 0.8 PB coming from \ac{cc}.
%, a popular source of data for training language models. 
We use CC crawls conducted mostly between 2014 and 2022. A detailed description of the crawls we use is in \Cref{sec:appendix_crawlsources}.

% \subsubsection{Sources of web crawls}
% % : briefly describe IA, CC; Appendix - a list of crawls and sizes
% Our main source of web crawls is Internet Archive (IA)
% \footnote{\url{https://archive.org/}} providing roughly 3.7 PB of data crawled mostly over the years 2012-2020. For comparison, we additionally employ Common Crawl
% %\footnote{\url{https://commoncrawl.org/}}
% which is a popular source of crawls for training language models, it contributes 0.8 PB of crawls mostly from 2014-2022. Thus, the total size of our source web collections is $\approx4.5$ PB. See appendix~\ref{sec:appendix_crawlsources} for a detailed description of the employed crawls.

% Though the CC crawls are less than 20\% of the total size of all crawls, they source about 60\% of text (as measured in either characters or documents) form the final version of our dataset. This is probably due to their focus on the textual content, while IA crawls seem to equally care about multimedia content resulting in 4-8x lower yields in general. However, for some languages including Chinese, Persian and a few smaller languages, IA still provides much more texts than CC. Appendix~\ref{sec:appendix_yields} presents a detailed study of contributions of different sources of crawls. 

\paragraph{Extracting HTML} Both \ac{ia} and \ac{cc} crawls are provided in the Web ARChive (WARC) format\footnote{\url{https://www.iso.org/standard/68004.html}} which stores HTTP requests and responses between a web crawler and web servers. We use the \texttt{warc2text} tool\footnote{\url{https://github.com/bitextor/warc2text}} to extract HTML and related metadata from these WARC files. It selects relevant WARC records containing HTML pages, removes documents from a list of known trash websites,\footnote{Mostly containing auto-generated lists of phone numbers, addresses, etc.: \url{https://github.com/paracrawl/cirrus-scripts/blob/master/url-filter-list.annotated}} and finally saves the results in the ZSTD-compressed JSONL format. The extracted metadata includes document URLs, paths to the original WARC files and record positions inside, content types and timestamps. Additionally, WARC records with URLs ending with "\texttt{robots.txt}" are stored for later use in filtering. 

% \subsubsection{Extraction of HTMLs from web crawls}
% Both IA and CC crawls are provided in the Web ARChive (WARC)\footnote{\url{https://www.iso.org/standard/68004.html}} format, which is consumed by our data processing software. WARC files store HTTP requests from a web crawler to web servers and their responses. We start with extracting responses containing HTML pages and dumping these pages and the related metadata to disk for further processing.\footnote{Separating this processing step from the following processing is required because it needs a huge amount of disk space but not much compute power, thus, should be executed on the quite weak compute nodes directly attached to the storage. The extracted HTMLs are 5x smaller compared to the original WARC files, this allows transferring them to a cluster and running there further processing that requires significantly more compute.}

% Extraction of HTMLs from WARCs is performed by the \texttt{warc2text} tool.\footnote{\url{https://github.com/bitextor/warc2text}} Specifically, it selects relevant WARC records containing HTML pages, removes documents matching the URL block list, and finally dumps the results to disk in the ZSTD-compressed JSONLines\footnote{\url{https://jsonlines.org/}} format. The extracted metadata includes document URLs, paths to the original WARC files and record positions inside, content types and timestamps. Additionally, WARC records with URLs ending with "\texttt{robots.txt}" are dumped and later used for filtering. 

\paragraph{Extracting text} This stage of the pipeline extracts the main textual content from HTML pages and groups it into language-specific subsets. It first parses the HTML pages into a tree representation. Next, it removes likely machine-translated texts by searching for indicative HTML tags and attributes. It then removes boilerplate (i.e. parts of a web page that do not contribute to its main content) using Trafilatura 1.8.0 \citep{barbaresi-2021-trafilatura}. Following hyperparameter experimentation, we set \texttt{include\_comments=False}, \texttt{include\_tables=False}, \texttt{no\_fallback=False} and \texttt{MIN\_EXTRACTED\_SIZE=0}, with all other hyperparameters set to their defaults. We chose not to use fallback to Trafilatura's simple extraction baseline since it leaves most boilerplate intact, and we preferred sacrificing some documents but avoiding extra boilerplate in \HPLT{}. Finally, we predict the language of the text using a modified version of the OpenLID model\footnote{\url{https://data.statmt.org/lid/lid193_merged_arabics.bin}} \citep{burchell-etal-2023-open,burchell2024}, where the Arabic dialects are combined under one macrolanguage label and the model training data has undergone improved pre-processing. These changes are intended to improve classification reliability. After text extraction, the dataset size reduces to 62 TB, 15 times smaller than the HTML data and 75 times smaller than the original web crawls. 

\subsection{Monolingual text processing}
\label{sec:monolingual_proc}
% Jaume
Following text extraction, we proceed to monolingual text cleaning in which we apply various criteria to select the cleanest documents. With the exception of fixing encoding, we do not alter the text in this process.

We first discard all documents for which the predicted probability of the language label is $<0.5$. We then perform crawl-level deduplication with a MinHash index \cite{minhash}, using 240 hashes and a Jaccard similarity threshold of 0.8. We keep one document from each computed disjoint-set \cite{unionfind}, thus removing near-duplicates within each crawl. 

To respect \texttt{robots.txt}\footnote{\url{https://www.robotstxt.org/}} rules specified by each domain, we use the extracted \texttt{robots.txt} records to identify patterns disallowing the crawlers we use.\footnote{\texttt{*}, \texttt{CCBot}, \texttt{ia-archiver}} We use the \texttt{fst}\footnote{\url{https://burntsushi.net/transducers/}} tool to create a compressed index of URLs to exclude and use it to remove documents originating from these URLs.

We then use a range of heuristics to discard low-quality documents. We calculate a document quality score using Web Docs Scorer (WDS),\footnote{\url{https://github.com/pablop16n/web-docs-scorer/}} discarding documents with a score $<5$. We remove any documents where the length of the document is $<500$ characters, or where the average number of words per segment is $<5$ ($<10$ characters for Japanese, Chinese or Korean). We also filter documents where the URL is in the UT1 adult list.\footnote{\url{https://dsi.ut-capitole.fr/blacklists}} 

Finally, we enrich documents with additional metadata. We add a unique identifier for the document hash derived from the WARC file name, the URL and the timestamp. We also carry out segment-level \ac{lid} using the Rust port\footnote{\url{https://github.com/ZJaume/heliport}} of HeLI-OTS \cite{jauhiainen-etal-2022-heli}, trained on the OpenLID dataset. Finally, we add the Unicode character offsets of any personally identifiable information found by the PII tool.\footnote{\url{https://github.com/mmanteli/multilingual-PII-tool}}

Although the \ac{cc} crawls are less than 20\% of the input data, they are the source of about 60\% of the final text. This is likely because \ac{cc} focuses on textual content whereas \ac{ia} includes much multimedia content, resulting in 4-8x lower yields in general. However, for some languages (e.g. Chinese, Persian, and a few smaller languages), \ac{ia} provides more texts than \ac{cc}. \Cref{sec:appendix_yields} presents a detailed study of the contributions of different source crawls to the final dataset.

\subsection{Parallel data extraction}
\label{sec:parallel_extraction}
% Ona and Jaume
%REMEMBER: In contrast to the previous version, where the bilingual texts were extracted from the WARC files and went through a different preprocessing pipeline, the new version departs from the clean monolingual documents.

Our parallel data extraction pipeline is adapted from Bitextor.\footnote{\url{https://github.com/bitextor/bitextor}} We make the following changes to increase the quality of the final dataset:
\begin{compactitem}
    \item Input data comes from cleaned monolingual \HPLT{} rather than WARCs.
    \item We use Loomchild, a SRX-based sentence splitter \cite{srx}, to cover more languages.
    \item During sentence splitting, paragraph and sentence identifiers are added as persistent metadata through the pipeline.
    \item Minimal length rule and fluency filtering in bicleaner-hardrules are disabled as this duplicates other processing steps.
    \item Bicleaner AI \citep{zaragoza-bernabeu-etal-2022-bicleaner} uses a multilingual model able to handle unseen language pairs during training.
    \item Document-level output from the document-matching step is collected to allow the creation of document-level parallel data. 
\end{compactitem}

To avoid the possible introduction of new bugs in the pipeline, given that many of the steps in it are made for 2-letter language codes, we convert 3-letter language codes to 2-letter before processing.

\section{Data analysis}
\label{sec:analysis}
In this section, we present our analysis of the \HPLT{} data based on indirect quality indicators, manual inspection, and register labels.

\subsection{Indirect quality indicators}
\label{sec:analysis_quality_indicators}
% c. 1 page
%based on statistics
% Gema and Marta. Nikolay.
We consider two types of indirect quality indicators: descriptive statistics and website domains.

\paragraph{Descriptive statistics} We calculate descriptive statistics for \HPLT{} using the HPLT Analytics tool.\footnote{\url{https://github.com/hplt-project/data-analytics-tool}} We compare to the cleaned HPLT~v1.2 dataset (our previous release).

% \paragraph{Analytics} First, we use the HPLT Analytics tool\footnote{\url{https://github.com/hplt-project/data-analytics-tool}} which shows corpora details, volumes, languages, lengths, and other quality indicators. We apply it to our cleaned dataset as well as to HPLT-clean v1.2 for comparison. 

For the monolingual data, there are far more unique segments (22.2\% of HPLT~v1.2 vs. 40.9\% of \HPLT{}) but far fewer documents longer than 25 segments (90.8\% vs. 23.2\%). Similarly, the proportion of short segments is reduced (39.6\% vs. 13.3\%). These changes can be attributed to the use of Trafilatura and WDS. More segments match the document language (58.6\% vs. 81.5\%), driven by improvements in \ac{lid} accuracy and a more aggressive document filtering strategy. Finally, we identify frequent $n$-grams and find that a substantial amount of textual boilerplate remains, particularly from Wikipedia and blogging platforms. 

% For monolingual data, the number of unique segments is significantly larger (22.2\%  (HPLT~v1.2) vs. 40.9\% (\HPLT{})), while the number of documents containing more than 25 segments is smaller (90.8\% vs. 23.2\%). Similarly, the proportion of short segments is reduced (39.6\% vs. 13.3\%). These changes can be attributed to the use of Trafilatura and \HPLT{}-Scorer. Additionally, the proportion of segments in the correct document language increases (58.6\% vs. 81.5\%), driven by improvements in language identification accuracy and a more aggressive document filtering strategy. Finally, we identify frequent $n$-grams, revealing that a substantial amount of textual boilerplate remains, particularly from Wikipedia and blogging platforms. Examples can be found in Appendix \ref{sec:appendix_ngrams}.

Regarding parallel data, we find that the number of source and target tokens per language pair is much higher in \HPLT{} (by 47\% and 49\%  on average) than in HPLT~v1.2. Furthermore, 80\% of the sentence pairs have a translation likelihood score from 0.8 to 1 (as computed by Bicleaner AI) which attests to their high quality. The frequent $n$-grams in the parallel datasets are similar among all languages: larger datasets tend to focus on hotels and legal notices, whereas the smaller datasets exhibit more variety and the frequent $n$-grams in these datasets reflect local content likely from news websites, e.g. political figures and place names. \Cref{sec:appendix_ngrams} contains further examples.
% \begin{comment}
% \textbf{Monolingual datasets} We extract several indicators from clean versions of the HPLT~v1.2 and \HPLT{} datasets using the \HPLT{}-tool\footnote{The link is removed to maintain anonymity} analysis tool. Main findings are:

% \emph{Unique segments:} increase from 22.2\% (HPLT~v1.2) to 49.9\% (\HPLT{}) on average. We find two possible reasons: 1) boilerplate removal by Trafilatura removes a lot of duplicate segments, and 2) document filtering by \HPLT{}-Scorer removes poor content, template-generated documents (i.e., real estate or hotel booking pages). 

% \emph{Documents containing more than 25 segments:} drop from 90.8\% (clean HPLT~v1.2) to 23.2\% (\HPLT{}) on average. From a closer inspection, we determine that this massive drop is linked to the usage of Trafilatura which produces shorter documents.

% \emph{Segments in the document language:} grow from 58.6\% (HPLT~v1.2) to  81.5\% (\HPLT{}) on average. This might be explained by the improved accuracy in language identification in the pipeline and the introduction of a much more aggressive document filtering strategy that strongly penalises documents with segments in the wrong language.

% \emph{Short segments:} drop from 39.6\% (HPLT~v1.2) version to 13.3\% (\HPLT{}) on average, as a result of boilerplate removal by Trafilatura and document filtering by \HPLT{}-Scorer, which penalises documents containing too many short segments.

% \end{comment}

%We also extract some statistics from metadata related to website domain names and Top-Level Domains (TLDs):

\paragraph{Domains} We explore the website domain names and geographic \acp{tld} present in the data in order to understand its origins better.

% Next, to gain insights into the origins of our data, we explore the dataset's domains. Our analysis focuses on two dimensions: website domain names and geographic \acp{tld}.

We find different patterns of website domain names in the corpora depending on the size of the language dataset. Languages with more data available contain a diverse range of website domain names in the monolingual data but more travel-related webpages in the parallel data. However, smaller language datasets tend to contain more Wikipedia and religious content in both the monolingual and parallel data. \Cref{sec:appendix_domains} contains further information about common domains.

% In the monolingual dataset, medium-to-large language datasets exhibit a diverse range of \textbf{website domain names} without a single dominating source. Religious content is predominant in small-sized datasets, particularly for African languages, as well as Wikipedia. In the bilingual dataset, small datasets similarly contain a high proportion of content from Wikipedia and religious websites. However, medium-to-large bilingual datasets show a clear predominance of hotel booking and travel-related webpages.
%, which is not observed in the monolingual datasets.

% More examples of our findings regarding the most common domains can be found in Appendix \ref{sec:appendix_domains}.

Whilst most of the \acp{tld} in our dataset are general purpose (e.g. \texttt{.com}, \texttt{.org}), we found that the most common geographic \acp{tld} in the monolingual language corpora were usually from the country with the most speakers. This gave us confidence in the reliability of the text. We found the proportion of geographic \acp{tld} from an indicative country was highest for those in Europe, whereas the datasets for many languages primarily spoken in Africa mostly consisted of general-purpose \acp{tld}. The parallel data exhibits more diversity in \acp{tld} than the monolingual data. For example, \texttt{.eu} is much more frequent, appearing in the top-10 TLDs of all mid-size and large parallel datasets of nearly all European languages. A more detailed discussion of our observations is in \Cref{sec:appendix_tlds}.

\subsection{Manual data inspection} 
\label{sec:manual_inspection}

%Gema and Marta. Nikolay.
% To get an idea about the quality of our data as perceived by humans, we manually inspected a random sample of documents from the cleaned monolingual datasets in 22 languages. 
To assess human-perceived quality, we manually inspected a random sample of documents from the cleaned monolingual datasets in 22 languages. Specifically, for each language spoken by the authors, we sampled 50 random documents extracted from each of the four groups of crawls: the older \ac{cc}/\ac{ia} crawls from 2012--2014, and the newer \ac{cc}/\ac{ia} crawls from 2017--2020. The main goal of this stratification was to compare the quality of texts we get depending on the crawl source and age, and select the most promising crawls for the next release of our datasets. 

We asked participants to annotate any documents which look like pornographic content, look unnatural, and/or are not in the target language. \Cref{sec:appendix_r2inspection} describes the inspection procedure and the results. Overall, for most languages, both the proportions of pornographic content and texts not in the target language are around 0--3\%, with no significant difference between groups of crawls. Asturian, Scottish Gaelic and Norwegian Nynorsk are notable exceptions, with 31, 11 and 7 percent of texts not in the target language respectively. The proportion of unnatural texts is around 10\% on average and up to 30\% for some languages, leaving space for improvements. We also observe that the probability of getting an unnatural text from the newer CC crawls is roughly half of that of the other three inspected groups of crawls. This is probably related to the introduction of harmonic centrality ranking for domain prioritization in the CC crawler queue since 2017 \citep{nagel2023common}, which is stated to be more efficient in avoiding spam compared to the previously used techniques. 

\subsection{Register labels}
% UTU
As noted in \Cref{sec:analysis_quality_indicators}, web crawls cover a vast range of different kinds of documents from various sources. We use  automatic register (or genre) classification to create metadata about this variation, allowing users to make informed decisions when sampling from the data.

% Automatic register (or genre) classification provides a valuable tool for creating metadata about this variation and the origins of the documents, allowing one to make informed decisions when sampling from the data.
%how language is used across different communicative contexts in web text and allows for \textcolor{red}{(end user application, why did we do this)}. 
%Registers, as described by \citet{Biber_1988}, refer to distinct forms of language use associated with specific situations of use and functionally related linguistic features. Recent advances in multilingual register classification have achieved strong, near-human performance using deep learning \cite{kuzman-ljubesic-automatic-genre-identification,Skantsi_Laippala_2023,henriksson2024automaticregisteridentificationopen}. 

%We use the multilingual register classifier described in \citet{henriksson2024automaticregisteridentificationopen} to register label the entire monolingual \HPLT{} dataset. The classifier, based on XLM-RoBERTa Large \cite{conneau-etal-2020-unsupervised}, has been fine-tuned on the Multilingual CORE corpora, a manually register annotated multilingual web corpus similarly introduced by \citet{henriksson2024automaticregisteridentificationopen} and covering altogether 16 languages. The classifier uses a hierarchical taxonomy with 25 register classes organized into 9 main categories listed in Table \ref{tab:registers}.
%such as Narrative, Informational description, Opinion, and Lyrical.
We use the multilingual register classifier described in \citet{henriksson2024automaticregisteridentificationopen} to label the entire monolingual \HPLT{}. This classifier covers 16 languages and is based on an XLM-R Large model \citep{conneau-etal-2020-unsupervised}, fine-tuned on a multilingual web corpus manually annotated with register information. The classifier employs a hierarchical taxonomy with 25 register classes organized into 9 main categories (listed in \Cref{tab:registers}). This label scheme is specifically designed for the linguistic characteristics of web texts and is therefore well-suited for our dataset.

The system achieves a mean micro F1 score of 77\% on the 5 languages used during fine-tuning. It also demonstrates good performance for 11 unseen languages, with a mean micro F1 score of 66\%. These results allow us to extend register labelling to a broad range of languages, though we limit predictions to languages within the 100 languages covered by XLM-RoBERTa.

We provide the classification certainty as well as the label, so that the threshold can be optimized by use case. \Cref{tab:registers} presents the distribution for register labels in our English data for a classification threshold of 0.4. Further work could use our derived labels to improve dataset quality, by e.g. filtering out \ac{mt} content.

%The classification system achieves a mean micro F1 score of 77\% (SD = 3.16, n = 5) on the languages used in fine-tuning.
%The model has also demonstrated good register classification performance for languages not seen during fine-tuning, with an average micro F1 of 66\% (SD = 8.29, n = 11). These results allow us to extend register labelling to a broad range of languages despite not being covered by the training data of the classifier. However, we limit our predictions to languages within the coverage of XLM-RoBERTa, as the model's performance on languages outside this scope has not been evaluated.
%The classification process took altogether 36,712 GPU hours, specifically 4,589 hours on 8-GPU nodes on \LUMI{} supercomputer. 

\begin{table}[t]
    \centering\small
    \begin{tabular}{lr}
    \toprule
        \multicolumn{1}{c}{\textbf{Register}} & \multicolumn{1}{c}{\textbf{Percentage}} \\
    \midrule
         How-to \/ Interactive (HI) & 1.8 \% \\
        Interactive Discussion (ID) & 6.5 \% \\
        Informative Description (IN) & 27.1 \%\\
        Informative Persuasion (IP) & 10.8 \%\\
        Lyrical (LY) & 0.5 \% \\ 
        Machine Translated (MT) & 3.3 \% \\ 
        Narrative (NA) & 18.1 \% \\
        Opinion (OP) & 5.4 \% \\ 
        Spoken (SP) & 0.2 \% \\ 
        Multiple labels & 23.6 \% \\
        No label & 2.5 \% \\ 
    \bottomrule
    \end{tabular}
    \caption{Register label distribution in \HPLT{} English dataset for classification threshold 0.4. See \citet{henriksson2024automaticregisteridentificationopen} for the full scheme and explanation of the contents of the classes.}
    \label{tab:registers}
\end{table}

\section{Empirical evaluation}
\label{sec:evaluation}

In this section, we describe our empirical evaluation of the quality of the \HPLT{} monolingual corpora. We conduct this evaluation by employing the datasets as training material for several \ac{nlp} models.

\subsection{Basic linguistic tasks and MLMs}
\label{sec:eval_mlm}
% Mariia Fedorova [UiO]
We train \acp{mlm} on 52 different languages from the \HPLT{} datasets, choosing those with available benchmarks.\footnote{\url{https://github.com/hplt-project/HPLT-WP4}} We use LTG-BERT \citep{samuel-etal-2023-trained} to allow comparison with HPLT~v1.2. We give full details about LTG-BERT in \Cref{sec:appendix_LTG-BERT}. 

We evaluate the trained \acp{mlm} on part-of-speech tagging, lemmatization and dependency parsing using the Universal Dependencies (UD) treebanks \citep{de-marneffe-etal-2021-universal}, as well as named entity recognition (NER) using WikiAnn datasets \citep{pan-etal-2017-cross}. We compare to mBERT \citep{devlin-etal-2019-bert} and XLM-R \citep{conneau-etal-2020-unsupervised} models as multilingual baselines, and to HPLT~v1.2 BERT models\footnote{\url{https://hf.co/collections/HPLT/hplt-bert-models-6625a8f3e0f8ed1c9a4fa96d}} as monolingual baselines. The performance is measured using the official CoNLL 2018 evaluation code \citep{zeman-etal-2018-conll} for the UD tasks, and \texttt{seqeval} \cite{seqeval} balanced F1 score for the NER task.

% We train \acp{mlm} on 52 different languages from the \HPLT{} datasets, choosing those with available benchmarks. For comparability, we employed LTG-BERT \cite{samuel-etal-2023-trained}, the architecture used to evaluate HPLT~v1.2 datasets.\footnote{\url{https://hplt-project.org/HPLT_D4_1___First_language_models_trained.pdf}, section 3.3} See more details about LTG-BERT in Appendix~\ref{sec:appendix_LTG-BERT}.

% The trained MLMs were evaluated on the tasks of part-of-speech tagging, lemmatization and dependency parsing using the Universal Dependencies (UD) treebanks
% %\footnote{\url{https://universaldependencies.org/}}
% \cite{de-marneffe-etal-2021-universal}; and named entities recognition (NER) using WikiAnn datasets
% %\footnote{\url{https://huggingface.co/datasets/unimelb-nlp/wikiann}}
% \cite{pan-etal-2017-cross}. We compare to mBERT
% %\footnote{\url{https://huggingface.co/google-bert/bert-base-multilingual-cased}} 
% \cite{devlin-etal-2019-bert} and XLM-R
% %\footnote{\url{https://huggingface.co/FacebookAI/xlm-roberta-base}}
% \cite{conneau-etal-2020-unsupervised} models as multilingual baselines, and to HPLT~v1.2 BERT models\footnote{\url{https://huggingface.co/collections/HPLT/hplt-bert-models-6625a8f3e0f8ed1c9a4fa96d}} as monolingual baselines. The performance is measured using the official CoNLL 2018 evaluation code \cite{zeman-etal-2018-conll} for the UD tasks, and \texttt{seqeval} \cite{seqeval} balanced F1 score for the NER task.

%Here must be the analysis. 

\begin{figure}[tb]
\small\centering
\includegraphics[width=0.95\columnwidth]{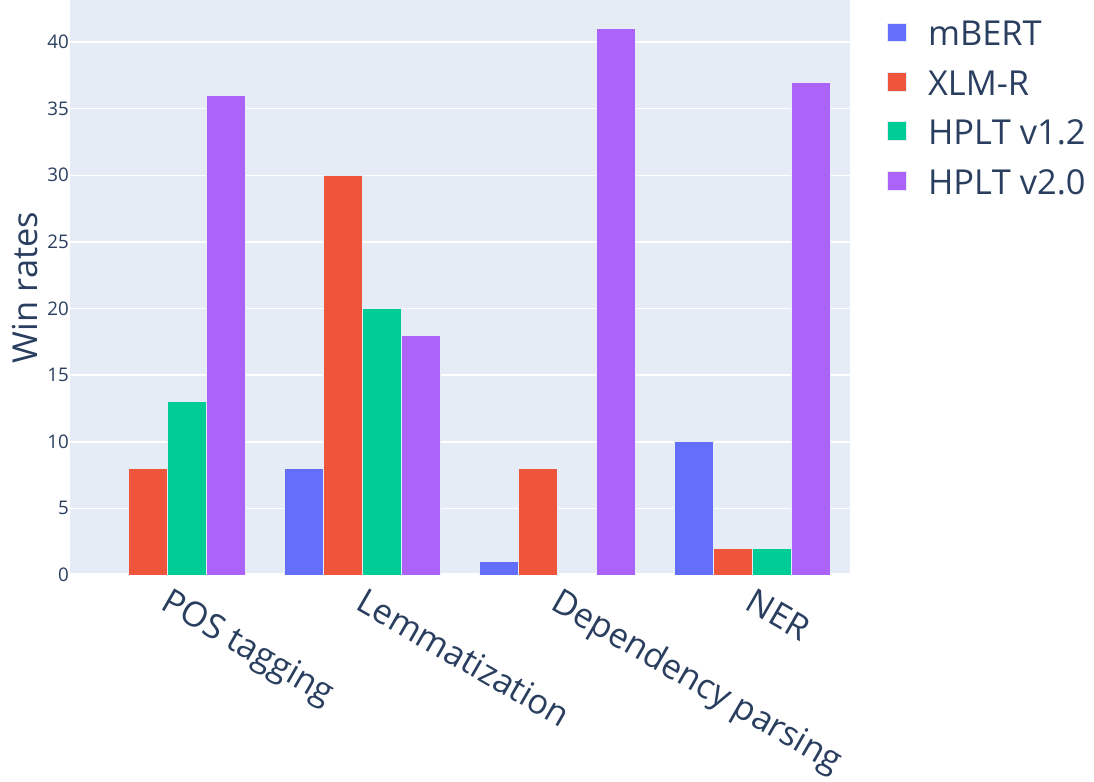}
    \caption{Win rates for \acp{mlm} at part-of-speech tagging, lemmatisation, dependency parsing, and named entity recognition. 
    %Zero bars mean the model never was the best for the task.
    }
    \label{fig:masked}
\end{figure}

\Cref{fig:masked} shows the win rates achieved by the models for the four tasks (`win rate' here is the number of languages on which a given model outperforms other models). Models trained on the \HPLT{} datasets show a considerably higher win rate compared to the baselines in all the tasks except lemmatization, where XLM-R and HPLT~v1.2 yield competitive results. However, we note that the difference between XLM-R, HPLT~v1.2 and \HPLT{} on the lemmatization task is less than 1\% of accuracy, meaning that no model significantly outperforms any other. Detailed scores by language and task are to be found in \Cref{tab:mlm}. We make the \HPLT{} BERT models with intermediate checkpoints publicly available.\footnote{\url{https://hf.co/collections/HPLT/hplt-20-bert-models-67ba52ae96b1fb8aae673493}}

\subsection{NLU tasks and large generative LMs}
\label{subsec:nlu_evaluation_ablations}
% Sampo [UTU]? Vlad [UiO]? 
% use sentences from here (internal evaluation): https://hplt-project.org/hplt-v2-dataset-performance 

Pretraining generative \acp{lm} and evaluating their downstream performance on advanced \ac{nlu} tasks is an established way to measure and compare training data quality \cite{thepile, refinedweb, longpre-etal-2024-pretrainers}. Following \citet{fineweb}, we compare various large web-crawled pretraining corpora using this method for one high-resource and one low-resource language: English and Norwegian. We train 1.7B decoder-only \acp{lm} using 100B/30B tokens sampled from the English/Norwegian parts of our \HPLT{} dataset respectively. We compare our English and Norwegian models with models trained on same-sized samples of HPLT~v1.2 \cite{de-gibert-etal-2024-new} and FineWeb \cite{fineweb}, and additionally compare our Norwegian models with FineWeb-2 \cite{penedo2024fineweb-2}, CulturaX \cite{nguyen-etal-2024-culturax}, and mC4 \cite{xue-etal-2021-mt5}.
%and compare them with models trained on same-sized samples of HPLT~v1.2 \cite{de-gibert-etal-2024-new}, FineWeb \cite{fineweb}, FineWeb-2 \cite{penedo2024fineweb-2}, CulturaX \cite{nguyen-etal-2024-culturax}, and mC4 \cite{xue-etal-2021-mt5}. 
We replicate the design by \citet{fineweb} and train the models with a fixed pretraining setup except for the pretraining corpus (English: four corpora; Norwegian: five corpora). We provide full details on pretraining and evaluation in \Cref{sec:appendix_LLMtrainingEval} and describe our key results below.

%Following \citet{fineweb}, we train 1.71 billion parameter decoder-only language models using 100 billion tokens sampled from the English part of our \HPLT{} dataset, and compare them with models trained on same-sized samples of similar large web-crawled datasets, HPLT~v1.2 \cite{de-gibert-etal-2024-new} and FineWeb \cite{fineweb}. We replicate the setup used by \citet{fineweb}, with the tokenizer, model architecture, and hyper-parameters matching those used in the original FineWeb experiments. Models are trained identically, with the only exception being the data: two models are trained on the \HPLT{} data (cleaned and deduplicated separately), one on the FineWeb data, and one on HPLTv1.2. The full model training setup can be found in Appendix \ref{sec:appendix_LLMtrainingEval}. We use \texttt{Megatron-LM}\footnote{\url{https://github.com/NVIDIA/Megatron-LM}} training framework by NVIDIA as opposed to the HuggingFace's \texttt{nanotron}\footnote{\url{https://github.com/huggingface/nanotron}} framework used by \citet{fineweb}. Evaluation is done using the HuggingFace \texttt{LightEval} tool \cite{lighteval}, specifically with benchmarks listed in Appendix \ref{sec:appendix_LLMtrainingEval}.

\begin{figure}[t]
    \centering
    \includegraphics[width=0.95\linewidth]{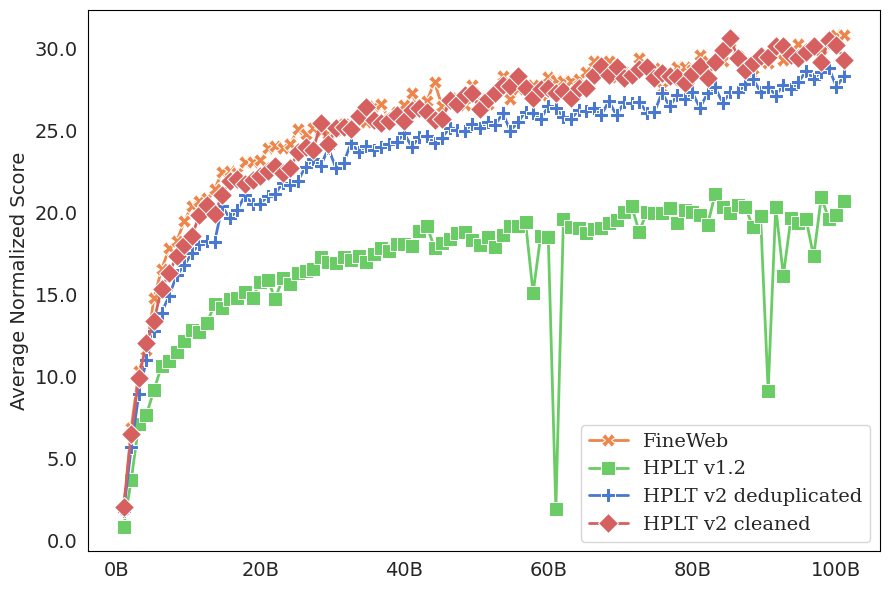}
    \caption{Performance comparison of the trained generative \acp{lm} on English.}
    \label{fig:LLM-results-english}
\end{figure}

\paragraph{English} Average results over the English benchmarks are presented in Figure \ref{fig:LLM-results-english}. Our models trained on the cleaned \HPLT{} datasets reach similar performance to the models trained on FineWeb data in downstream tasks, and considerably outperform the models trained on HPLT~v1.2 and on the deduplicated subset of \HPLT{}. 
%Specifically, the model trained on the cleaned subset of \HPLT{} is on par with the model trained on FineWeb data in downstream tasks, and shows improvement over the model trained on the deduplicated subset of \HPLT{}. 
This implies that our cleaning approach has successfully improved the data quality with respect to these benchmarks.

\begin{figure}[tb]
    \centering
    \includegraphics[width=0.95\linewidth]{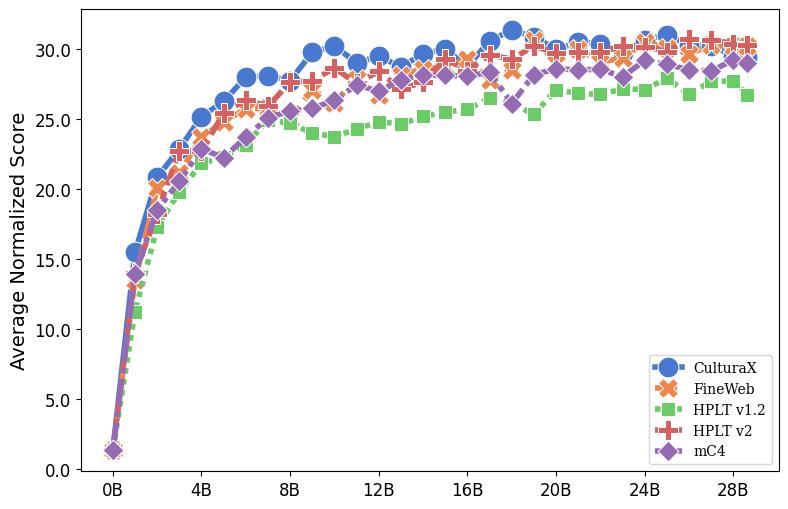}
    \caption{Performance comparison of the trained generative \acp{lm} on Norwegian.}
    \label{fig:LLM-results-norwegian}
\end{figure}

\paragraph{Norwegian} Average normalized scores over the Norwegian tasks are shown in Figure \ref{fig:LLM-results-norwegian}. We observe that the Norwegian models trained on FineWeb, CulturaX, and mC4 perform on par with \HPLT{}  and outperform those trained on HPLT~v1.2. Performance gains start to level off after 16B tokens, with the FineWeb and \HPLT{} scores being more stable during pretraining. This suggests that CulturaX, FineWeb, and \HPLT{} are more effective corpora for Norwegian, and their mixtures potentially provide further benefits.

\subsection{Machine translation tasks}
% Dušan and Jindra 
%results: https://docs.google.com/spreadsheets/d/1MaNKz7WrIXl-s81DzKlDVmg-OuPyr4hI97SaakHCZRg/edit?usp=sharing

We extrinsically evaluate the quality of the \HPLT{} parallel data by measuring the performance of MT models trained on it. This section considers two settings:
\begin{compactitem}
    \item We compare \HPLT{} to its predecessor, HPLT~v1.2, where we measure their performance in 10 overlapping languages. 
    \item We assess \HPLT{} as a complementary dataset to existing MT resources. We compare models in three data conditions: 1) solely on \HPLT{} parallel data; 2) on the data from Tatoeba Challenge (which includes most of the OPUS collection for each direction \citealp{tiedemann-2012-parallel,tiedemann-2020-tatoeba}); and 3) on a combination of the two.
\end{compactitem}

We carry out training for each individual language pair (to and from English) using the Transformer-base architecture \citep{transformer} and the Marian NMT toolkit \citep{junczys-dowmunt-etal-2018-marian}. Data processing and training are streamlined with OpusPocus\footnote{\url{https://github.com/hplt-project/OpusPocus}} following the configuration of \citet{arefyev-etal-2024-hplts}. 
We evaluate all models on the FLORES-200 benchmark \citep{nllb2022} using BLEU \citep{papineni-etal-2002-bleu}, chrF++ \citep{popovic-2017-chrf}, and COMET-22-DA \citep{rei-etal-2022-comet}. 
We use sacrebleu's implementation of BLEU\footnote{\scriptsize \texttt{nrefs:1|case:mixed|eff:no|smooth:exp|version:2.5.1},\\and where applicable, \texttt{tok:ja-mecab}, \texttt{tok:ko-mecab}, or \texttt{tok:13a}} and chrF++\footnote{\scriptsize \texttt{nrefs:1|case:mixed|eff:yes|nc:6|nw:0|space:no|version:2.5.1}} with signatures footnoted \citep{post-2018-call}. Full results for each metric and each direction are given in \Cref{tab:para_mt_eval_from_en} and \Cref{tab:para_mt_eval_into_en} in \Cref{sec:app-full-MT-results}.

\begin{table}[t]
\centering\small
\adjustbox{max width=0.99\linewidth}{
\begin{tabular}{lcccc}
\toprule
   & \multicolumn{2}{c}{xx-en} & \multicolumn{2}{c}{en-xx} \\
\cmidrule(lr){2-3}\cmidrule(lr){4-5}
   & BLEU & COMET & BLEU & COMET \\
\midrule
HPLT~v1.2  & 28.5 & 0.7943 & 24.4 & 0.7623 \\
\HPLT{} & \textbf{32.7} & \textbf{0.8343} & \textbf{27.9} & \textbf{0.8137} \\
\midrule
\HPLT{} & 28.7 & 0.8144 & 23.4 & 0.7941 \\
OPUS & 29.6 & 0.8142 & 23.4 & 0.8074 \\
\HPLT{}+OPUS & \textbf{30.5} & \textbf{0.8237} & \textbf{24.2} & \textbf{0.8083} \\
\bottomrule
\end{tabular}
}
\caption{Average scores for the HPLT v1.2 and v2 comparison (top) and \HPLT{} as a complimentary resource to OPUS (bottom). Only numbers that are available for all models in a comparison are averaged.}
\label{tab:average_numbers_MT}
\end{table}

% \Cref{fig:mt_eval_avg_bleu} summarises the average BLEU score for different settings, with detailed results in \Cref{tab:para_mt_eval_from_en,tab:para_mt_eval_into_en} in the appendix.
% Table~\ref{tab:para_mt_eval_from_en} and Table~\ref{tab:para_mt_eval_into_en} in Appendix~\ref{sec:appendix_para_mt_evaluation} show the detailed results. Figure~\ref{fig:mt_eval_avg_bleu} summarizes the average BLEU score for different settings. 

We report average BLEU and COMET scores in \Cref{tab:average_numbers_MT}. It is worth noting that we only average numbers for translation directions that are covered by all models in a comparison. The top half features a clear advantage of \HPLT{} over HPLT~v1.2, reflecting the effect of our improved data extraction pipeline discussed in \Cref{sec:construction}. Results in the bottom of \Cref{tab:average_numbers_MT} shows that \HPLT{} and OPUS perform comparably. However, combining the two datasets leads to improvements in both BLEU and COMET. This indicates that \HPLT{} contains non-overlapping content relative to existing OPUS corpora, making it a valuable complementary resource for \ac{mt}.

\section{Conclusions and future work}
% Laurie, Ona 
This paper introduced the \HPLT{} dataset, a large-scale multilingual collection of openly-available monolingual and parallel web-crawled data. Building on our previous HPLT~v1 effort, we focused on improving the quality of data available for a wider range of languages, and we make our data processing pipelines publicly available for easy reuse. We presented extensive data analysis as well as intrinsic and extrinsic evaluation, demonstrating the value of \HPLT{} for various \ac{nlp} tasks. 

Further work will focus on expanding language coverage and data quality, particularly for under-served languages. We also plan to conduct more experiments on the document-level aligned parallel corpus. 

% We introduce the \HPLT{} dataset, a large-scale multilingual collection of monolingual and parallel web-crawled data, released under the permissive \ac{cc0} license. Using \ac{cc} and \ac{ia} as sources, we provide insights into which crawls are more effective for obtaining clean data. Our data processing and cleaning pipelines are publicly available for easy reuse. Additionally, we present extensive data analysis and both intrinsic and extrinsic evaluations, demonstrating the dataset's value for various \ac{nlp} tasks. \HPLT{} represents a significant milestone for advancing the field of open \ac{llm} training.

% Our next steps focus on expanding coverage for underrepresented languages. However, we face a challenge: adding new languages to our dataset requires the use of language tools for our pipeline (such as language identification, translation models, and Bicleaner models), which creates a cyclical dependency. Addressing this dependency is crucial for breaking the cycle and achieving broader language inclusion.
% We also plan to implement a more aggressive Wikipedia boilerplate removal, as our analysis in Section \ref{sec:analysis_quality_indicators} reveals that our data still contains a substantial amount of this content.
% Additionally, we plan to release a document-level aligned parallel corpus, as we have all the necessary metadata to support it.

%We hope this work not only advances the development of multilingual NLP but also inspires future efforts to create more inclusive, high-quality datasets that better represent the world's rich linguistic diversity.

\section*{Limitations} % add ideas

% Corpus is heavily English-biased, because it reflects the internet
% Evaluation only necessarily has partial coverage of languages, we cannot look into all.
Like many large-scale corpora, the majority of the data in \HPLT{} is in Indo-European languages, especially English, and the parallel data is English-centric. To an extent, this is a result of the dominance of these languages in the source web-crawl data. In addition, the evaluation in the paper only covers a subset of the languages in \HPLT{} due to a lack of resources for all languages present. We hope that the data we release in multiple under-served languages will be used to improve language technologies for more communities.

% Residual errors in language identification, cleaning and boilerplate removal
% Limited removal of artificially generated content (LLM and MT)
%% (Looking for a reference on rise of synthetic data in corpora but could
%% not find one)
Whilst we focus on improving the \HPLT{} data processing pipeline, there are still residual errors in the final dataset in \ac{lid}, boilerplate removal (particularly Wiki* boilerplate) and other cleaning steps. We make the code for our pipeline available to facilitate its evaluation and improvement. We note that there is only limited removal of machine-generated content in \HPLT{} (i.e., content generated by technologies like \ac{mt} and \acp{llm}), as detecting such content remains a difficult task \citep{yang-etal-2024-survey}. 

% From Maria, on Mattermost: as for limitations, data contamination, like I was testing BERTs for NER on a 2017 old dataset based on wikipedia and we know for sure that wikipedia could have been in the pretraining data (could, because we did not used all data if there were more of them than required for 31250 steps)
It is possible that some of the test data we use for evaluation is contained within \HPLT{} (for example, the Wikipedia-based test set for named entity recognition). Nevertheless, we believe that the results reported in \Cref{sec:evaluation} are still indicative of the quality of \HPLT{}, since the large-scale datasets we compare against are likely to have similar contamination issues.

% Chinese (and probably Korean) has issues with punctuation
During the evaluation, we discovered that the punctuation for Chinese languages (and probably Korean and Japanese) in \HPLT{} had been normalised incorrectly to its Latin equivalent, causing a drop in measured performance for languages in this script. We will fix this in the next iteration of the HPLT pipeline.

\section*{Ethical considerations} % add ideas
% crawling of copyrighted data
% scary malware, API keys
% Carbon footprint
We source our data from web crawls and since Internet text is largely unregulated, our final dataset may contain harmful content or amplify existing biases, despite the extensive filtering applied to mitigate these issues. One notable bias is the over-representation of religious content in smaller language corpora, which could lead to models trained on this data being biased towards this particular domain. 
%which could unintentionally introduce undesirable behaviour in models trained on this data.

Another pressing ethical consideration is the significant environmental impact of producing large-scale datasets. We mitigate this impact by making the data openly available in multiple formats, limiting the need to reproduce the processing pipeline. 

We report the estimated CPU and GPU cost in hours for our work to allow for more informed decision-making in future research efforts:
\begin{compactitem}
    %\item Data download:
    \item WARC to HTML extraction: 250K CPU
    \item HTML to text extraction: 1.7M CPU
    \item Monolingual data processing: 600K CPU
    \item Parallel data cleaning and deduplication: 1.8M CPU and 23K GPU
    % ONA: Taken from CSC projects
    % project 462000688: 1 000 000 CPU 22875 GPU
    % project 462000764: 806453 CPU 210 GPU
    \item Register labels classification: 36.7K GPU
    \item \ac{mlm} experiments: 1.8K CPU and 4K GPUs
    \item Generative \acp{lm} training and evaluation: 21.5K CPU and 43K GPU
    \item MT models training and evaluation: 20K GPU
\end{compactitem}
The total amount of hours spent would be roughly 4.4M CPU hours and 106K GPU hours. The most expensive task is the evaluation of our data through generative model training. We mitigate the environmental impact of our work by using one of the most eco-efficient HPC clusters in the world (LUMI) to carry out much of our computation. The LUMI supercomputer uses renewable, carbon-neutral energy.\footnote{\url{https://lumi-supercomputer.eu/sustainable-future/}}
%if it's not easy to calculate carbon footprint, saying "we use blah blah blah hours which entails significant resource consumption. We mitigate this by making our data available so that resource intensive analysis need not be repeated. We also use one of the most eco-efficient data centers in the world to carry out much of our computation." 

\section*{Acknowledgements}

We acknowledge Tsz Kin Lam, Otto Tarkka, and Maryam Teimouribadelehdareh for their annotation work; we thank Jelmer van der Linde for his help with Go.

This project has received funding from the European Union's Horizon Europe research and innovation programme under Grant agreement No 101070350 and from UK Research and Innovation (UKRI) under the UK government's Horizon Europe funding guarantee [grant number 10052546]. The contents of this publication are the sole responsibility of its authors and do not necessarily reflect the opinion of the European Union. 

The authors wish to thank CSC~-~IT Center for Science, Finland for computational resources and support. The project has also been supported by the Czech MEYS project No. CZ.02.01.01/00/23\_025/0008691 and Research Infrastructure project LM2023062. 

\bibliography{anthology,custom}

%%%%%% APPENDIX STARTS HERE %%%%%%%
\appendix

\onecolumn
\section{Comparison of multilingual collections}
\label{sec:multi_collections}
% Working in this spreadsheet https://docs.google.com/spreadsheets/d/1OrCObkZ4b06cdaMNDav-zOIlEP55Ed66R_TLPSNub1k/edit?usp=sharing
% If you want to edit, ask Ona for permission
\begin{table}[htbp]
    \centering
    \adjustbox{max width=\linewidth}{
    \begin{tabular}{lrrrrr}
    \toprule
    \textbf{Dataset} & \textbf{Size (TB)} & \textbf{Tokens (T)} & \textbf{Langs}  & \textbf{\% English} & \textbf{Source} \\
    \midrule
     \rowcolor{gray!20} 
     \multicolumn{6}{c}{English only} \\
    \midrule
    The Pile \cite{thepile} & 0.8 & 0.39 & 1 & 100 & various (c.f Section 2) \\
    C4.en \cite{c4, dodge2021documenting} & 0.3 & 0.16 & 1 & 100 & Common Crawl \\
    RefinedWeb \cite{refinedweb} & 2.8 & 0.6 & 1 & 100 & Common Crawl \\
    Dolma-Web \cite{soldaini-etal-2024-dolma} & - & 2.28 & 1 & 100 & Common Crawl \\
    FineWeb \cite{fineweb} & - & 15 & 1 & 100 & Common Crawl \\     
         \midrule
\rowcolor{gray!20} 
     \multicolumn{6}{c}{Multilingual} \\
    \midrule
OSCAR-23.01 \cite{oscar} & - & 1.1 & 153 & 48.43 & Common Crawl \\
    CC-100 \cite{conneau-etal-2020-unsupervised} & 2.39 & 0.3 & 100 & 18.84 & Common Crawl \\
mC4 \cite{xue-etal-2021-mt5} & - & 6.3 & 101 & 5.67 & Common Crawl \\
ROOTS \cite{roots} & 1.6 & 0.4 & 46 & 30.03 & \makecell[r]{BigScience Catalogue Data, \\Common Crawl, OSCAR} \\
Glot500-c \cite{glot500} & 0.6 & - & 511 & *2.16 & various (c.f. Appendix C) \\
Serengeti \cite{serengeti} & 0.042 & - & 517 & - & various (c.f. Appendix C) \\
CulturaX \cite{culturax} & 27 & 6.3 & 167 & 45.13 & OSCAR, mC4 \\
MADLAD-400-clean \cite{madlad400} & - & 2.6 & 419 & 50 & Common Crawl \\
MaLA \citep{ji2024emma}  & - & 0.074 & 939 & 4 & various (c.f. Section 2.1.4)  \\
monoHPLT v1.2-dedup \cite{de-gibert-etal-2024-new} & 11 & 5.6 & 75 & 41 & \makecell[r]{Common Crawl, \\ Internet Archive}\\
    \midrule
    \HPLT{} (monolingual, deduplicated) & 21  &  7.6 & 193  & 44 & \makecell[r]{Common Crawl, \\ Internet Archive} \\
    
    \bottomrule
    \end{tabular}}
    \caption{Comparison of selected massively multilingual collections of monolingual data listed in chronological order. We report size, token counts, language coverage, and the proportion of English content. - indicates that data is not available. * indicates that the English percentage was computed over sentence counts, instead of token counts.}
    \label{tab:datasets}
\end{table}

\newpage  % to get section title on same page as table
\section{Parallel and monolingual data statistics}
\label{sec:parallel_stats}
\begin{table*}[!h]
\centering\small
\setlength{\tabcolsep}{1.2ex}
\begin{tabular}{lrrrrrr}
\toprule
\textbf{} &\multicolumn{2}{c}{\textbf{Raw}} &\multicolumn{2}{c}{\textbf{Filtered}} &\multicolumn{2}{c}{\textbf{TMX}} \\
\cmidrule(lr){2-3}
\cmidrule(lr){4-5}
\cmidrule(lr){6-7}
\textbf{Language} &\textbf{Sentence Pairs} &\textbf{English Words} &\textbf{Sentence Pairs} &\textbf{English Words} &\textbf{Sentence Pairs} & \textbf{English Words} \\\midrule
sin\_Sinh &929,844 &15,647,062 &450,122 &8,248,007 &273,430 &5,932,234 \\
npi\_Deva &1,058,740 &18,514,145 &523,022 &10,176,931 &317,120 &7,145,363 \\
xho\_Latn &1,223,514 &16,524,728 &655,790 &9,339,359 &405,605 &5,998,358 \\
mal\_Mlym &1,686,113 &25,600,791 &795,653 &12,475,940 &547,168 &9,656,086 \\
nno\_Latn &2,358,129 &34,771,540 &1,175,108 &21,352,540 &563,791 &10,548,302 \\
mar\_Deva &2,067,311 &34,952,324 &952,116 &19,606,305 &656,962 &15,113,175 \\
guj\_Gujr &2,134,977 &38,906,708 &1,165,483 &23,631,881 &716,777 &16,564,683 \\
kan\_Knda &2,354,299 &37,451,816 &1,238,033 &21,344,021 &720,157 &13,965,655 \\
tel\_Telu &2,924,532 &46,227,504 &1,513,237 &25,963,464 &902,962 &17,487,796 \\
tam\_Taml &3,859,610 &55,779,718 &1,759,372 &28,369,233 &1,111,471 &20,718,487 \\
uzn\_Latn &2,791,412 &37,715,209 &1,571,871 &25,823,124 &1,159,869 &19,667,785 \\
urd\_Arab &3,866,815 &101,346,427 &2,200,602 &65,830,839 &1,399,893 &47,591,409 \\
eus\_Latn &5,907,808 &79,485,282 &2,526,198 &38,107,950 &1,491,873 &24,303,464 \\
epo\_Latn  & 5,664,237&91,081,114 &3,190,135 &60,141,119 &1,521,821 & 30,986,721 \\
mlt\_Latn &7,434,717 &114,046,030 &2,651,758 &50,044,197 &1,529,471 &32,243,598 \\
kaz\_Cyrl &3,827,170 &55,027,673 &2,628,328 &39,138,283 &1,943,935 &30,216,073 \\
swh\_Latn &10,125,330 &145,685,653 &3,680,151 &68,766,541 &1,985,899 &39,952,916 \\
ben\_Beng &6,376,109 &106,303,435 &3,920,955 &70,350,081 &2,328,136 &49,851,040 \\
isl\_Latn &11,929,153 &146,981,787 &6,624,589 &91,089,371 &2,694,541 &47,440,271 \\
gle\_Latn &7,685,880 &133,441,028 &4,421,130 &89,932,030 &2,697,582 &59,065,530 \\
glg\_Latn &8,680,808 &145,602,784 &5,166,276 &99,562,132 &2,783,727 &58,437,672 \\
bel\_Cyrl &11,493,046 &154,657,914 &6,092,481 &90,760,902 &3,140,958 &50,113,002 \\
azj\_Latn &8,506,772 &118,079,751 &4,765,278 &72,026,597 &3,188,231 &51,425,346 \\
pes\_Arab &9,434,306 &192,718,387 &5,391,049 &130,840,005 &3,448,296 &95,822,037 \\
cym\_Latn &9,390,284 &156,087,956 &6,348,606 &125,442,540 &3,867,402 &82,244,645 \\
afr\_Latn &15,901,372 &246,703,185 &7,452,216 &139,249,006 &3,987,340 &80,857,410 \\
mkd\_Cyrl &10,815,504 &185,651,668 &7,175,217 &131,839,062 &3,991,617 &78,629,993 \\
tha\_Thai &13,818,095 &102,830,296 &7,551,187 &52,171,172 &4,088,354 &34,155,503 \\
als\_Latn &11,171,352 &208,460,688 &6,943,910 &145,918,660 &4,166,536 &94,263,904 \\
bos\_Latn &12,480,871 &193,998,734 &7,527,232 &139,457,685 &4,559,328 &92,723,229 \\
srp\_Cyrl &17,605,882 &244,921,478 &9,618,806 &153,171,621 &5,291,686 &90,518,351 \\
zsm\_Latn &47,173,963 &558,911,698 &22,298,471 &301,446,773 &8,432,285 &147,009,838 \\
heb\_Hebr &34,004,891 &431,453,938 &21,600,460 &279,563,405 &8,686,089 &162,768,846 \\
est\_Latn &29,934,421 &362,843,780 &16,629,846 &223,025,816 &8,797,574 &133,824,400 \\
hin\_Deva &26,345,062 &500,967,390 &16,337,324 &372,581,817 &9,926,620 &263,709,932 \\
slv\_Latn &30,956,083 &449,919,060 &18,290,300 &301,699,622 &10,336,528 &188,709,019 \\
lvs\_Latn &39,599,210 &476,030,125 &24,504,355 &316,955,851 &11,294,618 &183,588,490 \\
lit\_Latn &47,035,968 &553,786,167 &27,879,310 &351,354,617 &12,881,354 &205,285,778 \\
cat\_Latn &40,922,098 &671,563,410 &26,451,844 &521,397,424 &13,080,859 &292,854,267 \\
hrv\_Latn &45,617,022 &627,929,707 &27,783,979 &420,953,899 &14,263,908 &250,103,294 \\
ara\_Arab &41,671,896 &759,192,353 &31,389,602 &618,611,620 &17,505,366 &424,460,900 \\
kor\_Hang &76,980,595 &865,790,290 &46,010,282 &530,544,997 &18,393,859 &294,720,264 \\
jpn\_Jpan &105,291,263 &575,207,953 &48,568,992 &178,462,644 &18,894,019 &80,778,230 \\
vie\_Latn &47,831,389 &1,077,677,445 &35,072,681 &831,280,934 &19,231,770 &502,683,444 \\
slk\_Latn &62,840,882 &798,323,342 &40,704,524 &566,060,898 &20,056,339 &332,818,300 \\
tur\_Latn &84,823,944 &1,054,338,198 &46,493,019 &656,237,839 &21,616,652 &402,325,110 \\
bul\_Cyrl &63,059,982 &939,666,726 &40,936,972 &670,141,259 &22,725,326 &420,768,808 \\
nob\_Latn &71,277,875 &969,526,777 &45,204,041 &700,446,147 &22,912,722 &395,447,564 \\
ukr\_Cyrl &68,111,464 &862,726,167 &46,141,511 &637,243,802 &25,125,019 &400,949,773 \\
fin\_Latn &98,138,078 &1,028,494,366 &59,836,942 &667,840,953 &29,067,875 &383,463,787 \\
%\midrule
%\textbf{Total} &1,277,120,078 &116,849,551,707 &759,810,366 &11,206,020,915 &380,710,720 &6,779,910,082 \\
%\textbf{Median} &11,332,199 &170,869,812 &6,784,250 &128,141,273 &3,927,371 &79,704,112 \\
\bottomrule
\end{tabular}
\caption{Statistics for the parallel portion of \HPLT{} before filtering (Raw), after Bicleaner AI (Filtered) and after deduplication (TMX). Languages are in increasing order of deduplicated sentence pairs.}
\label{tab:v2_bitext_data_stats_full}
\end{table*}

\twocolumn
% \section{Sizes of monolingual datasets}
\label{sec:mono_langs_stats}
\onecolumn
\footnotesize
\begin{longtable}{lcccc}
    % \centering
    \toprule
    \textbf{Language} & \textbf{Segments} & \textbf{Tokens} & \textbf{Characters} & \textbf{Documents} \\
    \midrule
    \endhead
    \bottomrule
    \\
    \caption{Counts of segments, tokens, characters and documents for each language in the monolingual \HPLT{} datasets. Tokens are words as defined by Unix \texttt{wc}.}
    \endfoot
    ace\_Arab & 1.170e+02 & 8.363e+03 & 4.973e+04 & 1.600e+01 \\
    ace\_Latn & 2.062e+05 & 8.196e+06 & 5.083e+07 & 1.293e+04 \\
    afr\_Latn & 3.774e+07 & 1.000e+09 & 5.947e+09 & 1.457e+06 \\
    als\_Latn & 9.510e+07 & 2.713e+09 & 1.610e+10 & 5.385e+06 \\
    amh\_Ethi & 7.006e+06 & 1.959e+08 & 1.031e+09 & 2.955e+05 \\
    ara\_Arab & 2.200e+09 & 4.814e+10 & 2.795e+11 & 8.267e+07 \\
    asm\_Beng & 2.677e+06 & 7.344e+07 & 4.757e+08 & 1.757e+05 \\
    ast\_Latn & 7.426e+06 & 1.950e+08 & 1.244e+09 & 2.732e+05 \\
    awa\_Deva & 1.315e+05 & 6.049e+06 & 2.877e+07 & 7.281e+03 \\
    ayr\_Latn & 1.885e+05 & 3.068e+06 & 2.508e+07 & 9.223e+03 \\
    azb\_Arab & 2.389e+06 & 3.958e+07 & 2.602e+08 & 6.611e+04 \\
    azj\_Latn & 1.266e+08 & 2.569e+09 & 1.962e+10 & 6.485e+06 \\
    bak\_Cyrl & 3.139e+06 & 7.533e+07 & 5.585e+08 & 1.708e+05 \\
    bam\_Latn & 9.172e+04 & 3.982e+06 & 2.074e+07 & 5.721e+03 \\
    ban\_Latn & 6.011e+05 & 1.134e+07 & 7.724e+07 & 1.070e+04 \\
    bel\_Cyrl & 4.884e+07 & 1.212e+09 & 8.540e+09 & 2.320e+06 \\
    bem\_Latn & 1.335e+05 & 4.523e+06 & 3.232e+07 & 6.136e+03 \\
    ben\_Beng & 1.760e+08 & 4.639e+09 & 3.016e+10 & 1.104e+07 \\
    bho\_Deva & 4.583e+05 & 1.347e+07 & 6.865e+07 & 2.864e+04 \\
    bjn\_Arab & 1.953e+04 & 5.482e+05 & 3.317e+06 & 1.112e+03 \\
    bjn\_Latn & 3.663e+05 & 8.048e+06 & 5.597e+07 & 1.876e+04 \\
    bod\_Tibt & 4.650e+05 & 5.781e+06 & 2.685e+08 & 2.744e+04 \\
    bos\_Latn & 2.682e+08 & 7.255e+09 & 4.607e+10 & 1.461e+07 \\
    bug\_Latn & 3.855e+04 & 2.705e+06 & 1.931e+07 & 2.023e+03 \\
    bul\_Cyrl & 6.814e+08 & 1.530e+10 & 9.693e+10 & 2.809e+07 \\
    cat\_Latn & 3.833e+08 & 1.002e+10 & 6.019e+10 & 1.855e+07 \\
    ceb\_Latn & 2.865e+06 & 8.589e+07 & 5.157e+08 & 1.388e+05 \\
    ces\_Latn & 1.927e+09 & 4.208e+10 & 2.739e+11 & 7.529e+07 \\
    cjk\_Latn & 3.670e+04 & 9.647e+05 & 7.432e+06 & 1.196e+03 \\
    ckb\_Arab & 5.226e+06 & 1.426e+08 & 9.128e+08 & 2.737e+05 \\
    crh\_Latn & 1.381e+06 & 3.676e+07 & 2.811e+08 & 1.227e+05 \\
    cym\_Latn & 1.557e+07 & 4.090e+08 & 2.402e+09 & 7.581e+05 \\
    dan\_Latn & 8.730e+08 & 2.120e+10 & 1.334e+11 & 3.384e+07 \\
    deu\_Latn & 1.113e+10 & 2.515e+11 & 1.782e+12 & 4.821e+08 \\
    dik\_Latn & 3.465e+04 & 2.295e+06 & 1.154e+07 & 2.325e+03 \\
    dyu\_Latn & 2.456e+04 & 1.194e+06 & 5.552e+06 & 1.390e+03 \\
    dzo\_Tibt & 3.997e+04 & 4.222e+05 & 7.375e+06 & 1.626e+03 \\
    ell\_Grek & 1.849e+09 & 4.270e+10 & 2.835e+11 & 7.033e+07 \\
    eng\_Latn & 1.165e+11 & 2.862e+12 & 1.708e+13 & 4.389e+09 \\
    epo\_Latn & 2.035e+07 & 4.716e+08 & 2.976e+09 & 8.189e+05 \\
    est\_Latn & 2.644e+08 & 4.742e+09 & 3.602e+10 & 8.449e+06 \\
    eus\_Latn & 3.762e+07 & 7.767e+08 & 6.052e+09 & 1.974e+06 \\
    ewe\_Latn & 1.434e+05 & 4.308e+06 & 2.132e+07 & 3.772e+03 \\
    fao\_Latn & 4.526e+06 & 9.345e+07 & 5.818e+08 & 2.399e+05 \\
    fij\_Latn & 1.789e+05 & 7.263e+06 & 3.769e+07 & 8.914e+03 \\
    fin\_Latn & 9.766e+08 & 1.845e+10 & 1.557e+11 & 3.482e+07 \\
    fon\_Latn & 1.476e+04 & 1.233e+06 & 5.335e+06 & 1.226e+03 \\
    fra\_Latn & 1.056e+10 & 2.370e+11 & 1.457e+12 & 4.018e+08 \\
    fur\_Latn & 7.300e+05 & 2.082e+07 & 1.147e+08 & 3.667e+04 \\
    fuv\_Latn & 1.340e+05 & 5.143e+06 & 2.990e+07 & 7.760e+03 \\
    gaz\_Latn & 9.736e+05 & 2.888e+07 & 2.192e+08 & 4.914e+04 \\
    gla\_Latn & 3.307e+06 & 8.066e+07 & 4.836e+08 & 1.374e+05 \\
    gle\_Latn & 1.099e+07 & 2.957e+08 & 1.749e+09 & 4.908e+05 \\
    glg\_Latn & 6.118e+07 & 1.639e+09 & 1.011e+10 & 3.020e+06 \\
    grn\_Latn & 1.713e+06 & 3.072e+07 & 2.186e+08 & 7.342e+04 \\
    guj\_Gujr & 2.064e+07 & 5.768e+08 & 3.386e+09 & 1.134e+06 \\
    \pagebreak
    hat\_Latn & 4.635e+06 & 1.223e+08 & 6.389e+08 & 2.127e+05 \\
    hau\_Latn & 5.688e+06 & 1.526e+08 & 8.535e+08 & 3.159e+05 \\
    heb\_Hebr & 4.666e+08 & 9.966e+09 & 5.682e+10 & 1.712e+07 \\
    hin\_Deva & 2.674e+08 & 8.637e+09 & 4.396e+10 & 1.365e+07 \\
    hne\_Deva & 5.500e+04 & 2.199e+06 & 1.059e+07 & 2.806e+03 \\
    hrv\_Latn & 2.971e+08 & 7.307e+09 & 4.800e+10 & 1.230e+07 \\
    hun\_Latn & 1.419e+09 & 3.052e+10 & 2.252e+11 & 5.187e+07 \\
    hye\_Armn & 6.524e+07 & 1.405e+09 & 1.072e+10 & 3.599e+06 \\
    ibo\_Latn & 1.411e+06 & 3.829e+07 & 2.052e+08 & 5.629e+04 \\
    ilo\_Latn & 1.120e+06 & 2.478e+07 & 1.568e+08 & 4.875e+04 \\
    ind\_Latn & 2.389e+09 & 5.462e+10 & 3.842e+11 & 9.814e+07 \\
    isl\_Latn & 6.964e+07 & 1.536e+09 & 9.593e+09 & 2.841e+06 \\
    ita\_Latn & 5.127e+09 & 1.274e+11 & 8.206e+11 & 2.218e+08 \\
    jav\_Latn & 6.431e+06 & 1.378e+08 & 9.375e+08 & 1.960e+05 \\
    jpn\_Jpan & 2.327e+10 & 4.236e+10 & 9.011e+11 & 4.177e+08 \\
    kab\_Latn & 3.452e+05 & 9.222e+06 & 5.419e+07 & 1.510e+04 \\
    kac\_Latn & 1.594e+05 & 5.955e+06 & 2.840e+07 & 7.587e+03 \\
    kam\_Latn & 1.426e+04 & 6.740e+05 & 4.645e+06 & 1.183e+03 \\
    kan\_Knda & 2.493e+07 & 5.329e+08 & 4.298e+09 & 1.336e+06 \\
    kas\_Arab & 2.711e+04 & 6.780e+05 & 3.468e+06 & 9.490e+02 \\
    kas\_Deva & 1.357e+03 & 3.194e+04 & 1.854e+05 & 1.060e+02 \\
    kat\_Geor & 6.372e+07 & 1.244e+09 & 1.016e+10 & 3.335e+06 \\
    kaz\_Cyrl & 8.101e+07 & 1.409e+09 & 1.113e+10 & 2.637e+06 \\
    kbp\_Latn & 4.679e+04 & 4.258e+06 & 2.090e+07 & 7.075e+03 \\
    kea\_Latn & 4.391e+04 & 1.143e+06 & 6.144e+06 & 1.962e+03 \\
    khk\_Cyrl & 5.347e+07 & 1.342e+09 & 9.327e+09 & 2.121e+06 \\
    khm\_Khmr & 9.864e+06 & 1.138e+08 & 2.122e+09 & 7.010e+05 \\
    kik\_Latn & 5.193e+04 & 1.428e+06 & 9.292e+06 & 3.995e+03 \\
    kin\_Latn & 1.917e+06 & 5.074e+07 & 3.671e+08 & 9.270e+04 \\
    kir\_Cyrl & 1.004e+07 & 2.467e+08 & 1.925e+09 & 6.761e+05 \\
    kmb\_Latn & 1.180e+04 & 3.831e+05 & 2.068e+06 & 5.310e+02 \\
    kmr\_Latn & 7.147e+06 & 1.959e+08 & 1.123e+09 & 3.643e+05 \\
    knc\_Arab & 1.083e+04 & 2.620e+05 & 1.302e+06 & 2.450e+02 \\
    knc\_Latn & 1.052e+04 & 2.409e+06 & 1.195e+07 & 2.472e+03 \\
    kon\_Latn & 4.748e+04 & 1.944e+06 & 1.127e+07 & 2.542e+03 \\
    kor\_Hang & 1.358e+09 & 1.970e+10 & 8.923e+10 & 3.887e+07 \\
    lao\_Laoo & 3.200e+05 & 5.178e+06 & 8.468e+07 & 2.950e+04 \\
    lij\_Latn & 1.577e+05 & 5.593e+06 & 3.146e+07 & 8.371e+03 \\
    lim\_Latn & 7.140e+06 & 1.806e+08 & 1.125e+09 & 3.679e+05 \\
    lin\_Latn & 2.003e+05 & 5.555e+06 & 3.292e+07 & 7.588e+03 \\
    lit\_Latn & 3.222e+08 & 6.676e+09 & 5.039e+10 & 1.334e+07 \\
    lmo\_Latn & 2.125e+06 & 5.964e+07 & 3.454e+08 & 1.462e+05 \\
    ltg\_Latn & 1.514e+05 & 3.790e+06 & 2.688e+07 & 9.209e+03 \\
    ltz\_Latn & 5.059e+06 & 1.072e+08 & 7.104e+08 & 2.469e+05 \\
    lua\_Latn & 3.869e+04 & 1.368e+06 & 9.005e+06 & 1.083e+03 \\
    lug\_Latn & 4.075e+05 & 9.176e+06 & 6.796e+07 & 2.128e+04 \\
    luo\_Latn & 8.412e+04 & 3.727e+06 & 2.033e+07 & 4.153e+03 \\
    lus\_Latn & 3.433e+06 & 1.252e+08 & 6.520e+08 & 1.604e+05 \\
    lvs\_Latn & 1.738e+08 & 3.461e+09 & 2.518e+10 & 6.772e+06 \\
    mag\_Deva & 1.929e+04 & 8.906e+05 & 4.283e+06 & 3.280e+02 \\
    mai\_Deva & 6.455e+05 & 1.779e+07 & 9.674e+07 & 2.498e+04 \\
    mal\_Mlym & 4.800e+07 & 9.737e+08 & 9.489e+09 & 3.105e+06 \\
    mar\_Deva & 3.632e+07 & 9.807e+08 & 6.622e+09 & 2.080e+06 \\
    min\_Latn & 6.008e+05 & 1.098e+07 & 7.477e+07 & 2.504e+04 \\
    mkd\_Cyrl & 5.701e+07 & 1.485e+09 & 9.440e+09 & 3.566e+06 \\
    mlt\_Latn & 8.675e+06 & 1.958e+08 & 1.442e+09 & 3.673e+05 \\
    \pagebreak
    mni\_Beng & 6.576e+04 & 1.627e+06 & 1.179e+07 & 2.934e+03 \\
    mos\_Latn & 1.910e+04 & 8.075e+05 & 3.864e+06 & 9.310e+02 \\
    mri\_Latn & 2.795e+06 & 8.676e+07 & 4.243e+08 & 1.083e+05 \\
    mya\_Mymr & 3.050e+07 & 4.532e+08 & 5.819e+09 & 1.368e+06 \\
    nld\_Latn & 3.075e+09 & 7.141e+10 & 4.511e+11 & 1.387e+08 \\
    nno\_Latn & 3.460e+07 & 8.603e+08 & 5.404e+09 & 1.423e+06 \\
    nob\_Latn & 6.760e+08 & 2.154e+10 & 1.332e+11 & 2.705e+07 \\
    npi\_Deva & 3.714e+07 & 1.128e+09 & 7.256e+09 & 2.778e+06 \\
    nso\_Latn & 1.433e+05 & 5.322e+06 & 2.749e+07 & 6.066e+03 \\
    nus\_Latn & 8.514e+03 & 3.932e+05 & 1.882e+06 & 2.720e+02 \\
    nya\_Latn & 1.344e+06 & 2.706e+07 & 2.029e+08 & 5.312e+04 \\
    oci\_Latn & 4.195e+06 & 1.027e+08 & 6.354e+08 & 1.899e+05 \\
    ory\_Orya & 3.596e+06 & 1.201e+08 & 7.815e+08 & 4.129e+05 \\
    pag\_Latn & 8.583e+04 & 5.657e+06 & 3.352e+07 & 6.900e+03 \\
    pan\_Guru & 1.174e+07 & 3.722e+08 & 1.902e+09 & 5.846e+05 \\
    pap\_Latn & 1.387e+06 & 4.671e+07 & 2.541e+08 & 8.981e+04 \\
    pbt\_Arab & 8.455e+06 & 2.794e+08 & 1.304e+09 & 4.665e+05 \\
    pes\_Arab & 3.963e+09 & 8.855e+10 & 4.551e+11 & 9.050e+07 \\
    plt\_Latn & 4.736e+06 & 1.171e+08 & 8.103e+08 & 2.078e+05 \\
    pol\_Latn & 4.461e+09 & 8.953e+10 & 6.316e+11 & 1.754e+08 \\
    por\_Latn & 6.125e+09 & 1.463e+11 & 8.965e+11 & 2.378e+08 \\
    prs\_Arab & 6.900e+07 & 1.844e+09 & 9.567e+09 & 2.839e+06 \\
    quy\_Latn & 4.943e+05 & 1.731e+07 & 1.434e+08 & 3.694e+04 \\
    ron\_Latn & 1.697e+09 & 4.005e+10 & 2.507e+11 & 6.588e+07 \\
    run\_Latn & 1.752e+06 & 4.444e+07 & 3.165e+08 & 1.373e+05 \\
    rus\_Cyrl & 2.629e+10 & 5.409e+11 & 3.908e+12 & 8.847e+08 \\
    sag\_Latn & 5.190e+04 & 3.612e+06 & 1.674e+07 & 3.161e+03 \\
    san\_Deva & 3.281e+06 & 4.380e+07 & 3.592e+08 & 5.491e+04 \\
    sat\_Olck & 4.580e+04 & 1.085e+06 & 6.266e+06 & 2.566e+03 \\
    scn\_Latn & 1.650e+06 & 4.239e+07 & 2.523e+08 & 8.197e+04 \\
    shn\_Mymr & 9.214e+04 & 1.648e+06 & 2.121e+07 & 6.003e+03 \\
    sin\_Sinh & 3.371e+07 & 7.956e+08 & 4.981e+09 & 1.153e+06 \\
    slk\_Latn & 4.943e+08 & 1.063e+10 & 7.037e+10 & 2.183e+07 \\
    slv\_Latn & 2.386e+08 & 5.435e+09 & 3.526e+10 & 1.028e+07 \\
    smo\_Latn & 1.012e+06 & 3.709e+07 & 1.861e+08 & 4.586e+04 \\
    sna\_Latn & 1.202e+06 & 2.392e+07 & 1.926e+08 & 6.108e+04 \\
    snd\_Arab & 2.826e+06 & 8.953e+07 & 4.286e+08 & 1.003e+05 \\
    som\_Latn & 1.638e+07 & 3.888e+08 & 2.565e+09 & 9.665e+05 \\
    sot\_Latn & 1.085e+06 & 3.100e+07 & 1.715e+08 & 4.392e+04 \\
    spa\_Latn & 1.212e+10 & 3.220e+11 & 1.954e+12 & 5.031e+08 \\
    srd\_Latn & 9.171e+05 & 2.389e+07 & 1.487e+08 & 5.382e+04 \\
    srp\_Cyrl & 9.381e+07 & 2.519e+09 & 1.616e+10 & 4.123e+06 \\
    ssw\_Latn & 6.213e+04 & 9.943e+05 & 8.821e+06 & 2.036e+03 \\
    sun\_Latn & 3.238e+06 & 6.963e+07 & 4.753e+08 & 1.148e+05 \\
    swe\_Latn & 1.755e+09 & 4.011e+10 & 2.511e+11 & 6.681e+07 \\
    swh\_Latn & 3.431e+07 & 7.177e+08 & 4.664e+09 & 1.374e+06 \\
    szl\_Latn & 6.366e+05 & 1.468e+07 & 1.038e+08 & 4.093e+04 \\
    tam\_Taml & 1.686e+08 & 2.981e+09 & 2.624e+10 & 6.106e+06 \\
    taq\_Latn & 1.388e+04 & 1.544e+06 & 8.845e+06 & 1.747e+03 \\
    tat\_Cyrl & 1.345e+07 & 2.967e+08 & 2.157e+09 & 6.307e+05 \\
    tel\_Telu & 3.919e+07 & 8.354e+08 & 6.505e+09 & 2.058e+06 \\
    tgk\_Cyrl & 2.485e+07 & 6.248e+08 & 4.590e+09 & 1.261e+06 \\
    tgl\_Latn & 5.288e+07 & 1.346e+09 & 8.131e+09 & 1.869e+06 \\
    tha\_Thai & 3.391e+08 & 3.506e+09 & 5.998e+10 & 1.770e+07 \\
    tir\_Ethi & 1.128e+06 & 3.672e+07 & 1.816e+08 & 6.469e+04 \\
    tpi\_Latn & 2.824e+05 & 1.251e+07 & 6.453e+07 & 1.398e+04 \\
    \pagebreak
    tsn\_Latn & 1.322e+05 & 5.273e+06 & 2.767e+07 & 6.050e+03 \\
    tso\_Latn & 2.212e+05 & 8.668e+06 & 4.929e+07 & 1.101e+04 \\
    tuk\_Latn & 3.355e+06 & 7.068e+07 & 5.700e+08 & 1.710e+05 \\
    tum\_Latn & 9.901e+04 & 2.876e+06 & 2.110e+07 & 4.384e+03 \\
    tur\_Latn & 2.575e+09 & 5.167e+10 & 3.896e+11 & 1.166e+08 \\
    twi\_Latn & 1.256e+05 & 4.696e+06 & 2.418e+07 & 5.860e+03 \\
    uig\_Arab & 8.982e+06 & 2.239e+08 & 1.747e+09 & 4.424e+05 \\
    ukr\_Cyrl & 1.169e+09 & 2.523e+10 & 1.829e+11 & 4.740e+07 \\
    umb\_Latn & 5.991e+04 & 2.431e+06 & 1.541e+07 & 2.471e+03 \\
    urd\_Arab & 5.063e+07 & 2.126e+09 & 1.001e+10 & 3.194e+06 \\
    uzn\_Latn & 1.480e+07 & 3.513e+08 & 2.846e+09 & 7.069e+05 \\
    vec\_Latn & 1.579e+06 & 3.526e+07 & 2.180e+08 & 8.480e+04 \\
    vie\_Latn & 3.020e+09 & 8.320e+10 & 3.795e+11 & 1.007e+08 \\
    war\_Latn & 2.009e+05 & 5.889e+06 & 3.557e+07 & 1.387e+04 \\
    wol\_Latn & 1.615e+05 & 5.463e+06 & 2.754e+07 & 5.679e+03 \\
    xho\_Latn & 1.821e+06 & 3.034e+07 & 2.587e+08 & 6.309e+04 \\
    ydd\_Hebr & 2.940e+06 & 7.753e+07 & 4.585e+08 & 1.283e+05 \\
    yor\_Latn & 1.469e+06 & 4.281e+07 & 2.178e+08 & 6.613e+04 \\
    yue\_Hant & 1.235e+06 & 3.268e+06 & 7.430e+07 & 6.129e+04 \\
    zho\_Hans & 4.245e+10 & 7.403e+10 & 2.352e+12 & 1.247e+09 \\
    zho\_Hant & 4.480e+09 & 9.510e+09 & 2.868e+11 & 1.571e+08 \\
    zsm\_Latn & 5.798e+08 & 1.148e+10 & 7.843e+10 & 1.842e+07 \\
    zul\_Latn & 2.710e+06 & 4.436e+07 & 3.808e+08 & 1.136e+05 \\

\end{longtable}
\label{tab:monostats_long}

\twocolumn
\normalsize

\twocolumn

\section{Sources of web crawls}
\label{sec:appendix_crawlsources}
\begin{table}[ht]
\centering
\small
\begin{tabular}{p{3.7cm}cr}
\toprule
\textbf{Name} & \textbf{Years} & \textbf{Size (TB)} \\ 
\midrule
\textbf{IA full crawls} & \textbf{2012--2020} & \textbf{3390} \\ 
\midrule
wide5 & 2012--2012 & 365 \\ 
wide6 & 2012--2013 & 204 \\ 
wide10 & 2014--2014 & 91 \\ 
wide11 & 2014--2014 & 420 \\ 
wide12 & 2015--2015 & 449 \\ 
wide15 & 2016--2017 & 358 \\ 
wide16 & 2017--2018 & 768 \\ 
wide17 & 2018--2020 & 641 \\ 
survey3 & 2015--2016 & 94 \\ 
\midrule
\textbf{CC full crawls} & \textbf{2014--2022} & \textbf{743} \\ 
\midrule
CC-MAIN-2014-35 & 2014 & 43 \\ 
CC-MAIN-2014-42 & 2014 & 54 \\ 
CC-MAIN-2015-11 & 2015 & 29 \\ 
CC-MAIN-2015-48 & 2015 & 30 \\ 
CC-MAIN-2017-04 & 2017 & 54 \\ 
CC-MAIN-2018-05 & 2018 & 75 \\ 
CC-MAIN-2018-22 & 2018 & 52 \\ 
CC-MAIN-2018-43 & 2018 & 59 \\ 
CC-MAIN-2021-43 & 2021 & 86 \\ 
CC-MAIN-2022-27 & 2022 & 85 \\ 
CC-MAIN-2022-40 & 2022 & 83 \\ 
CC-MAIN-2022-49 & 2022 & 93 \\ 
\midrule
\textbf{Partial crawls} & \textbf{2013-2023} & \textbf{317} \\ 
\midrule
1\% of WARCs from 81 \ac{cc} crawls & 2013-2023 & 46 \\ 
7\% of \ac{ia} ArchiveBot & 2013-2023 & 271 \\ \bottomrule
\end{tabular}
\caption{List of web crawls used to construct \HPLT{}. From \ac{ia}, we use 8 Wide crawls, 1 Survey crawl containing main pages of websites and a random sample of 7\% of items from \ac{ia} ArchiveBot. From \ac{cc}, we use 12 randomly-selected full crawls, plus a 1\% sample of WARCs from each of the other 81 available crawls.}
\label{tab:crawl-sources}
\end{table}

\section{Yields of different crawls}
\label{sec:appendix_yields}

\begin{figure*}[htbp]
    \centering % Center the image
    \includegraphics[width=0.99\textwidth]{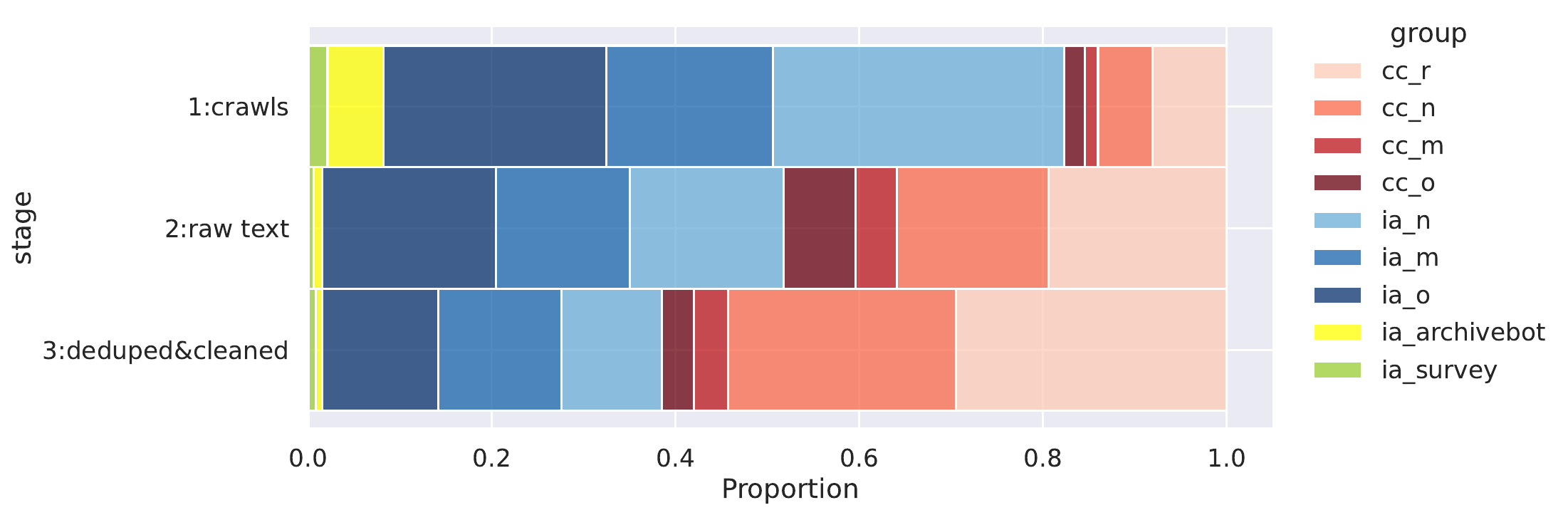} 
    \caption{Proportions of data from different groups of crawls at various processing stages. Crawls were quantified in TB of compressed WARC files, while raw texts and deduplicated cleaned texts in characters.} 
    \label{fig:stage_groups_proportions}
\end{figure*}

\begin{figure}[htbp]
    \centering % Center the image
    \includegraphics[width=0.5\textwidth]{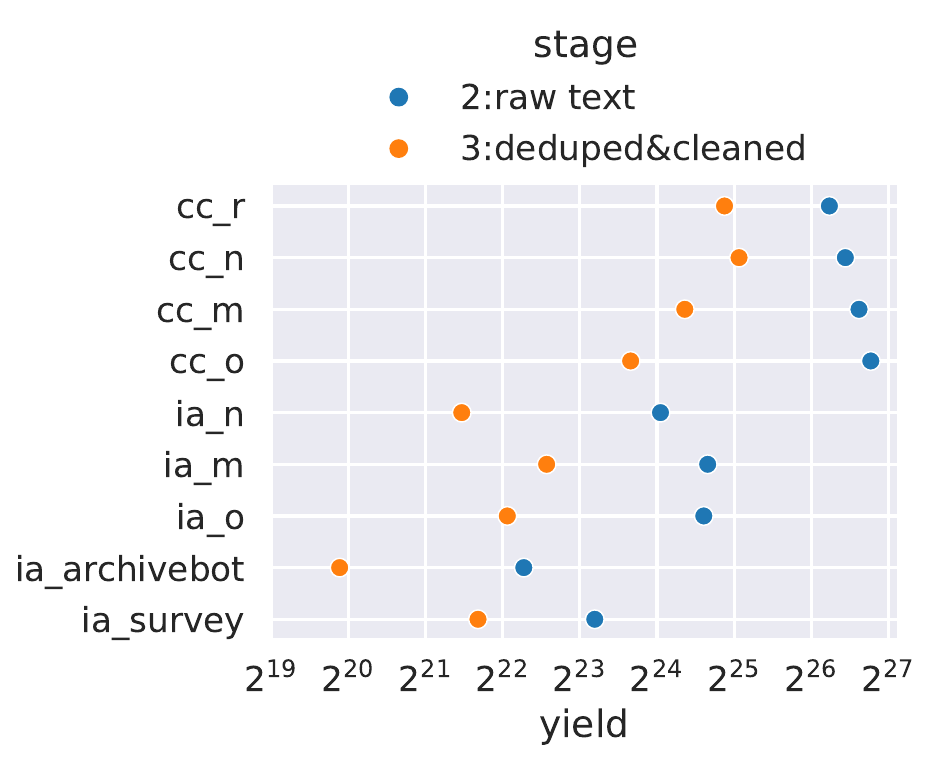} 
    \caption{Yields (in characters per 1 GB of raw compressed crawls) of different crawls at different stages.} 
    \label{fig:stage_groups_yields}
\end{figure}

To figure out how different web crawls contribute to our datasets and which crawls are the most promising sources of monolingual corpora in general, we compared crawls from two points of view: the amount of texts extracted from each crawl and the quality of these texts. In this section, we study crawls from the first point of view, while in~\ref{sec:appendix_r2inspection} the results of manual quality inspection are presented.

To make a comparison, we group all crawls into groups according to their age and source. The oldest IA wide crawls from 2012-2014 (from wide5 up to wide11) are assigned to the group \texttt{ia\_o}, the newest wide16, wide17 crawls from 2017-2020 to the group \texttt{ia\_n}, and the wide12, wide15 crawls in the middle to the group \texttt{ia\_m}. CC crawls are split by age following the same time periods, but additionally a group \texttt{cc\_r} is introduced for the recent CC crawls from 2021-2023 (we don't have IA wide crawls from this time period). Finally, the IA survey3 and ArchiveBot crawls form their own groups \texttt{ia\_survey} and \texttt{ia\_archivebot}. In total, we have 9 groups of crawls. 

For different processing stages, Figure~\ref{fig:stage_groups_proportions} visualizes how much data comes from different groups of crawls. While originally less than 20\% of our crawls are CC crawls, they contribute about half of the raw text before duplication and more than 60\% of the text after deduplication and cleaning. Especially high-yielding are the new and recent CC crawls, they are only 6\% and 8\% of all crawls in size but contribute 28\% and 30\% of text (both when counting in characters and in documents) to the cleaned version. On the other hand, the newest IA wide crawls are 32\% of all crawls in size but contribute only about 11\% of text.

Figure~\ref{fig:stage_groups_yields} suggests another point of view showing yields for different crawls, or more specifically, how much text (measured in the number of characters) is extracted from 1 GB of compressed WARC files for each crawl. Evidently, CC crawls have the highest yields, especially the newer ones. Compared to the newer CC crawls, for the older CC crawls more data is filtered during deduplication and cleaning,  giving finally lower yields despite a bit higher yields of raw texts. IA wide crawls have 4-8x smaller yields than CC crawls. The survey IA crawl has a comparable yield to the wide crawls in the final dataset. Since they are publicly available, it probably makes sense to employ more of these crawls in the future. Finally, the ArchiveBot IA crawl has remarkably low yields.

\begin{figure}[htb]
    \centering
    \includegraphics[width=0.48\textwidth]{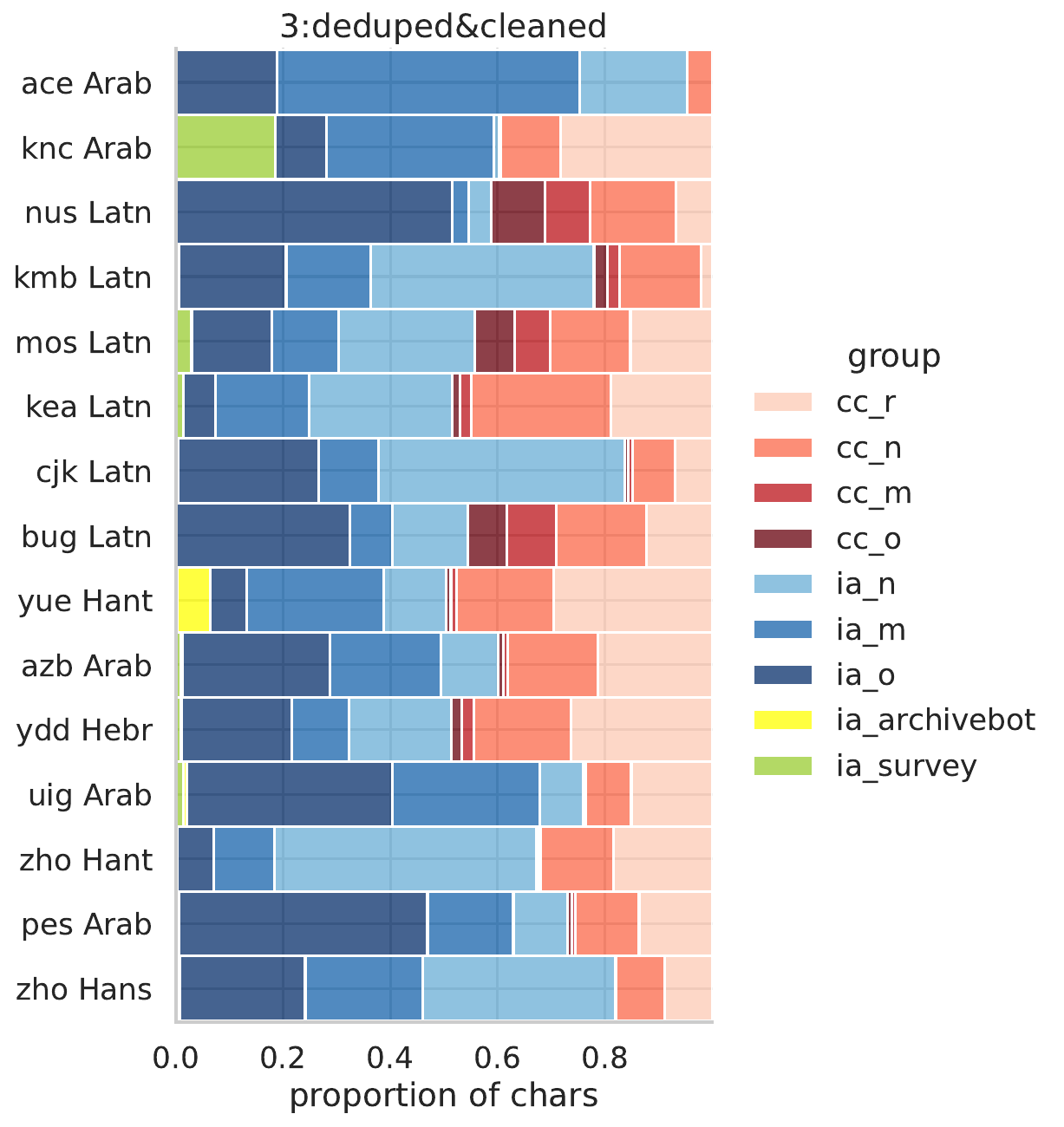}
    \caption{Proportions of texts from different groups of crawls for the 15 languages with the largest contribution of IA crawls.}
    \label{fig:fig:group-contrib-perlang}
\end{figure}

Despite having a lower contribution in general, for some languages, IA crawls supply the majority of texts. Figure~\ref{fig:fig:group-contrib-perlang} shows 15 languages with the highest proportion of texts from IA crawls. They include both high-resourced (Chinese, Western Persian) and low-resourced languages. Deduplication and cleaning significantly reduce the number of languages with high contribution of IA. For instance, before deduplication and cleaning there are 49 languages having more than 70\% of texts (characters) coming from IA and only 6 such languages after. 

\label{sec:appendix_ngrams}
\section{Frequent \textit{n}-grams}
We obtain frequent $n$-grams (up to order 5) in each dataset after tokenizing text and applying some restrictions:
\begin{compactitem}
\item $n$-grams must start and end in the same segment (i.e. no line breaks are allowed in the middle of a $n$-gram)
\item $n$-grams containing any punctuation are discarded
\item $n$-grams that start or end in stopwords are discarded
\item $n$-grams are calculated case-insensitive
\item all tokens in the $n$-gram must have at least one alphabetic character
\end{compactitem}

\begin{figure*}[ht]
    \centering %
    \includegraphics[width=\textwidth]{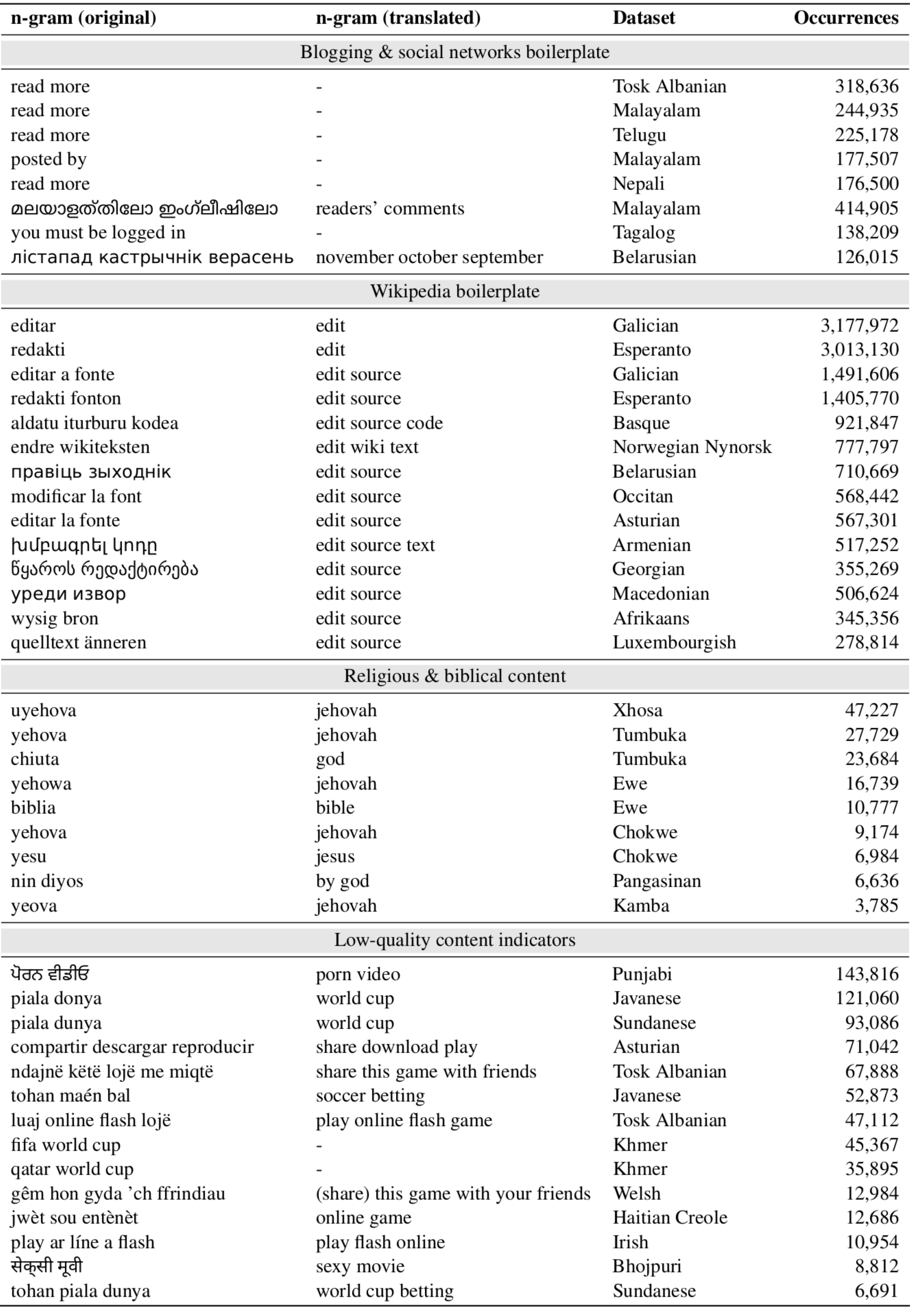} 
    \caption{Frequent n-grams in monolingual datasets.}
    \label{tab:mono_ngrams}
\end{figure*}

\begin{figure*}[ht]
    \centering %
    \includegraphics[width=\textwidth]{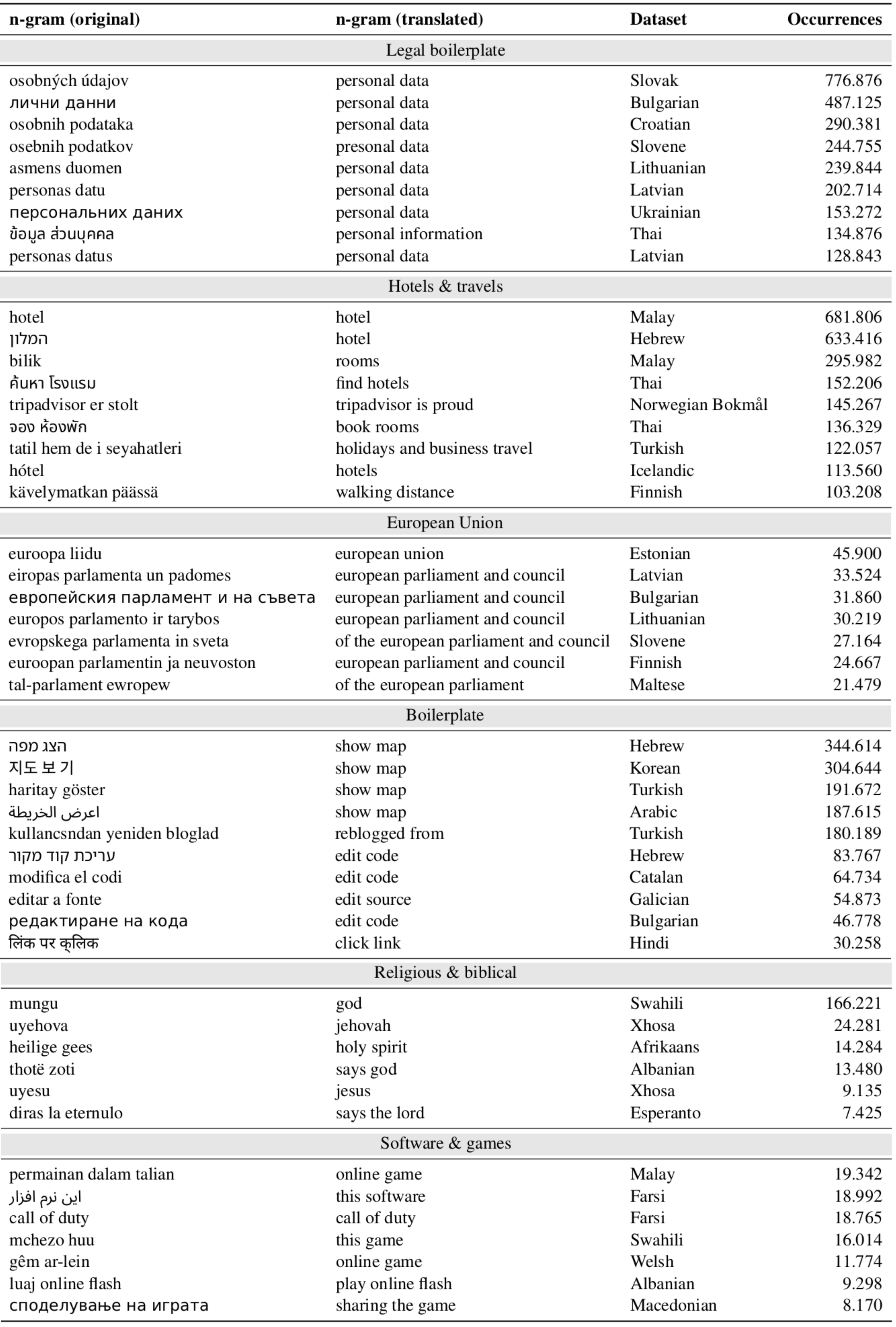}
    \caption{Frequent n-grams in parallel datasets (non-English side).}
    \label{tab:para_ngrams}
\end{figure*}

\Cref{tab:mono_ngrams,tab:para_ngrams} show the 5 most frequent $n$-grams (orders 1 to 5) in \HPLT{}. In the case of parallel datasets, $n$-grams are selected from the target (non-English) side of the segments. Translation to English is obtained with Google Translate.\footnote{\url{https://translate.google.com/}}

We find that most datasets (both monolingual and parallel) contain frequent $n$-grams that seem to be boilerplate, such as "edit source", "read more", "click button" or "view map". This kind of content usually comes from Wikipedia and Blogspot. In the monolingual datasets, there is a large amount of text that seems to come from headers or footers in news webpages, e.g. "latest news". Biblical $n$-grams (such as "god" or "jehovah") are also very frequent in some datasets, notably African languages, matching our observations about frequent domains (\Cref{sec:appendix_domains}). Some frequent $n$-grams suggest poor-quality content in some datasets, since they seem to be related to downloads webpages, online game platforms or betting sites. 

For the parallel datasets, we observe that on the English side the frequent $n$-grams are very similar across all languages. For the languages with the most data, hotels and legal notices are the most common kind of $n$-grams. The smaller parallel datasets tend to exhibit more variety of $n$-grams and include $n$-grams alluding to political leaders or city names, which suggest more locally-generated content (probably from news sites). Finally, frequent $n$-grams in parallel datasets from Eastern European countries usually contain mentions to European institutions (such as the European Parliament or the European Commission). This matches our observations on \acp{tld} in \Cref{sec:appendix_tlds}.

%\includepdf{tables/para_ngrams.pdf}
%\input{tables/para_ngrams}

% \begin{compactitem}
%    \item
%    %\item Similarly, in monolingual datasets, texts that seem to come from headers or footers in news webpages get a lots of counts, for example "latest news"
%     \item Biblical $n$-grams (such as "god" or "jehovah") are also very frequent in some datasets (noticeable, in African languages), matching our observations about frequent domains (\ref{sec:appendix_domains}.
%     \item Some frequent $n$-grams suggest poor-quality content in some datasets, since they seem to be related to downloads webpages, online game platforms or betting sites.
%     \item  In the case of bilingual datasets, we observe that the frequent $n$-grams are very similar across all languages (looking at the English side only), especially among larger ones where hotels and legal notices are the most common kind of $n$-grams.
%     \item Smaller bilingual datasets tend to exhibit more variety of $n$-grams. They also include $n$-grams alluding to political leaders or city names, suggesting a more locally-generated content (probably from news sites).
%     \item Finally, and matching our observations on TLDs distribution, frequent $n$-grams in bilingual datasets from Eastern European countries usually contain mentions to European institutions (such as the European Parliament or the European Commission)
% \end{compactitem}

% domains here
\section{Frequent domains}
\label{sec:appendix_domains}

Inspecting the most common domain names in the datasets is one way to understand the type of content we can find in it. \Cref{tab:mono_domains} gives the datasets with the highest proportion of frequent domain classes, and \Cref{tab:para_domains} gives the datasets with the highest proportion of frequent domain classes for the parallel data. We make the following general observations:

\begin{compactitem}
    \item Mid-to-large-sized datasets show a wider variety of domains with no clear majority source. However, in monolingual datasets, blogging platforms usually get a significant portion of the total (\Cref{tab:mono_domains}).
    \item Wikipedia tends to be among the most frequent domains for both monolingual and parallel datasets. It is usually the most frequent domain for smaller language datasets (\Cref{tab:mono_domains}).
    \item Hotel and travel webpages are much more frequent in the larger parallel datasets and very infrequent in the monolingual data (\Cref{tab:para_domains}).
    \item News and media outlets are also a frequent content source in monolingual datasets, with some news websites getting a significant percentage in different datasets: for example, regional websites from the Free Radio Europe\footnote{\url{https://www.rferl.org/navigation/allsites}} or Voice of America\footnote{\url{https://www.voanews.com/navigation/allsites}} networks (\Cref{tab:mono_domains}).
    \item Religious and biblical content is also very frequent in the smaller monolingual and parallel datasets. This is specially notable in the case of African languages, which often get more than three quarters of their content from such sites (\Cref{tab:mono_domains}).
    \item Software and online gaming websites are usually among the top-10 most frequent domains in almost all parallel datasets.  
    \item Chinese shopping websites are common in the larger parallel datasets of non-European languages (Vietnamese, Japanese, Korean, Arabic and Turkish).
    \item No pornographic webpages appear in the top domains, implying our filter for such content worked as expected.
\end{compactitem}

\begin{table}[ht]
    \centering
    \adjustbox{max width=\linewidth}{
    \footnotesize
    \begin{tabular}{lr}
    \toprule
    \textbf{Dataset} & \textbf{\% of documents} \\
    \midrule
     \rowcolor{gray!20} 
     \multicolumn{2}{c}{Blogging platforms} \\
    \midrule
        Standard Malay & 50\%   \\
        Magahi &    48\% \\
        %Kachin & 41\% \\
        %Banjar (Arabic) & 39\% \\
        %Dzongkha & 29\% \\
        Greek & 24\% \\
        Cantonese & 22\% \\
        %South Azerbaijani & 22\% \\
        %Balinese &  22\% \\
        %Sinhala & 21\%  \\
        Portuguese & 16\% \\
        Finnish & 16\% \\
        Swedish & 13\% \\
        Spanish & 10\% \\
    \midrule
    \rowcolor{gray!20} 
    \multicolumn{2}{c}{Wikipedia} \\
    \midrule
        Santali & 90\%  \\
        Ligurian & 80\% \\
        Waray & 74\% \\
        Iloko & 66\% \\
        Esperanto & 66\% \\
        Occitan & 62\% \\
        Sicilian & 55\% \\
        %Limburgish & 54\%  \\
        %Asturian & 54\% \\
        %Kabiyè & 54\% \\
    \midrule
    \rowcolor{gray!20} 
    \multicolumn{2}{c}{News \& media} \\
    \midrule
        Crimean Tatar & 61\% \\
        Tigrinya & 48\% \\
        Banjar (Arabic) & 46\% \\
        Nigerian Fulfulde & 38\% \\
        Turkmen & 32\% \\
        Kyrgyz & 30\% \\
        Rundi & 29\% \\
        %Oromo & 22\%  \\
        %Bashkir & 21\% \\
        %Ganda & 20\% \\       
    \midrule
    \rowcolor{gray!20} 
    \multicolumn{2}{c}{Religious \& biblical} \\
    \midrule
        Dyula & 99\% \\
        Fon & 96\% \\
        Bemba & 95\% \\
        Tumbuka & 94\% \\        
        Kamba & 93\% \\
        Chokwe & 92\% \\        
        Central Kanuri (Latin) & 91\% \\
        Luba-Lulua & 88\% \\
        Sango & 88\% \\
        Umbundu & 84\% \\
    
    \bottomrule
    \end{tabular}}
    \caption{Languages with the biggest proportion of frequent domain classes in the monolingual \HPLT{} corpora.}
    \label{tab:mono_domains}
\end{table}
%Table format copied from mono_datasets_table.tex  

\begin{table}[ht]
    \centering
    \adjustbox{max width=\linewidth}{
    \footnotesize
    \begin{tabular}{lr}
    \toprule
    \textbf{Language} & \textbf{\% of segments} \\
    \midrule
    \rowcolor{gray!20} 
    \multicolumn{2}{c}{Hotels \& travels} \\
    \midrule
        Icelandic & 42\% \\
        Malay & 34\% \\
        Hebrew & 26\% \\
        Lithuanian & 21\% \\
        Korean & 18\% \\
        Thai & 16\% \\
        Norwegian Bokmål & 16\% \\
        Japanese & 15\% \\
    \midrule
     \rowcolor{gray!20} 
     \multicolumn{2}{c}{Wikipedia} \\
    \midrule
        Norwegian Nynorsk & 72\% \\
        Galician & 37\% \\
        Esperanto & 36\% \\
        Kannada & 22\% \\
        Macedonian & 21\% \\
        Telugu & 19\% \\
        Catalan & 19\% \\
    \midrule
    
    \rowcolor{gray!20} 
    \multicolumn{2}{c}{Religious \& biblical } \\
    \midrule
        Xhosa & 70\% \\
        Esperanto & 28\% \\
        Swahili & 20\% \\
        Nepali & 16\% \\
        Icelandic & 14\% \\
        Albanian & 14\% \\
    
    \bottomrule
    \end{tabular}}
    \caption{Frequent domain classes in parallel \HPLT{} datasets for different languages (non-English side).}
    \label{tab:para_domains}
\end{table}

\section{Geographic TLDs}
\label{sec:appendix_tlds}
\Cref{tab:mono_tlds,tab:para_tlds} list the most frequent examples of geographic \acp{tld} in the monolingual and parallel \HPLT{} corpora respectively. We make the following observations in addition to those made in the main text:

% We inspect the distribution of \acp{tld} (especially, geographic TLDs) in each dataset can help us to get an idea of which language variants might be included in our datasets. 
\begin{compactitem}
    \item In general, the most frequent \acp{tld} in many of the datasets are generic (such as \texttt{.com}, \texttt{.org} or \texttt{.info}).
    \item Some \acp{tld} are frequent because they "sound good" rather than indicating the kind of content or language: \texttt{.icu} (because it reads like "I see you"), \texttt{.is} (official TLD for Iceland, but used as a verb and very noticeably in \url{bible.is}, a religious webpage whose domain is usually in the top-10 most frequent TLDs), \texttt{.tv} (for the country of Tuvalu, but widely used for TV-related web pages),  \texttt{.co} (for Colombia, but mostly used for companies), \texttt{.no} (for Norway, but used as a negative particle), \texttt{.nu} (for the island of Niue, but used because it sounds like "new"), etc.
    \item There are common \acp{tld} for super-national territories: \texttt{.eu} (European Union), \texttt{.africa}, \texttt{.asia}, etc.
    \item In our monolingual datasets, there is frequently one geographic \ac{tld} among the 10 most frequent ones that clearly surpasses the others. The "winning" \ac{tld} is usually from the country where the dataset language is spoken most, indicating that the text content is probably in the correct language. The percentage of this "winning country" varies depending on the amount of general purpose \ac{tld} in the dataset, but it is in general higher for European countries. This "winning" geographic \ac{tld}, in the case of parallel datasets, is less frequent and, when present, its portion of the total is noticeably lower than for the monolingual datasets.
    \item Many African languages do not have a significant portion of geographic \acp{tld} (beyond the aforementioned \texttt{.is}, \texttt{.no}, etc).
    \item For some languages, there are a few countries \acp{tld} in the top-10 from closely related countries or territories (for example, from former colonial rulers (i.e. African languages datasets) or with geostrategic interests (i.e. \texttt{.ru} (Russia) appearing in all former Soviet states). This may indicate "language contamination" in the data.
\end{compactitem}

\begin{table}[tp]
%    \centering
    \adjustbox{max width=\linewidth}{
    \footnotesize
    \begin{tabular}{lrcl}
    \toprule
    % \textbf{Dataset}  & \textbf{\% of documents} & \textbf{TLD}  & \textbf{Country/Territory}\\
    \multirow{2}{*}{\textbf{Language}}  & \textbf{\% of} & \multirow{2}{*}{\textbf{TLD}}  & \textbf{Country or}\\
     & \textbf{documents} & & \textbf{Territory}\\
    \midrule
     \rowcolor{gray!20} 
     \multicolumn{4}{c}{One geographic TLD} \\
    \midrule
        Manipuri & 80\% & .in & India \\
        Lithuanian & 79\% & .lt & Lithuania \\
        Polish &  77\% & .pl & Poland \\
        Hungarian & 76\% & .hu & Hungary \\
        Danish & 76\% & .dk & Denmark \\
        Icelandic & 73\% & .is & Iceland \\
        Faroese & 73\% & .fo & Faroe Islands \\
        Macedonian & 73\% & .mk & N. Macedonia \\
        Latgalian &  73\% & .lv & Latvia \\
        Latvian & 72\% & .lv & Latvia\\ 
        % Georgian: 70% .ge (Georgia)
        % Estonian: 67% .ee (Estonia)
        % Norwegian Bokmal: 65% .no (Norway)
        % Bashkir: 64% .ru + рф (Russian Federation)
        % North Azerbaijani: 63% .az (Azerbaijan)
        % Slovene: 61% .si (Slovenia)
        % Finnish: 61% .fi (Finland)
        % Hebrew: 61% .il (Israel)
        % Norwegian Nynorsk: 56% .no
        % Halh Mongolian: 54% .mn (Mongolia)
    \midrule
    \rowcolor{gray!20} 
    \multicolumn{4}{c}{Related territories} \\
    \midrule
        Slovak & 77\% & .sk & Slovakia \\
                & 3\% & .cz & Czechia \\
        Kazakh & 71\% & .kz & Kazakhstan \\
                & 3\% & .ru & Russia\\
        Russian & 65\% & .ru & Russia \\
                & 5\% & .ua  & Ukraine \\
                & 2\% & .by & Belarus \\
        Croatian & 47\% & .hr & Croatia \\
                & 3\% & .ba & Bosnia \\
                & 3\% & .rs & Serbia \\
        Kyrgyz & 33\% & .kg & Kyrgyzstan \\
                & 7\% & .ru & Russia \\            
        Bosnian & 29\% & .rs & Serbia \\
                & 13\% & .ba & Bosnia \\                
       
    \midrule
    \rowcolor{gray!20} 
    \multicolumn{4}{c}{Language variants} \\
    \midrule
        Romanian & 72\% & .ro & Romania \\
                & 3\% & .md & Moldova \\
        Dutch & 66\% & .nl & Netherlands \\
                & 12\% & .be & Belgium \\
        German & 60\% & .de & Germany \\
                & 6\% & .at & Austria \\
                & 5\% & .ch & Switzerland \\
        Portuguese & 45\% & .br & Brazil \\
                & 9\% & .pt & Portugal \\
        Lombard & 47\% & .ch & Switzerland \\
                & 5\% & .it & Italy \\
        Uyghur & 36\% & .cn & China \\
                & 3\% & .kz & Kazakhstan \\
        French & 30\% & .fr & France \\
                & 3\% & .be & Belgium \\
                & 2\% & .ca & Canada \\
                & 2\% & .ch & Switzerland \\
        Spanish & 15\% & .es & Spain \\
                & 4\% & .ar & Argentina \\
                & 4\% & .mx & Mexico \\
                & 2\% & .cl & Chile \\
                & 1\% & .pe & Peru \\
    \bottomrule
    \end{tabular}}
    \caption{Frequent geographic TLDs in monolingual \HPLT{} datasets for different languages.}
    \label{tab:mono_tlds}
\end{table}
%Table format copied from mono_datasets_table.tex  

\begin{table}[tp]
    \centering
    \adjustbox{max width=\linewidth}{
    \footnotesize
    \begin{tabular}{p{1.7cm}rcl}
    \toprule
    \multirow{2}{*}{\textbf{Language}}  & \textbf{\% of} & \multirow{2}{*}{\textbf{TLD}}  & \textbf{Country or}\\
     & \textbf{segments} & & \textbf{Territory}\\
    \midrule
     \rowcolor{gray!20} 
     \multicolumn{4}{c}{One geographic TLD} \\
    \midrule
        Norwegian Nynorsk & 36\% & .no & Norway \\
        Norwegian Bokmål & 34\% & .no & Norway  \\
        Azerbaijani & 33\% & .az & Azerbaijan \\
        Macedonian & 25\% & .mk & North Macedonia \\
        Vietnamese & 23\% & .vn & Vietnam \\
        Farsi & 22\% & .ir & Iran \\
        Hebrew & 20\% & .il & Israel \\
        Sinhala & 19\% & .lk & Sri Lanka \\
        Serbian & 16\% & .rs & Serbia \\
        Malay & 15\% & .my & Malaysia \\
        Hindi & 15\% & .in & India \\
        Japanese & 15\% & .jp & Japan \\
        Korean & 15\% & .kr & South Korea \\
    \midrule
    \rowcolor{gray!20} 
    \multicolumn{4}{c}{Related territories (European Union)} \\
    \midrule
        Maltese & 68\% & .eu & European Union \\
                & 4\% & .mt & Malta \\
        Slovene & 31\% & .si  & Slovenia \\
                & 17\% & .eu & European Union \\
        Estonian & 35\% & .ee & Estonia \\
                & 16\% & .eu & European Union\\
        Latvian & 30\% & .lv & Latvia \\
                & 16\% & .eu & European Union \\
        Lithuanian & 35\% & .lt & Lithuania \\
                & 14\% & .eu & European Union \\
        Slovak & 44\% & .sk & Slovakia \\
                & 11\% & .eu & European Union \\
                & 4\% & .cz & Czechia \\
        Croatian & 26\% & .hr & Croatia \\
                & 10\% & .eu & European Union \\
                & 2\% & .ba & Bosnia \\
        Bulgarian & 26\% & .bg & Bulgaria \\
                &  10\% & .eu & European Union \\
        Irish & 20\% & .ie & Ireland \\
                & 10\% & .eu & European Union \\
        Finnish & 44\% & .fi & Finland \\
                & 7\% & .eu & European Union \\
        
    %Not adding the "multiple territories - others" in this table because I don't think they are super relevant (figures are low, etc)
    %English - Welsh: 47% .uk (United Kingdom) + 8% .cymru (Wales) + 6% .wales (Wales)
    %English - Kazakh: 47% .kz (Kazakhstan) + 5% .ru (Russia)
    %English - Ukrainian: 29% .ua (Ukrania) + 2% .ru (Russia)
    %English- Bosnian: 23% .rs (Serbia) + 7% .ba (Bosnia) + 2% .gr (Greece)
    %English -  Uzbek: 20% .uz (Uzbekistan) + 8% .ru (Russia)
    %English - Catalan: 16% .cat (Catalonia) + 9% .es (Spain) + 1% .ad (Andorra)
    %English - Tamil: 8% .in (India) + 3% .lk (Sri Lanka)
    %English - Arabic: 3% .ae (United Arab Emirates) + 1% .eg (Egypt) + 1% .sa (Saudi Arabia) + 1% .ma (Morocco)
    \bottomrule
    \end{tabular}}
    \caption{Frequent geographic TLDs in the parallel \HPLT{} datasets for different languages (non-English side).}
    \label{tab:para_tlds}
\end{table}

\section{Manual quality inspection}
\label{sec:appendix_r2inspection}
\begin{table}[htbp]
\footnotesize
\begin{tabular}{p{1.8cm}ccc}
\toprule
\textbf{Language} & \textbf{\% Porn $\downarrow$} & \textbf{\% Unnat. $\downarrow$} & \textbf{\% LID $\uparrow$} \\
\midrule
Arabic & 0 (-) & 9 (5-13) & 100 (-) \\ 
Asturian & 0 (-) & 28 (22-35) & 69 (62-75) \\ 
Bengali & 1 (-) & 0 (-) & 100 (-) \\ 
Catalan & 0 (-) & 14 (9-19) & 99 (-) \\ 
Czech & 0 (-) & 9 (4-13) & 100 (-) \\ 
Chinese & 0 (-) & 25 (18-31) & 99 (-) \\
Dutch & 1 (-) & 5 (-) & 100 (-) \\ 
English & 1 (-) & 13 (8-18) & 100 (-) \\ 
Finnish & 1 (-) & 4 (-) & 100 (-) \\ 
German & 1 (-) & 2 (-) & 98 (-) \\ 
Hindi & 2 (-) & 2 (-) & 98 (-) \\ 
Iran. Persian & 0 (-) & 25 (18-31) & 99 (-) \\ 
Marathi & 0 (-) & 6 (-) & 97 (-) \\ 
Modern Greek & 0 (-) & 3 (-) & 100 (-) \\ 
Nor. Bokmål & 2 (-) & 8 (4-11) & 99 (-) \\ 
Nor. Nynorsk & 0 (-) & 3 (-) & 93 (-) \\ 
Polish & 1 (-) & 7 (3-11) & 100 (-) \\ 
Russian & 2 (1-3) & 18 (15-21) & 98 (-) \\ 
Scot. Gaelic & 0 (-) & 3 (-) & 89 (85-93) \\ 
Slovak & 0 (-) & 10 (6-14) & 100 (-) \\ 
Spanish & 1 (-) & 9 (5-13) & 100 (-) \\ 
Turkish & 6 (-) & 10 (5-14) & 99 (-) \\ 
\bottomrule
\end{tabular}

\caption{Manual quality inspection of a random sample of documents from the cleaned version, stratified by crawls groups. Percentages of extracted texts considered as pornography (\% Porn), unnatural texts (\% Unnat.), and texts correctly classified by language identification (\% LID) (the 95\% confidence intervals for the percentage estimates are given in brackets when applicable).}
\label{tab:r2inspection}
\end{table}
In this section, we study how the quality of the extracted texts varies between older and newer crawls, and also between IA and CC crawls. More specifically, for a particular language we wanted to understand if there are any substantial differences in the proportions of texts classified as this language by mistake or just undesirable texts. 

For this study, we carried out manual annotation of documents from the cleaned version of our dataset asking our annotators to provide three binary annotations for each document.
\begin{compactitem}
    \item \textbf{LID ok:} 0 if most of the text is not in the target language, otherwise 1;
    \item \textbf{Unnatural:} 1 if most of the text looks unnatural (e.g. word lists for SEO, mostly boilerplate, etc.), otherwise leave empty;
    \item \textbf{Porn:} 1 if the text looks like pornographic content, otherwise leave empty.
\end{compactitem}

We compared four groups of crawls: among wide IA crawls and CC crawls separately we selected old crawls from 2012-2014 and new crawls from 2017-2020. Among languages spoken by the paper authors, 22 languages were selected for annotation. 

\begin{figure}[ht]
    \centering % Center the image
    \includegraphics[width=0.45\textwidth]{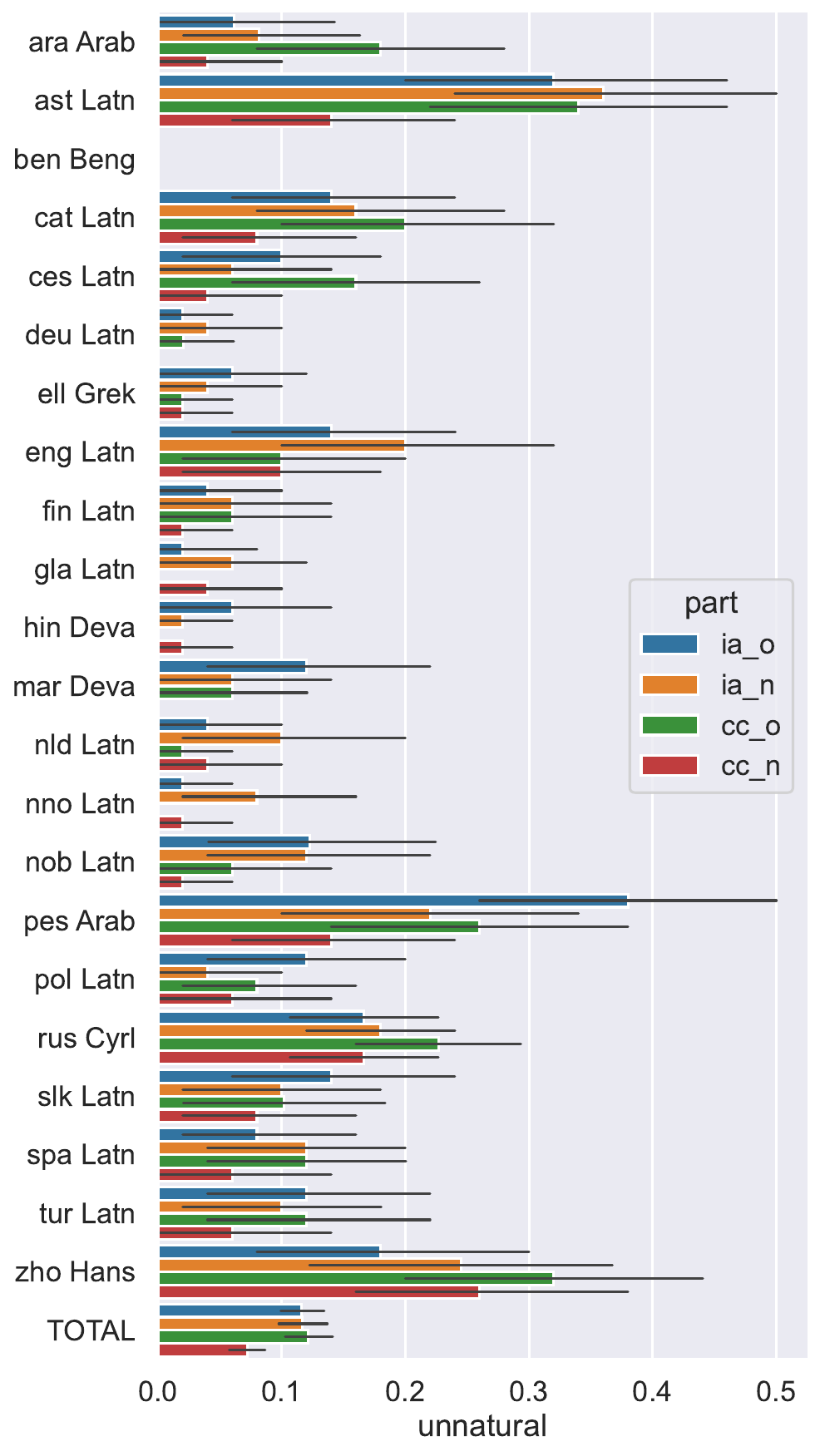} 
    \caption{Proportions of unnatural texts among the cleaned texts extracted from four selected groups of crawls, according to manual inspection of a sample. Error bars correspond to the 95\% confidence intervals.} 
    \label{fig:props-unnatural-texts}
\end{figure}

For each language and each group of crawls, 50 random documents from the cleaned version of our datasets were annotated by a native or a fluent speaker of this language. In total, 200 documents for each language were annotated, except for Russian where three native speakers annotated 600 documents. Only texts extracted from the documents were shown to the annotators, they did not know which crawl each text came from or any other meta-information. For documents longer than 1000 characters, the first 500 characters and 500 characters from the beginning of the second half were shown.

\Cref{tab:r2inspection} shows the results for the four groups combined together.\footnote{Since the sample is stratified by group and the crawls from these groups give about 52\% of all texts in our dataset, one should carefully interpret these statistics in the context of the full dataset.} We see that the proportion of pornographic content is low, usually between 0-2\% with a maximum of 6\% for Turkish. The precision of our LID model for the inspected languages is above 97\%, with a few notable exceptions. The worst precision is for Asturian where we observed about 30\% of texts being in Spanish or other Spanish minority languages (e.g. Extremeño, Aragonese), or just SEO lists consisting of e.g. song names not in Asturian. The proportions of unnatural texts vary a lot from language to language. Annotators report the following major types of unnaturalness: lists of services and goods, commercial ads with varying degrees of grammaticality, traces of Wikipedia markup, documents consisting mostly of menus, and boilerplate missed by boilerplate removal.

\Cref{fig:props-unnatural-texts} shows proportions of unnatural texts for each language and group of crawls. Looking at individual languages, for most of them the group of new CC crawls give a much lower proportion of unnatural texts than other groups. However, since only 50 documents were labelled from each group and language, the confidence intervals are large and statistically significant conclusions cannot be made for each individual language. However, when annotations for all languages are combined (denoted as TOTAL on the figure) it becomes clear that for a random language (among those annotated) a random document has about 2x lower probability to be unnatural if it comes from the group of newer CC crawls compared to older CC crawls or any of two groups of IA crawls. For the proportions of pornographic content and documents misclassified by \ac{lid} we did not observe any consistent differences for different groups of crawls.

\section{Model training and evaluation}
\label{sec:appendix_LLMtrainingEval}
\subsection{Corpora comparison: English}
\label{sec:appendix_english}

\paragraph{Pretraining} We fully replicated the original FineWeb training and evaluation setup by \citet{fineweb}, with the same architecture and pretraining settings (1.71B parameters, Llama architecture with a sequence length of 2048 tokens, GPT 2 tokenizer, and a global batch size of \textasciitilde2 million tokens). We train 4 models that are differentiated only by training data, and evaluate their performance at different stages of model training. Each model is trained on 100 billion tokens, randomly sampled from the following datasets:
\begin{compactitem}
    \item English \HPLT{} data, cleaned
    \item English \HPLT{} data, deduplicated
    \item English HPLT~v1.2 \cite{de-gibert-etal-2024-new} %\cite{aulamo-etal-2023-hplt}
    \item FineWeb dataset \cite{fineweb}
\end{compactitem}

\noindent We use NVIDIA's \texttt{Megatron-LM} (\url{https://github.com/NVIDIA/Megatron-LM}) training framework instead of HuggingFace's \texttt{nanotron} (\url{https://github.com/huggingface/nanotron}) framework used by \citet{fineweb}. Each model is trained on the \LUMI{} supercomputer with 16 nodes, each with 4 AMD MI250x GPUs with dual-GCD (graphics compute die) design, amounting to 8 logical devices. In total, we used 128 devices and a single 64-core CPU for approximately 84 hours, totalling 11 008 GPU hours per model.

% Andrey said that this can be removed.
%located in Finland. The LUMI-G partition has 2978 nodes, with each node having four AMD MI250x GPUs with 128GB of memory each, and a single 64-core CPU. The MI250x is a multi-chip module (MCM), with dual-GCD (graphics compute die) design, which in practice means a node has ight logical devices, each logical device with access to 64GB of high bandwidth memory. Each node has four 200Gbps Slingshot-11 network interconnects. The nodes are connectedtogether in a dragonfly topology.

\begin{figure*}[t]
    \centering % Center the image
    \includegraphics[width=0.7\textwidth]{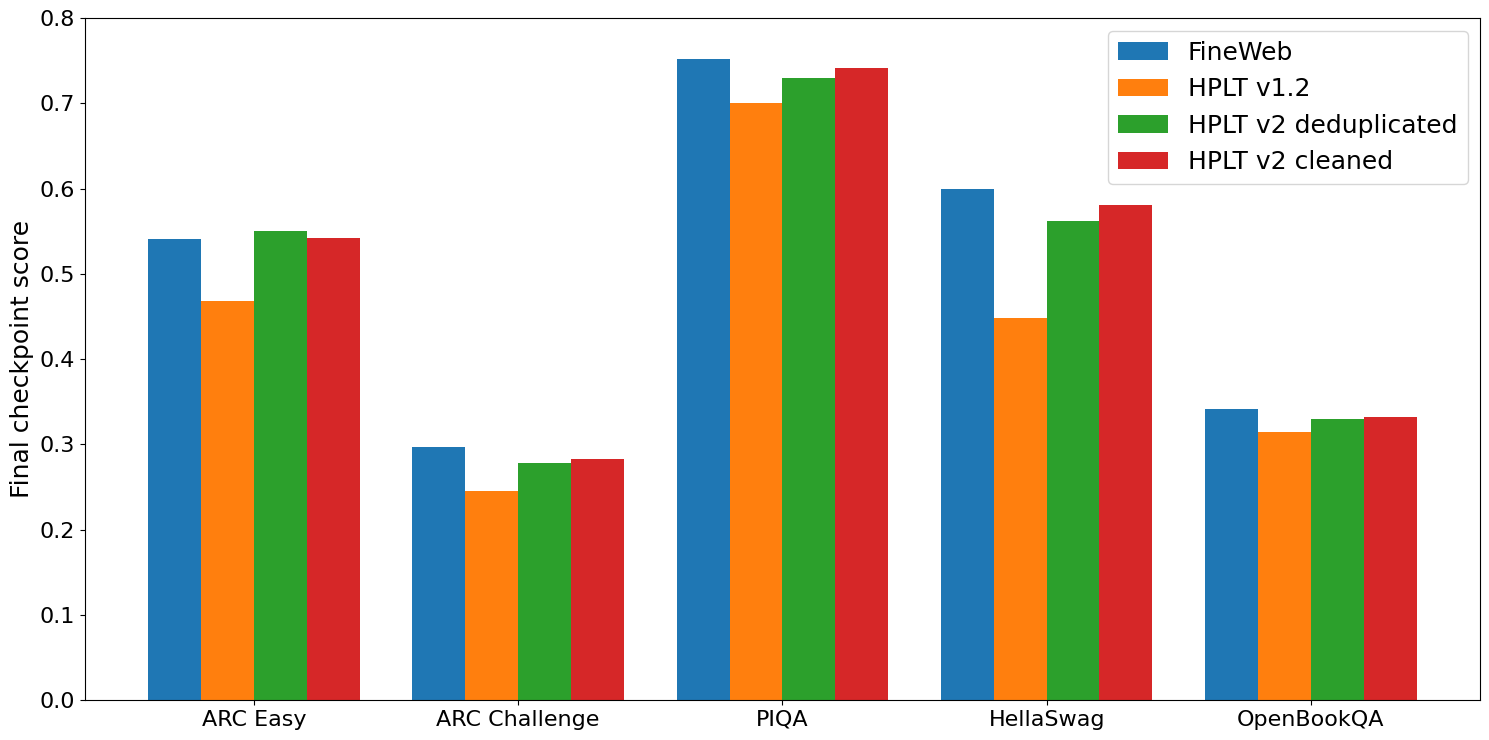} 
    \caption{Final checkpoint scores of the English models trained on the datasets shown, grouped based on the benchmarks conducted. The models perform quite similarly with the exception of the model trained on the HPLT~v1.2 dataset, the scores of which are noticeably lower.}
    \label{fig:LLM-results-final-scores}
\end{figure*}

\paragraph{Evaluation} Evaluation is performed using HuggingFace's \texttt{LightEval} tool \cite{lighteval} on the tasks listed below. Results per task are presented in  Figure \ref{fig:LLM-results-final-scores}.
%in \url{https://huggingface.co/datasets/HuggingFaceFW/fineweb/blob/main/lighteval_tasks.py} by \citet{fineweb}. The tasks are
\begin{compactitem}
    \item \textbf{HellaSwag}: a dataset to evaluate commonsense reasoning. Its questions are designed to be trivial for humans but challenging for LLMs \cite{zellers-etal-2019-hellaswag}.
    %\item WinogGande \cite{sakaguchi-2021-winogrande}
    \item \textbf{PIQA}: a dataset focusing on reasoning with multiple-choice questions about physical interactions, evaluating the LLM's understanding of how different objects are used in various situations \cite{Bisk2020piqa}.
    %\item SIQA \cite{sap-etal-2019-social}
    \item \textbf{OpenBookQA}: a dataset consisting of multiple-choice questions which require understanding concepts and their relations, benchmarking the complex reasoning and inference performance of the LLM \cite{OpenBookQA2018}.
    \item \textbf{ARC Easy and ARC Challenge}: both parts of the AI2 Reasoning Challenge dataset, containing easier and more complex questions to test the LLM's reasoning skills \cite{clark-2018-ARC}.
    %\item CommonsenseQA \cite{talmor-etal-2019-commonsenseqa}
    %\item MMLU \cite{hendrycks2021ethics,hendryckstest2021}
\end{compactitem}

\subsection{Corpora comparison: Norwegian}
% from Vlad: David, please add pretraining details similar to I.1
\paragraph{Pretraining} We mirrored the pretraining setup used for the English ablation studies in \Cref{sec:appendix_english}, except for two details: 1) we trained a new tokenizer specifically for Norwegian, using a single tokenizer for all experiments trained on equal number of samples from all ablated corpora using the \href{https://github.com/huggingface/tokenizers}{\texttt{tokenizers}} library; 2) we pretrained the models for 30B tokens (roughly corresponding to 1 epoch on most of the ablated corpora) instead of 100B, mirroring the multilingual experiments for FineTasks \citep{kydlicek2024finetasksmultilingualtasks}.

We compared five different filtered corpora that support Norwegian. Most of these discriminate between two written variants of Norwegian -- Bokmål and Nynorsk -- in those cases, we simply concatenate these subcorpora. The ablated corpora are:

\begin{compactitem}
    \item Norwegian \HPLT{} data, cleaned;
    \item Norwegian CulturaX \citep{culturax};
    \item Norwegian HPLT~v1.2 \citep{de-gibert-etal-2024-new};
    \item Norwegian FineWeb-2 \citep{penedo2024fineweb-2};
    \item Norwegian mC4 \cite{xue-etal-2021-mt5}.
\end{compactitem}

\noindent The pretraining code is built on the Megatron-DeepSpeed framework \citep{smith2022usingdeepspeedmegatrontrain}. All models were trained on the LUMI supercomputer using 32 compute nodes, each with 4 AMD MI250x GPUs. The full pretraining run of each model took approximately 15 hours (wall-clock time), or 1\,920 GPU-hours ($15 \times 32 \times 4$ hours), respectively.

\paragraph{Evaluation} Evaluation is performed using \texttt{NorEval} \cite{mikhailov2025noreval}, an open-source benchmark for Norwegian built upon LM Evaluation Harness \cite{eval-harness}. We consider the following ten multiple-choice QA, generative QA, sentence completion, and sentence pair ranking tasks that target different aspects of the model understanding and generation abilities in Norwegian Bokmål and Nynorsk:
%\footnote{Norwegian has two official written standards, which differ from one another at the lexical level: Bokmål and Nynorsk.}

\begin{compactitem}
    \item \textbf{Commonsense reasoning:} performing logical and commonsense reasoning (NorCommonsenseQA, \citealp{mikhailov-etal-2025-collection}).
    \item \textbf{Norwegian-specific \& world knowledge:} answering questions about facts and Norwegian culture (NorOpenBookQA and NRK-Quiz-QA, \citealp{mikhailov-etal-2025-collection}).
    \item \textbf{Norwegian language knowledge:} understanding Norwegian punctuation rules (NCB, \citealp{mikhailov2025noreval})\footnote{\url{https://huggingface.co/datasets/hcfa/ncb}} and idioms (NorIdiom, \citealp{mikhailov2025noreval}).\footnote{\url{https://huggingface.co/datasets/Sprakbanken/Norwegian_idioms}}
    \item \textbf{Machine reading comprehension:} understanding a given text and extracting an answer from it (NorQuAD, \citealp{ivanova-etal-2023-norquad}).
\end{compactitem}

\noindent We aim to find tasks that provide a reliable signal during pretraining. We evaluate the models in a zero-shot regime at regular checkpoint intervals (approx. 1B tokens) on all tasks. Next, we discard tasks that provide a low signal based on two criteria \cite{penedo2024fineweb-2}: 

\begin{compactitem}
    \item \textbf{Monotonicity:} the Spearman rank correlation between the number of steps and the target performance score is at least 0.5 across all model checkpoints. 
    \item \textbf{Non-random performance:} the difference between the random baseline (zero for generative tasks, one divided by the number of answer choices for multiple-choice tasks, and a coin flip probability for sentence pair ranking tasks) and the maximum score across all models is positive and satisfactory. 
\end{compactitem}

\begin{figure*}[t]
    \centering 
    \includegraphics[width=0.7\textwidth]{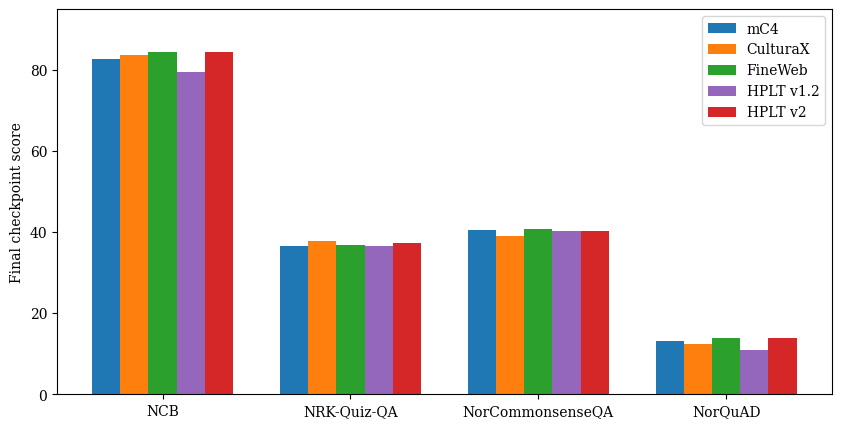} 
    \caption{Final checkpoint scores of the Norwegian models trained on the datasets shown, grouped based on the evaluation datasets in \texttt{NorEval}. The models perform quite similarly with the exception of the model trained on the HPLT~v1.2 dataset, the NorQuAD and NCB scores of which are generally lower.}
    \label{fig:LLM-results-final-scores-norwegian}
\end{figure*}

\noindent The filtering results in four datasets: NCB (accuracy), NRK-Quiz-QA Bokmål (accuracy), NorCommonsenseQA Bokmål (accuracy), and NorQuAD (F1-score). We aggregate the performance across the datasets using the average normalized score \cite{myrzakhan2024openllmleaderboard}. We report the performance results for our 150 checkpoints in Figure \ref{fig:LLM-results-norwegian} (see \S\ref{subsec:nlu_evaluation_ablations}) and final checkpoint performance in Figure \ref{fig:LLM-results-final-scores-norwegian}.

\section{LTG-BERTs training and evaluation details}
\label{sec:appendix_LTG-BERT}

Following the HPLT~v1.2,\footnote{\url{https://hplt-project.org/HPLT\_D4\_1\_\_\_First\_language\_models\_trained.pdf}} we use UD treebanks of version 2.13\footnote{\url{https://lindat.mff.cuni.cz/repository/xmlui/handle/11234/1-5287}} for  most languages, except for Albanian and Georgian. These languages were not used in the HPLT~v1.2 report due to missing training and development splits in UD 2.13. However, UD 2.15 does contain the required splits, and we use them. We do not evaluate NER on Maltese, since its WikiAnn training split contains only 100 samples. Table~\ref{tab:mlm} shows detailed MLM evaluation results by language and task.

LTG-BERT architecture \cite{samuel-etal-2023-trained} is a version of the original masked BERT model \cite{devlin-etal-2019-bert}. The differences include
removing the next sentence prediction objective, swapping subword masking to span masking, and other minor architectural improvements. LTG-BERT was shown to perform well for small-sized training datasets \cite{samuel-2023-mean}, which fits our evaluation setup. The models were trained with the same hyperparameters as in the aforementioned HPLT report.

We trained separate models for Bosnian and Croatian, in addition to the joint Bosnian-Croatian model. Since the UD does not provide Bosnian treebanks, we evaluated all three models on the Croatian datasets. We did not include Serbian, because it uses the Cyrillic writing system in \HPLT{}, while UD features Serbian data only in Latin. Exploring whether mixing the scripts still improves the results is left for future work. It is difficult to give any clear recommendations on which of the three models to use for practical tasks, since all of them yield satisfactory evaluation results (ranking varies from task to task). 

LTG-BERT models were trained for 31 250 steps on 16 compute nodes with 4 physical AMD INSTINCT MI250x GPUs each for approximately 9.8 hours.
Sharding, training a tokenizer and tokenizing for larger languages required up to 3.5, 0.5 and 1 hours correspondingly on 7 AMD EPYC 7763 CPUs  (these numbers are estimated from the processing of English, the largest data subset in \HPLT{}. 
%For languages with more than one archive of data available, we processed only a first one. 
The processing time of different languages may vary, for instance, languages without whitespace separation between words require an additional pretokenizing step).
UD fine-tuning and NER fine-tuning required 1.1 hours and 8 minutes correspondingly on 1 GPU (estimated for English).

\begin{table*}[!htp]
\centering
\small
\adjustbox{max width=\linewidth}{
\begin{tabular}{lcccccccccccccccc}
\toprule
\textbf{} &\multicolumn{4}{c}{\textbf{POS tags}} &\multicolumn{4}{c}{\textbf{Lemmas}} &\multicolumn{4}{c}{\textbf{Dependency parsing}} &\multicolumn{4}{c}{\textbf{NER}} \\
\cmidrule(lr){2-5} \cmidrule(lr){6-9} \cmidrule(lr){10-13} \cmidrule(lr){14-17}
\multirow{2}{*}{\textbf{Language}} & \multirow{2}{*}{mBERT} & \multirow{2}{*}{XLM-R} & HPLT & \multirow{2}{*}{\HPLT{}} & \multirow{2}{*}{mBERT} & \multirow{2}{*}{XLM-R} & HPLT & \multirow{2}{*}{\HPLT{}}  & \multirow{2}{*}{mBERT} & \multirow{2}{*}{XLM-R} & HPLT & \multirow{2}{*}{\HPLT{}}  & \multirow{2}{*}{mBERT} & \multirow{2}{*}{XLM-R} & HPLT & \multirow{2}{*}{\HPLT{}}\\
&  & & v1.2 &  & & & v1.2 & & & & v1.2 & & & & v1.2 & \\
\midrule
als\_Latn & 59.1 & 61.6 & 64.0 & \textbf{64.5} & \textbf{78.2} & 75.0 & 76.3 & 77.2 & \textbf{33.1} & 29.3 & 25.3 & 24.7 & 92.3 & 92.9 & 92.4 & \textbf{93.9} \\
bel\_Cyrl & 94.1 & 94.6 & 95.5 & \textbf{95.7} & 93.2 & 93.8 & 93.8 & \textbf{97.1} & 88.1 & 89.9 & 91.1 & \textbf{91.7} & 91.7 & 90.3 & 90.1 & \textbf{92.8} \\
bos\_Latn & 95.5 & 96.2 & 96.4 & \textbf{96.6} & 97.2 & \textbf{97.4} & 97.2 & 97.1 & 90.2 & 91.3 & 91.3 & \textbf{91.7} & 91.5 & 91.6 & 89.3 & \textbf{92.8} \\
hrv\_Latn  & 95.5 & 96.2 & 96.4 &  \textbf{96.8} & 97.2 & \textbf{97.4} & 97.2 & 97.1 & 90.2 & 91.3 & 91.3 & \textbf{91.6} & 91.5 & 91.6 & 89.3 & \textbf{92.5} \\
bul\_Cyrl & 97.0 & 97.5 & 97.8 & \textbf{97.9} & 97.5 & \textbf{97.7} & 97.3 & 97.3 & 92.7 & 94.4 & 94.0 & \textbf{94.5} & 92.2 & 92.2 & 91.5 & \textbf{93.0} \\
cat\_Latn & 97.1 & 97.2 & 97.4 & \textbf{97.5} & 99.4 & 99.4 & 99.4 & \textbf{97.5} & 93.6 & 94.1 & 94.4 & \textbf{99.4} & 92.1 & 91.0 & 90.1 & \textbf{94.5} \\ 
ces\_Latn & 97.8 & 98.0 & 98.3 & \textbf{98.4} & 99.3 & 99.3 & 99.4 & 99.4 & 93.5 & 94.2 & 94.4 & \textbf{94.6} & 91.2 & 91.2 & 89.0 & \textbf{91.8} \\
cym\_Latn & 87.2 & 88.3 & \textbf{89.2} & 89.0 & \textbf{94.6} & 94.4 & 93.7 & 92.3 & 80.8 & \textbf{82.8} & 82.3 & \textbf{82.8} & 92.5 & 90.0 & 89.4 & \textbf{93.4} \\
dan\_Latn & 96.7 & 97.8 & 97.8 & \textbf{97.9} & 97.2 & \textbf{97.6} & 97.1 & 97.1 & 86.7 & 89.1 & 88.8 & \textbf{89.5} & 91.2 & 91.6 & 90.3 & 92.0 \\
deu\_Latn & 88.8 & 89.4 & 80.7 & \textbf{89.9} & 97.6 & \textbf{97.7} & 95.5 & 97.5 & 84.6 & 87.1 & 76.4 & \textbf{87.6} & \textbf{89.4} & 87.7 & 64.1 & 89.2 \\
ell\_Grek  & 94.6 & 95.7 & 96.1 & \textbf{96.2} & 94.6 & \textbf{94.7} & 94.1 & 94.1 & 91.7 & \textbf{93.5} & 92.2 & 93.2 & 90.2 & 90.7 & 90.2 & \textbf{92.6} \\
eng\_Latn  & 96.1 & 96.8 & 96.7 & \textbf{97.0} & 97.8 & 98.0 & 97.9 & \textbf{98.1} & 91.3 & 92.6 & 92.2 & \textbf{93.0} & 2.2 & 81.1 & 81.0 & \textbf{82.7} \\
spa\_Latn  & 95.7 & 95.9 & 96.0 & \textbf{96.2} & \textbf{99.4} & \textbf{99.4} & \textbf{99.4} & 99.4 & 92.3 & 93.0 & 93.1 & \textbf{93.4} & \textbf{90.9} & 89.9 & 89.6 & 90.8 \\
est\_Latn & 96.0 & 96.6 & \textbf{97.1} & \textbf{97.1} & 94.8 & 95.0 & \textbf{95.2} & \textbf{95.2} & 88.1 & 89.7 & 90.8 & \textbf{91.0} & 91.8 & 90.4 & 89.6 & \textbf{93.0} \\
eus\_Latn & 91.0 & 91.4 & \textbf{92.3} & \textbf{92.3} & 95.7 & 95.9 & \textbf{96.0} & 95.9 & 85.3 & 87.3 & 88.1 & \textbf{88.2} & 91.3 & 90.7 & 89.8 & \textbf{92.9} \\
pes\_Arab & 95.9 & 96.3 & \textbf{96.4} & 96.3 & 99.1 & 99.4 & 99.4 & \textbf{99.5} & 92.7 & 93.8 & 93.9 & \textbf{94.1} & 92.0 & 92.9 & 91.8 & \textbf{93.9} \\
fin\_Latn & 95.1 & 96.4 & 96.8 & \textbf{97.0} & 90.6 & 91.5 & \textbf{91.6} & 91.4 & 90.2 & 93.0 & 93.3 & \textbf{94.0} & 90.2 & 90.0 & 89.2 & \textbf{91.6} \\
fra\_Latn & 97.8 & \textbf{98.1} & \textbf{98.1} & 98.0 & 98.6 & \textbf{98.8} & 93.8 & 98.6 & 93.8 & 94.4 & 94.5 & \textbf{94.8} & \textbf{90.5} & 88.7 & 87.2 &90.0 \\
gle\_Latn & 86.5 & 87.1 & 88.7 & \textbf{89.3} & 95.5 & 95.8 & \textbf{96.1} & 95.6 & 81.3 & 82.7 & 83.4 & \textbf{84.3} & \textbf{80.8} & 78.0 & 55.9 & 78.2 \\
glg\_Latn & 96.9 & \textbf{97.1} & \textbf{97.1} & 97.0 & \textbf{98.3} & \textbf{98.3} & 98.2 & 98.0 & 82.3 & \textbf{82.6} & 82.3 & 82.2 & 92.5 & 93.3 & 91.1 & \textbf{94.1} \\
heb\_Hebr & 95.6 & 96.1 & 96.5 & \textbf{96.7} & 97.0 & \textbf{97.2} & 97.1 & \textbf{97.2} & 89.8 & 91.6 & 91.0 & \textbf{91.9} & 2.6 & 84.2 & 88.4 & \textbf{89.3} \\
hin\_Deva & 92.4 & 93.3 & 93.6 & \textbf{93.7} & 98.9 & \textbf{99.0} & \textbf{99.0} & \textbf{99.0} & 92.6 & 93.3 & 93.5 & \textbf{93.6} & 88.6 & 88.0 & 84.3 & \textbf{89.5} \\
hrv\_Latn  & 95.5 & 96.2 &  96.4 &  \textbf{96.7} & 97.2 & \textbf{97.4} & 97.2 & 97.2 & 90.2 & 91.3 & 91.3 & \textbf{91.8} & 91.5 & 91.6 & 89.3 & \textbf{92.0} \\
hun\_Latn & 93.0 & \textbf{94.3} &  93.0 &  94.1 & 93.0 & \textbf{94.3} & 93.0 & 92.3 & 84.3 & \textbf{86.7} & 82.4 & 86.1 & 92.2 & 91.9 & 92.8 & \textbf{93.1} \\
hye\_Armn  & 88.7 & 91.2 &  \textbf{92.7} & \textbf{92.7} & 94.4 & \textbf{94.9} & 93.9 & 94.7 & 80.4 & 85.3 & 84.1 & \textbf{86.8} & 95.7 & 95.3 & 94.8 & \textbf{95.9} \\
ind\_Latn & 89.5 & \textbf{89.8} &  89.6 & 89.1 & 98.2 & \textbf{98.3} & 98.0 & 97.5 & 82.4 & \textbf{82.7} & 81.7 & 81.8 & 91.3 & 91.6 & 89.1 & \textbf{92.0} \\
isl\_Latn & 87.7 & 88.1 & 88.6 & \textbf{88.7} & 96.2 & 96.4 & \textbf{96.5} & 96.4 & 85.2 & 86.6 & 86.9 & \textbf{87.4} & \textbf{81.7} & 63.9 & 55.9 & 78.3 \\
ita\_Latn & 98.0 & 98.0 & 98.1 & \textbf{98.3} & 98.6 & 98.7 & \textbf{98.8} & 98.7 & 94.1 & 94.4 & 94.6 & \textbf{95.1} & 90.5 & 89.7 & 87.8 & \textbf{91.2} \\
jpn\_Jpan & 97.5 & 97.7 & \textbf{97.8} & \textbf{97.8} & 98.3 & 98.3 & 98.3 & \textbf{98.4} & 94.1 & 94.6 & 94.6 & \textbf{94.8} & 66.5 & 65.9 & \textbf{67.4} & 67.2 \\
kat\_Geor & 91.3 & \textbf{92.6} & 92.4 & 92.4 & \textbf{92.8} & 93.7 & 92.5 & 92.5 & 79.5 & 80.9 & 80.8 & \textbf{81.3} & 87.2 &  4.7 & 89.6 & \textbf{90.7} \\
kor\_Hang & 88.6 & 89.7 & 89.9 & \textbf{90.1} & 94.0 & 94.3 & \textbf{94.4} & 94.4 & 88.0 & 89.0 & 89.4 & \textbf{89.7} & 87.8 & 87.0 & 88.3 & \textbf{89.3} \\
lvs\_Latn & 91.6 & 92.8 & 92.4 & \textbf{93.6} & 96.9 & 91.6 & 96.8 & \textbf{97.7} & 88.8 & 90.9 & 90.9 & \textbf{92.1} & 93.2 & 92.6 & 90.7 & \textbf{93.9} \\
lit\_Latn & 87.7 & 91.9 & 92.0 & \textbf{92.5} & 90.2 & \textbf{91.6} & 91.5 & 91.2 & 79.3 & 85.7 & 84.9 & \textbf{86.8} & 89.1 & 89.3 & 87.0 & \textbf{91.0} \\
ltz\_Latn & - & - & - & - & - & - & - & - & - & - & - & - & - & - & - & 89.2 \\
mkd\_Cyrl & - & - & - & - & - & - & - & - & - & - & - & - & - & - & - & 94.6 \\
mlt\_Latn & 94.7 & 94.5 & 97.0 & \textbf{97.7} & \textbf{100.0} & \textbf{100.0} & \textbf{100.0} & \textbf{100.0} & 78.2 & 78.5 & 83.2 & \textbf{87.2} & - & - & - & - \\
nob\_Latn & 97.0 & 97.4 & \textbf{97.6} & 97.5 & 98.5 & \textbf{98.8} & \textbf{98.8} & 98.7 & 93.2 & 94.3 & 94.5 & \textbf{94.7} & 91.9 & 92.6 & 91.1 & \textbf{93.2} \\
nld\_Latn & 96.2 & 96.9 & 97.1 & \textbf{97.2} & 94.1 & \textbf{94.7} & 94.4 & 94.1 & 91.6 & 92.9 & 93.8 & \textbf{94.1} & \textbf{91.7} & 90.4 & 88.6 & 91.0 \\
nno\_Latn & 96.6 & 97.0 & 97.7 & \textbf{97.8} & 98.2 & 98.4 & \textbf{98.5} & \textbf{98.5} & 92.9 & 93.9 & 94.6 & \textbf{95.0} & \textbf{95.8} & 93.6 & 93.2 & 95.5 \\
pol\_Latn & 95.6 & 95.5 & 96.9 & \textbf{97.2} & 97.8 & \textbf{98.2} & \textbf{98.2} & \textbf{98.2} & 93.7 & 95.2 & 95.3 & \textbf{95.6} & 12.9 & 88.8 & \textbf{89.7} & 89.6 \\
por\_Latn & 93.6 & 94.0 & \textbf{94.1} & \textbf{94.1} & 98.1 & \textbf{98.3} & \textbf{98.3} & 98.2 & 83.4 & 84.5 & 84.9 & \textbf{85.3} & 91.2 & 90.3 & 88.0 & \textbf{91.5} \\
ron\_Latn & 97.3 & 97.6 & 97.7 & \textbf{97.9} & 97.7 & \textbf{97.9} & 97.8 & 97.8 & 89.5 & 91.0 & 90.6 & \textbf{91.6} & \textbf{94.5} & 93.6 & 91.2 & 93.6 \\
rus\_Cyrl & 93.8 & 94.4 & 94.5 & \textbf{94.7} & 98.3 & 98.5 & \textbf{98.6} & \textbf{98.6} & 92.6 & 93.4 & 93.6 & \textbf{93.8} & 88.0 & 86.9 & 85.6 & \textbf{89.0} \\
slk\_Latn & 89.1 & 97.6 & \textbf{98.1} & 91.9 & 95.7 & \textbf{96.1} & 95.6 & 95.5 & 92.9 & 94.4 & 93.8 & \textbf{95.0} & 93.2 & 92.9 & 91.2 & \textbf{93.3} \\
slv\_Latn & 96.7 & 97.6 & 98.1 & \textbf{98.2} & 98.5 & \textbf{98.7} & 98.6 & \textbf{98.7} & 93.4 & 94.7 & 94.8 & \textbf{95.3} & 93.4 & 93.1 & 93.6 & \textbf{94.2} \\
srp\_Cyrl & - & - & - & - & - & - & - & - & - & - & - & - & 91.6 & 92.4 & - & \textbf{93.4} \\
swe\_Latn & 96.5 & \textbf{97.4} & \textbf{97.4} & 97.3 & 97.3 & \textbf{97.6} & 97.1 & 97.0 & 89.4 & \textbf{92.1} & 90.8 & 91.7 & 94.3 & \textbf{94.5} & 93.5 & 94.4 \\
tat\_Cyrl & - & - & - & - & - & - & - & - & - & - & - & - & \textbf{89.7} & 80.6 & 82.9 & 84.0 \\
tur\_Latn & 90.4 & 91.0 & \textbf{91.5} & 91.4 & 91.1 & 91.3 & \textbf{91.9} & 91.4 & 70.9 & 73.0 & 73.6 & \textbf{74.6} & 92.2 & 92.0 & 90.8 & \textbf{92.5} \\
ukr\_Cyrl & 93.1 & 94.7 & 72.9 & \textbf{95.3} & 87.0 & \textbf{97.2} & 87.0 & 97.0 & 89.4 & 91.8 & 61.3 & \textbf{92.1} & 92.0 & 91.7 & 77.5 & \textbf{92.8} \\
vie\_Latn & 89.8 & \textbf{92.1} & 91.8 & \textbf{92.1} & \textbf{99.9} & \textbf{99.9} & \textbf{99.9} & \textbf{99.9} & 66.5 & \textbf{70.3} & 68.0 & \textbf{70.3} & \textbf{91.9} & 90.6 & 89.2 & 90.3 \\
zho\_Hans & 96.2 & \textbf{96.3} & 96.0 & 96.0 & \textbf{99.9} & \textbf{99.9} & \textbf{99.9} & \textbf{99.9} & 86.1  & \textbf{86.9} & 84.6 & 85.6 & 0.1 & \textbf{76.5} & 75.5 & 74.5 \\
\bottomrule
\end{tabular}
}
\caption{Results of monolingual masked language models trained on the \HPLT{} datasets compared to the baselines on part-of-speech (POS) tagging, lemmatization, dependency parsing and named entity recognition. For POS tagging, we evaluate the AllTags performance, which is the exact
match accuracy of the UPOS, XPOS and UFeats UDtags. For dependency parsing, we report LAS, and for lemmatization accuracy.}
\label{tab:mlm}
\end{table*}

% Skipped this. Let's get this ready for the camera ready.
% \section{Training and Model Configurations for Translation Models}
% \label{sec:appendix_para_mt_configs}

\section{Full Results for Translation Models Built on Parallel Data}\label{sec:app-full-MT-results}

We compare models trained on \HPLT{}, Tatoeba \citep{tiedemann-2012-parallel,tiedemann-2020-tatoeba}, and the combination of the two datasets.
The language selection is the intersection of the languages covered by both datasets.
We evaluate the models on the FLORES-200 evaluation benchmark \citep{nllb2022} using SacreBLEU implementation of BLEU\footnote{\scriptsize \texttt{nrefs:1|case:mixed|eff:no|smooth:exp|version:2.5.1},\\ and where applicable, \texttt{tok:ja-mecab}, \texttt{tok:ko-mecab}, or \texttt{tok:13a}} and chrF++\footnote{\scriptsize \texttt{nrefs:1|case:mixed|eff:yes|nc:6|nw:0|space:no|version:2.5.1}} metrics \citep{post-2018-call} and COMET-22-DA \citep{rei-etal-2022-comet}.

\Cref{tab:para_mt_eval_from_en,tab:para_mt_eval_into_en} present the full results of the MT models for translation into English and from English respectively.
For reference, we also include the performance of models trained on the HPLT~v1.2 dataset, which shares the same underlying extraction pipeline. 
Note that we did not perform any language-specific hyper-parameter tuning which possibly led to low scores for a few model instances. 

\label{sec:appendix_para_mt_evaluation}
\begin{table*}[!htp]
\centering
\small
\adjustbox{max width=\linewidth}{
\begin{tabular}{lrrrrrrrrrrrr}
\toprule
      & \multicolumn{3}{c}{HPLT~v1.2}                                                    & \multicolumn{3}{c}{\HPLT{}} & \multicolumn{3}{c}{Tatoeba (OPUS)}                                           & \multicolumn{3}{c}{\HPLT{}+OPUS}                                 \\
\cmidrule(lr){2-4}\cmidrule(lr){5-7}\cmidrule(lr){8-10}\cmidrule(lr){11-13}
 & BLEU & chrF++ & COMET & BLEU & chrF++ & COMET & BLEU & chrF++ & COMET & BLEU & chrF++ & COMET \\
\midrule
en-af &                          &                          &                            & 39.2    & 64.5   & 0.8398   & 38.3                 & 63.6                 & 0.8397               & 38.8                 & 63.8                 & 0.8409               \\
en-ar &                          &                          &                            & 26.6    & 55.0   & 0.8442   & \multicolumn{1}{l}{} & \multicolumn{1}{l}{} & \multicolumn{1}{l}{} & \multicolumn{1}{l}{} & \multicolumn{1}{l}{} & \multicolumn{1}{l}{} \\
en-az &                          &                          &                            & 12.2    & 41.0   & 0.8128   & 11.3                 & 38.7                 & 0.8074               & 11.5                 & 38.7                 & 0.8011               \\
en-be &                          &                          &                            & 11.6    & 39.0   & 0.7756   & 11.2                 & 37.4                 & 0.7767               & 11.8                 & 38.2                 & 0.7800               \\
en-bg &                          &                          &                            & 38.0    & 62.4   & 0.8680   & 0.9                  & 14.5                 & 0.6774               & 30.0                 & 51.9                 & 0.8122               \\
en-bn &                          &                          &                            & 16.0    & 45.2   & 0.8109   & 16.6                 & 45.9                 & 0.8282               & 16.8                 & 46.1                 & 0.8275               \\
en-bs & \multicolumn{1}{r}{4.7}  & \multicolumn{1}{r}{26.0} & \multicolumn{1}{r}{0.4314} & 26.8    & 53.9   & 0.8672   & \multicolumn{1}{l}{} & \multicolumn{1}{l}{} & \multicolumn{1}{l}{} & \multicolumn{1}{l}{} & \multicolumn{1}{l}{} & \multicolumn{1}{l}{} \\
en-ca & \multicolumn{1}{r}{38.4} & \multicolumn{1}{r}{61.7} & \multicolumn{1}{r}{0.8461} & 37.8    & 61.0   & 0.8334   & 39.8                 & 62.2                 & 0.8440               & 39.5                 & 62.1                 & 0.8425               \\
en-cy &                          &                          &                            & 50.4    & 69.9   & 0.8611   & 47.7                 & 67.6                 & 0.8536               & 48.4                 & 67.8                 & 0.8506               \\
en-eo &                          &                          &                            & 27.2    & 54.7   & 0.8264   & 25.5                 & 54.1                 & 0.8500               & 25.9                 & 54.4                 & 0.8523               \\
en-et & \multicolumn{1}{r}{23.7} & \multicolumn{1}{r}{53.4} & \multicolumn{1}{r}{0.8664} & 24.5    & 53.8   & 0.8684   & 24.5                 & 53.3                 & 0.8600               & 24.4                 & 53.3                 & 0.8578               \\
en-eu & \multicolumn{1}{r}{12.1} & \multicolumn{1}{r}{43.4} & \multicolumn{1}{r}{0.7674} & 16.5    & 49.5   & 0.8215   & 14.9                 & 47.2                 & 0.8098               & 14.8                 & 47.1                 & 0.8122               \\
en-fa &                          &                          &                            & 21.5    & 47.5   & 0.7947   & 23.4                 & 50.0                 & 0.8336               & 23.6                 & 50.0                 & 0.8349               \\
en-fi &                          &                          &                            & 21.3    & 51.1   & 0.8709   & 21.3                 & 51.2                 & 0.8725               & 22.1                 & 51.6                 & 0.8752               \\
en-ga & \multicolumn{1}{r}{27.3} & \multicolumn{1}{r}{52.6} & \multicolumn{1}{r}{0.7561} & 29.0    & 53.9   & 0.7543   & 30.2                 & 53.9                 & 0.7715               & 30.8                 & 54.6                 & 0.7717               \\
en-gl & \multicolumn{1}{r}{27.9} & \multicolumn{1}{r}{54.0} & \multicolumn{1}{r}{0.8033} & 30.0    & 55.7   & 0.8179   & 31.4                 & 56.1                 & 0.8302               & 31.4                 & 56.1                 & 0.8264               \\
en-gu &                          &                          &                            & 19.3    & 46.5   & 0.8066   & 22.5                 & 49.9                 & 0.8518               & 22.6                 & 49.9                 & 0.8479               \\
en-he &                          &                          &                            & 28.1    & 54.0   & 0.8320   & 29.7                 & 55.9                 & 0.8532               & 29.6                 & 55.4                 & 0.8503               \\
en-hi & \multicolumn{1}{r}{32.8} & \multicolumn{1}{r}{55.5} & \multicolumn{1}{r}{0.7621} & 32.0    & 54.6   & 0.7612   & 33.1                 & 55.5                 & 0.7728               & 32.5                 & 54.9                 & 0.7658               \\
en-hr &                          &                          &                            & 27.5    & 53.7   & 0.8504   & \multicolumn{1}{l}{} & \multicolumn{1}{l}{} & \multicolumn{1}{l}{} & \multicolumn{1}{l}{} & \multicolumn{1}{l}{} & \multicolumn{1}{l}{} \\
en-is & \multicolumn{1}{r}{20.6} & \multicolumn{1}{r}{45.1} & \multicolumn{1}{r}{0.7651} & 22.2    & 47.1   & 0.7766   & 22.8                 & 47.1                 & 0.7800               & 23.1                 & 47.5                 & 0.7859               \\
en-ja &                          &                          &                            & 27.0    & 26.6   & 0.8244   & 29.9                 & 30.2                 & 0.8640               & 29.6                 & 29.9                 & 0.8633               \\
en-kk &                          &                          &                            & 21.0    & 51.4   & 0.8651   & 16.5                 & 45.1                 & 0.8315               & 16.9                 & 45.3                 & 0.8347               \\
en-kn &                          &                          &                            & 13.8    & 43.5   & 0.7746   & 19.5                 & 50.8                 & 0.8348               & 19.2                 & 51.1                 & 0.8369               \\
en-ko &                          &                          &                            & 25.0    & 31.2   & 0.8268   & 26.6                 & 32.2                 & 0.8424               & 26.5                 & 32.0                 & 0.8402               \\
en-lt &                          &                          &                            & 22.9    & 50.9   & 0.8214   & 25.0                 & 52.7                 & 0.8486               & 24.4                 & 52.5                 & 0.8417               \\
en-lv &                          &                          &                            & 26.8    & 53.1   & 0.8214   & 23.9                 & 50.0                 & 0.7898               & 24.3                 & 50.6                 & 0.7891               \\
en-mk &                          &                          &                            & 32.2    & 58.6   & 0.8478   & 33.9                 & 60.1                 & 0.8718               & 34.1                 & 60.1                 & 0.8660               \\
en-ml &                          &                          &                            & 0.6     & 20.2   & 0.5753   & 14.4                 & 47.9                 & 0.8438               & 14.6                 & 48.0                 & 0.8427               \\
en-mr &                          &                          &                            & 11.0    & 37.9   & 0.6086   & 13.9                 & 42.3                 & 0.6808               & 14.2                 & 42.4                 & 0.6792               \\
en-ms &                          &                          &                            & 38.3    & 63.9   & 0.8580   & 24.4                 & 52.6                 & 0.8534               & 25.1                 & 53.2                 & 0.8540               \\
en-mt &                          &                          &                            & 34.7    & 63.2   & 0.7072   & 35.8                 & 64.1                 & 0.7102               & 36.0                 & 64.4                 & 0.7111               \\
en-nb &                          &                          &                            & 33.0    & 58.2   & 0.8661   & \multicolumn{1}{l}{} & \multicolumn{1}{l}{} & \multicolumn{1}{l}{} & \multicolumn{1}{l}{} & \multicolumn{1}{l}{} & \multicolumn{1}{l}{} \\
en-ne &                          &                          &                            & 0.9     & 20.9   & 0.4880   & 12.9                 & 42.8                 & 0.7404               & 12.8                 & 43.1                 & 0.7386               \\
en-nn &                          &                          &                            & 22.8    & 47.1   & 0.7766   & \multicolumn{1}{l}{} & \multicolumn{1}{l}{} & \multicolumn{1}{l}{} & \multicolumn{1}{l}{} & \multicolumn{1}{l}{} & \multicolumn{1}{l}{} \\
en-si &                          &                          &                            & 1.2     & 18.6   & 0.6289   & 13.1                 & 41.0                 & 0.8542               & 13.2                 & 41.2                 & 0.8548               \\
en-sk &                          &                          &                            & 29.3    & 54.0   & 0.8279   & 29.3                 & 53.9                 & 0.8353               & 29.9                 & 54.5                 & 0.8423               \\
en-sl &                          &                          &                            & 26.8    & 52.2   & 0.8295   & 26.5                 & 52.0                 & 0.8339               & 27.5                 & 52.6                 & 0.8414               \\
en-sq & \multicolumn{1}{r}{27.8} & \multicolumn{1}{r}{54.6} & \multicolumn{1}{r}{0.8509} & 27.7    & 54.2   & 0.8398   & 29.9                 & 55.8                 & 0.8659               & 29.3                 & 55.3                 & 0.8600               \\
en-sr &                          &                          &                            & 32.2    & 57.5   & 0.8512   & \multicolumn{1}{l}{} & \multicolumn{1}{l}{} & \multicolumn{1}{l}{} & \multicolumn{1}{l}{} & \multicolumn{1}{l}{} & \multicolumn{1}{l}{} \\
en-sw & \multicolumn{1}{r}{28.4} & \multicolumn{1}{r}{54.6} & \multicolumn{1}{r}{0.7743} & 32.5    & 58.2   & 0.7964   & 31.2                 & 57.0                 & 0.8058               & 31.3                 & 57.0                 & 0.8031               \\
en-ta &                          &                          &                            & 14.1    & 46.7   & 0.8139   & 16.7                 & 50.3                 & 0.8565               & 16.4                 & 50.1                 & 0.8553               \\
en-te &                          &                          &                            & 20.2    & 51.3   & 0.8104   & 22.1                 & 53.7                 & 0.8378               & 22.7                 & 53.9                 & 0.8383               \\
en-th &                          &                          &                            & 9.9     & 40.9   & 0.7977   & 8.1                  & 40.6                 & 0.8053               & 8.7                  & 40.9                 & 0.8053               \\
en-tr &                          &                          &                            & 25.3    & 53.7   & 0.8368   & 27.8                 & 56.4                 & 0.8685               & 27.5                 & 55.8                 & 0.8638               \\
en-uk &                          &                          &                            & 26.7    & 52.6   & 0.8457   & 27.2                 & 53.4                 & 0.8532               & 26.8                 & 52.8                 & 0.8471               \\
en-ur &                          &                          &                            & 18.9    & 43.2   & 0.7548   & 19.3                 & 44.0                 & 0.7537               & 19.5                 & 44.6                 & 0.7584               \\
en-uz &                          &                          &                            & 16.3    & 49.1   & 0.8397   & 15.9                 & 47.3                 & 0.8497               & 17.1                 & 48.8                 & 0.8532               \\
en-vi &                          &                          &                            & 37.8    & 55.8   & 0.8358   & 39.3                 & 57.1                 & 0.8489               & 38.8                 & 56.6                 & 0.8451               \\
en-xh &                          &                          &                            & 12.0    & 44.2   & 0.7323   & 0.0                  & 3.5                  & 0.2317               & 0.0                  & 3.5                  & 0.1647     \\
\bottomrule
\end{tabular}
}
\caption{MT results (BLEU, chrf++, COMET22-DA) for models translating from English, trained on our \HPLT{}, Tatoeba (OPUS), a combination of both, and the existing HPLT v1.2 (numbers reported where available).}
\label{tab:para_mt_eval_from_en}
\end{table*}

\begin{table*}[!htp]
\centering
\small
\adjustbox{max width=\linewidth}{
\begin{tabular}{rrrrrrrrrrrrr}
\toprule
      & \multicolumn{3}{c}{HPLT~v1.2}                                                    & \multicolumn{3}{c}{\HPLT{}} & \multicolumn{3}{c}{Tatoeba (OPUS)}                                           & \multicolumn{3}{c}{\HPLT{}+OPUS}                                 \\
\cmidrule(lr){2-4}\cmidrule(lr){5-7}\cmidrule(lr){8-10}\cmidrule(lr){11-13}
 & BLEU & chrF++ & COMET & BLEU & chrF++ & COMET & BLEU & chrF++ & COMET & BLEU & chrF++ & COMET \\
\midrule
az-en &                          &                          &                            & 18.5    & 47.1   & 0.8290   & 17.4                 & 44.7                 & 0.8039               & 18.6                 & 46.2                 & 0.8175               \\
be-en &                          &                          &                            & 16.1    & 46.7   & 0.7886   & 14.8                 & 44.7                 & 0.7621               & 15.5                 & 45.8                 & 0.7743               \\
bg-en &                          &                          &                            & 35.5    & 61.1   & 0.8556   & 7.4                  & 32.5                 & 0.5104               & 34.8                 & 60.6                 & 0.8524               \\
bn-en &                          &                          &                            & 27.9    & 53.7   & 0.8468   & 28.4                 & 53.7                 & 0.8498               & 29.0                 & 54.2                 & 0.8523               \\
bs-en & \multicolumn{1}{r}{12.8} & \multicolumn{1}{r}{38.0} & \multicolumn{1}{r}{0.5882} & 35.2    & 59.9   & 0.8508   & \multicolumn{1}{l}{} & \multicolumn{1}{l}{} & \multicolumn{1}{l}{} & \multicolumn{1}{l}{} & \multicolumn{1}{l}{} & \multicolumn{1}{l}{} \\
ca-en & \multicolumn{1}{r}{41.0} & \multicolumn{1}{r}{64.4} & \multicolumn{1}{r}{0.8676} & 39.2    & 63.1   & 0.8478   & 41.1                 & 64.4                 & 0.8580               & 40.3                 & 63.9                 & 0.8541               \\
cy-en &                          &                          &                            & 51.5    & 70.7   & 0.8615   & 50.0                 & 68.8                 & 0.8456               & 50.6                 & 69.1                 & 0.8455               \\
eo-en &                          &                          &                            & 35.5    & 59.2   & 0.8362   & 35.6                 & 59.2                 & 0.8437               & 35.6                 & 59.3                 & 0.8429               \\
et-en & \multicolumn{1}{r}{30.6} & \multicolumn{1}{r}{56.6} & \multicolumn{1}{r}{0.8611} & 30.3    & 55.8   & 0.8517   & 30.7                 & 56.1                 & 0.8510               & 30.7                 & 55.6                 & 0.8510               \\
eu-en & \multicolumn{1}{r}{19.4} & \multicolumn{1}{r}{45.7} & \multicolumn{1}{r}{0.7810} & 23.3    & 49.2   & 0.8121   & 22.2                 & 47.5                 & 0.8064               & 22.0                 & 47.4                 & 0.8042               \\
fa-en &                          &                          &                            & 31.1    & 56.4   & 0.8447   & 33.7                 & 58.2                 & 0.8585               & 32.7                 & 57.6                 & 0.8546               \\
fi-en &                          &                          &                            & 26.6    & 52.4   & 0.8442   & 26.6                 & 51.9                 & 0.8451               & 26.2                 & 51.6                 & 0.8421               \\
ga-en & \multicolumn{1}{r}{29.9} & \multicolumn{1}{r}{54.9} & \multicolumn{1}{r}{0.7653} & 34.1    & 58.8   & 0.8006   & 32.3                 & 56.3                 & 0.7754               & 33.1                 & 57.7                 & 0.7918               \\
gl-en & \multicolumn{1}{r}{31.4} & \multicolumn{1}{r}{57.2} & \multicolumn{1}{r}{0.8236} & 33.7    & 59.2   & 0.8374   & 34.5                 & 59.1                 & 0.8395               & 35.0                 & 59.9                 & 0.8441               \\
gu-en &                          &                          &                            & 28.5    & 54.6   & 0.8475   & 32.0                 & 57.0                 & 0.8646               & 33.0                 & 57.6                 & 0.8667               \\
he-en &                          &                          &                            & 38.2    & 62.2   & 0.8534   & 39.7                 & 62.9                 & 0.8622               & 40.4                 & 63.6                 & 0.8665               \\
hi-en & \multicolumn{1}{r}{35.2} & \multicolumn{1}{r}{59.9} & \multicolumn{1}{r}{0.8741} & 34.7    & 59.5   & 0.8701   & 35.8                 & 60.1                 & 0.8738               & 36.9                 & 61.0                 & 0.8773               \\
hr-en &                          &                          &                            & 31.7    & 56.4   & 0.8389   & \multicolumn{1}{l}{} & \multicolumn{1}{l}{} & \multicolumn{1}{l}{} & \multicolumn{1}{l}{} & \multicolumn{1}{l}{} & \multicolumn{1}{l}{} \\
is-en & \multicolumn{1}{r}{25.3} & \multicolumn{1}{r}{50.0} & \multicolumn{1}{r}{0.7815} & 29.0    & 53.4   & 0.8189   & 29.0                 & 52.8                 & 0.8163               & 28.7                 & 52.8                 & 0.8136               \\
ja-en &                          &                          &                            & 19.9    & 46.8   & 0.8255   & 24.6                 & 52.5                 & 0.8628               & 23.6                 & 50.6                 & 0.8533               \\
kk-en &                          &                          &                            & 27.0    & 53.4   & 0.8403   & 22.6                 & 47.8                 & 0.8003               & 22.6                 & 47.7                 & 0.7998               \\
kn-en &                          &                          &                            & 3.8     & 24.5   & 0.6246   & 27.9                 & 53.4                 & 0.8391               & 27.4                 & 53.2                 & 0.8396               \\
ko-en &                          &                          &                            & 24.1    & 51.3   & 0.8458   & 25.7                 & 52.7                 & 0.8586               & 25.8                 & 52.7                 & 0.8578               \\
lt-en &                          &                          &                            & 26.5    & 51.9   & 0.8138   & 27.2                 & 52.6                 & 0.8224               & 27.0                 & 52.2                 & 0.8201               \\
lv-en &                          &                          &                            & 29.3    & 56.0   & 0.8368   & 25.0                 & 50.9                 & 0.7862               & 26.6                 & 53.3                 & 0.8113               \\
mk-en &                          &                          &                            & 37.0    & 61.4   & 0.8522   & 38.2                 & 62.0                 & 0.8558               & 38.8                 & 62.6                 & 0.8596               \\
ml-en &                          &                          &                            & 2.9     & 23.5   & 0.5978   & 26.4                 & 51.7                 & 0.8342               & 26.4                 & 51.9                 & 0.8363               \\
mr-en &                          &                          &                            & 23.8    & 49.8   & 0.8063   & 26.1                 & 51.9                 & 0.8299               & 26.9                 & 52.2                 & 0.8320               \\
ms-en &                          &                          &                            & 37.2    & 61.3   & 0.8561   & 38.5                 & 61.8                 & 0.8579               & 38.0                 & 61.7                 & 0.8583               \\
mt-en &                          &                          &                            & 45.0    & 68.4   & 0.7777   & 47.1                 & 68.5                 & 0.7892               & 47.5                 & 68.9                 & 0.7895               \\
nb-en &                          &                          &                            & 37.3    & 61.0   & 0.8540   & \multicolumn{1}{l}{} & \multicolumn{1}{l}{} & \multicolumn{1}{l}{} & \multicolumn{1}{l}{} & \multicolumn{1}{l}{} & \multicolumn{1}{l}{} \\
ne-en &                          &                          &                            & 22.1    & 47.9   & 0.7929   & 26.6                 & 52.0                 & 0.8408               & 27.7                 & 52.8                 & 0.8436               \\
nn-en &                          &                          &                            & 32.9    & 55.8   & 0.7945   & \multicolumn{1}{l}{} & \multicolumn{1}{l}{} & \multicolumn{1}{l}{} & \multicolumn{1}{l}{} & \multicolumn{1}{l}{} & \multicolumn{1}{l}{} \\
si-en &                          &                          &                            & 3.0     & 24.2   & 0.5979   & 26.0                 & 51.2                 & 0.8381               & 26.9                 & 51.9                 & 0.8418               \\
sk-en &                          &                          &                            & 31.6    & 58.1   & 0.8456   & 32.7                 & 58.6                 & 0.8486               & 33.2                 & 59.0                 & 0.8535               \\
sl-en &                          &                          &                            & 29.2    & 55.0   & 0.8371   & 28.7                 & 54.4                 & 0.8345               & 29.7                 & 55.6                 & 0.8402               \\
sq-en & \multicolumn{1}{r}{31.7} & \multicolumn{1}{r}{58.3} & \multicolumn{1}{r}{0.8468} & 32.1    & 58.6   & 0.8453   & 33.7                 & 58.8                 & 0.8448               & 34.8                 & 59.8                 & 0.8534               \\
sr-en &                          &                          &                            & 37.4    & 62.7   & 0.8544   & \multicolumn{1}{l}{} & \multicolumn{1}{l}{} & \multicolumn{1}{l}{} & \multicolumn{1}{l}{} & \multicolumn{1}{l}{} & \multicolumn{1}{l}{} \\
sw-en & \multicolumn{1}{r}{27.2} & \multicolumn{1}{r}{51.0} & \multicolumn{1}{r}{0.7542} & 35.3    & 57.8   & 0.8086   & 34.3                 & 56.3                 & 0.7979               & 33.5                 & 55.6                 & 0.7932               \\
ta-en &                          &                          &                            & 24.0    & 49.6   & 0.8068   & 23.4                 & 49.0                 & 0.8139               & 24.3                 & 49.5                 & 0.8154               \\
te-en &                          &                          &                            & 30.2    & 55.3   & 0.8328   & 31.5                 & 55.9                 & 0.8438               & 31.9                 & 56.4                 & 0.8446               \\
th-en &                          &                          &                            & 24.9    & 52.3   & 0.8452   & 22.9                 & 51.0                 & 0.8381               & 23.7                 & 51.7                 & 0.8410               \\
tr-en &                          &                          &                            & 29.5    & 54.9   & 0.8392   & 32.2                 & 57.3                 & 0.8622               & 32.7                 & 57.4                 & 0.8602               \\
uk-en &                          &                          &                            & 33.1    & 58.7   & 0.8444   & 33.4                 & 59.2                 & 0.8470               & 33.9                 & 59.6                 & 0.8478               \\
ur-en &                          &                          &                            & 26.3    & 52.1   & 0.8138   & 26.2                 & 50.9                 & 0.8097               & 27.4                 & 52.0                 & 0.8144               \\
uz-en &                          &                          &                            & 24.8    & 51.5   & 0.8110   & 23.6                 & 48.6                 & 0.7990               & 24.8                 & 50.0                 & 0.8064               \\
vi-en &                          &                          &                            & 32.0    & 56.4   & 0.8514   & 33.5                 & 57.9                 & 0.8602               & 32.9                 & 57.2                 & 0.8543               \\
xh-en &                          &                          &                            & 22.1    & 45.5   & 0.6703   & 0.1                  & 11.8                 & 0.2871               & 0.2                  & 13.1                 & 0.2628       \\
\bottomrule
\end{tabular}
}
\caption{MT results (BLEU, chrf++, COMET22-DA) for models translating into English,  trained on our \HPLT{}, Tatoeba (OPUS), a combination of both, and the existing HPLT v1.2 (numbers reported where available).}
\label{tab:para_mt_eval_into_en}

\end{table*}

\end{document}